\documentclass[11pt]{article}

\usepackage{cite}
\usepackage{amsmath,amssymb,amsfonts}
\usepackage{algorithmic}
\usepackage{graphicx}
\usepackage{textcomp}
\usepackage{float}
\usepackage{placeins}
\usepackage{fontawesome5}
\usepackage{colortbl}
\usepackage{xcolor}
\usepackage{tabularx}
\usepackage{booktabs}
\usepackage{amssymb}    
\usepackage{pifont}     

\definecolor{cInternal}{HTML}{2E6DB4}   
\definecolor{cExplicit}{HTML}{3F9440}   
\definecolor{cRetrieval}{HTML}{8E5BA6}  
\definecolor{cAgentic}{HTML}{E07C24}    
\definecolor{rowfill}{HTML}{F5F7FA}     
\definecolor{openclr}{HTML}{B0B4BA}     
\newcommand{\open}{\textcolor{openclr}{\small$\varnothing$ \textit{open}}}
\newcommand{\mh}[2]{\textcolor{#1}{\textbf{#2}}}

\usepackage{bm}
\makeatletter
\AtBeginDocument{\DeclareMathVersion{bold}
\SetSymbolFont{operators}{bold}{T1}{times}{b}{n}
\DeclareSymbolFont{NewLetters}{T1}{times}{b}{it}{}
\SetSymbolFont{NewLetters}{bold}{T1}{times}{b}{it}
\SetMathAlphabet{\mathrm}{bold}{T1}{times}{b}{n}
\SetMathAlphabet{\mathit}{bold}{T1}{times}{b}{it}
\SetMathAlphabet{\mathbf}{bold}{T1}{times}{b}{n}
\SetMathAlphabet{\mathtt}{bold}{OT1}{pcr}{b}{n}
\SetSymbolFont{symbols}{bold}{OMS}{cmsy}{b}{n}
\renewcommand\boldmath{\@nomath\boldmath\mathversion{bold}}}
\makeatother

\def\BibTeX{{\rm B\kern-.05em{\sc i\kern-.025em b}\kern-.08em
    T\kern-.1667em\lower.7ex\hbox{E}\kern-.125emX}}

\usepackage[utf8]{inputenc}
\usepackage[T1]{fontenc}
\usepackage[margin=1in]{geometry}
\usepackage{amsmath,amssymb,amsthm}
\usepackage{graphicx}
\usepackage{booktabs}
\usepackage{tabularx}
\usepackage{multirow}
\usepackage{enumitem}
\usepackage{authblk}
\usepackage[table,dvipsnames,svgnames,x11names]{xcolor}

\usepackage[hidelinks]{hyperref}
\usepackage{cleveref}
\usepackage{framed}
\usepackage[square,numbers]{natbib}
\definecolor{memparam}{HTML}{7F8C8D}
\definecolor{memexplicit}{HTML}{2471A3}
\definecolor{memretrieval}{HTML}{2E86C1}
\definecolor{memagentic}{HTML}{7D3C98}
\definecolor{panelbg}{HTML}{F4F6F7}

\newtheorem{definition}{Definition}
\newtheorem{remark}{Remark}

\newcommand{\msec}[1]{\medskip\noindent\textbf{#1}\quad}

\title{Memory in Deep Time-Series Models}
\author[1]{Minh Hoang Nguyen}
\author[1]{Huu Hiep Nguyen}
\author[1]{Manh Nguyen}
\author[1]{Van Dai Do}
\author[1]{Dung Nguyen}
\author[1]{Hung Le}

\affil[1]{Deakin Applied AI Initiative, Deakin University, Geelong, VIC, Australia}

\date{}

\begin{document}
\maketitle



\begin{abstract}
Deep learning for time series has progressed through successive architectural paradigms, from recurrent networks and transformers to structured state-space models, retrieval-augmented predictors, foundation models, and tool-using agents. These developments are typically studied in isolation, organized by architecture or modeling era. We argue that they can instead be viewed through a common question of \emph{how does a time-series model retain and access information beyond its immediate input?} This question is motivated by a fundamental limitation of conventional time-series modeling: information relevant to a prediction may lie far beyond a feasible input window, while compressing history into a fixed-size state can discard information that may become useful later. We formulate this challenge as a \emph{memory} problem and organize existing time-series methods along a spectrum from internal memory, encoded in parameters and fixed-size states, to external memory that is addressable, retrievable, and increasingly maintained by agents. We then develop a unified taxonomy of memory mechanisms and review three classes of external memory, including explicit modules, retrieval augmentation, and agentic stores, under a common framework for what is retained, how it is written and accessed, and how it persists. A cross-cutting analysis maps these mechanisms to time-series tasks and identifies gaps in both methods and evaluation. We conclude by outlining open problems in building memory systems that can selectively retain, retrieve, revise, and forget information as temporal environments evolve. The result is a framework for studying memory as a first-class dimension of time series modeling, independent of the underlying backbone. The accompanying paper collection is available at \href{https://github.com/DA2I2-SLM/Time-series-memory}{\texttt{https://github.com/DA2I2-SLM/Time-series-memory}}.

\end{abstract}



\maketitle


\section{Introduction}
\label{sec:intro}


\begin{figure*}[h]
    \centering
    \includegraphics[
        width=0.9\linewidth,
        trim=0 0 0 0,
        clip
    ]{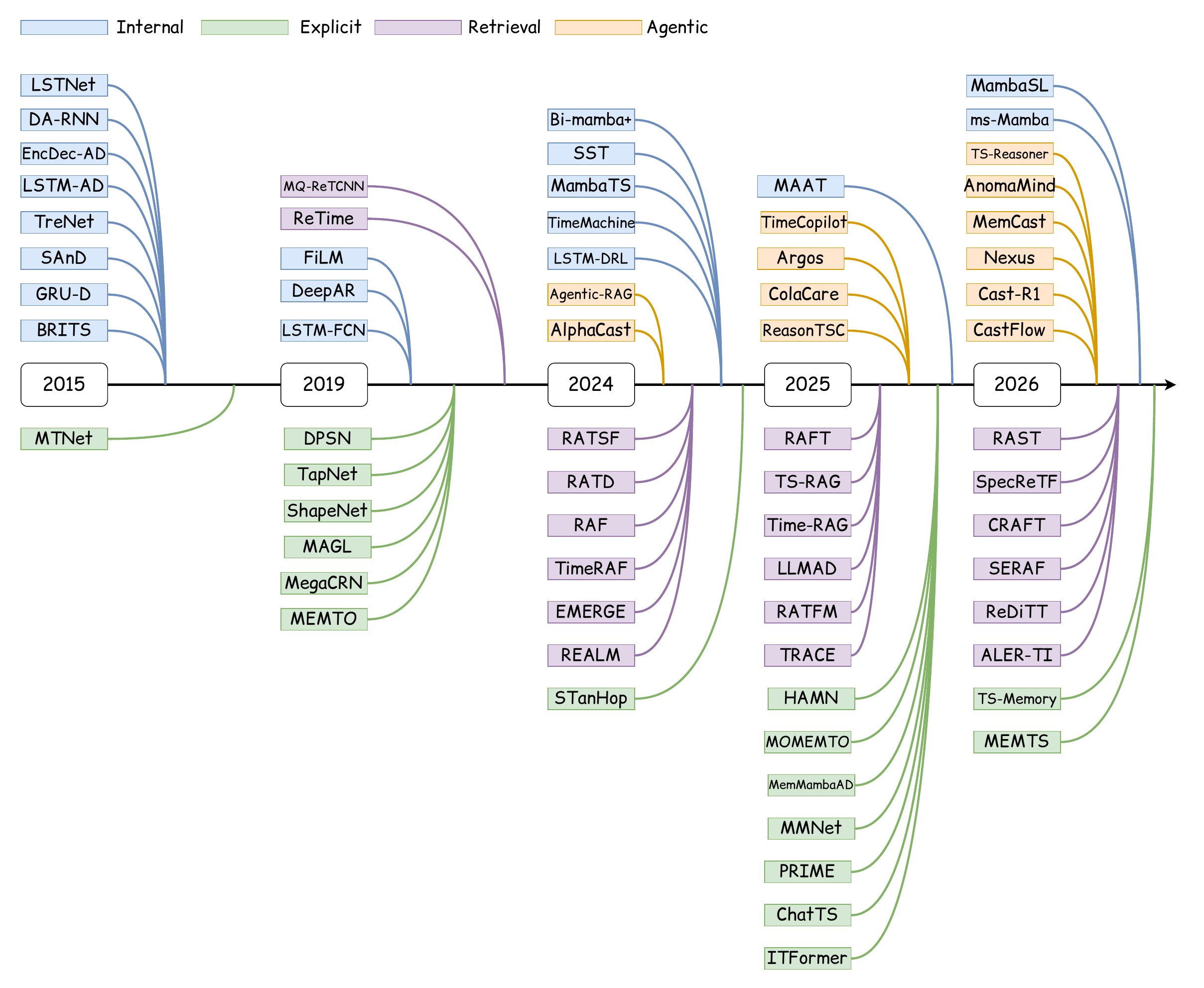}
    \caption{Evolution of memory mechanisms in time-series modeling from 2015 to 2026. Representative methods are organized along a temporal axis and categorized according to four increasingly explicit memory paradigms.}
    \label{fig:time_series_memory_evolution}
\end{figure*}

Time series underpin applications ranging from energy~\citep{tu2024powerpm,jiang2024probabilistic}, materials science \citep{11392029}, and finance~\citep{shi2026kronos} to healthcare~\citep{xu2023transehr,staniek2024early} and industrial monitoring~\citep{li2025MAD}, and the central modeling challenge is temporal dependence that reaches beyond any convenient input window. A decade of deep learning has produced a succession of architectural paradigms, including recurrence \cite{hochreiter1997lstm}, attention \cite{vaswani2017attention}, structured state \cite{gu2022s4,gu2023mamba}, retrieval \cite{liu2024ratd,han2025raft}, and, more recently, foundation models and agents \cite{garza2023timegpt,tao2026castr1}. These paradigms are typically surveyed apart, as though each represents a distinct stage in the evolution of time series modeling. We argue that they can instead be understood through a common notion of \emph{memory} (the mechanisms by which a model retains information from the past or broader context and accesses that information when producing an output). Recurrence compresses history into a hidden state, state-space models maintain a structured finite-dimensional summary, explicit memory modules retain addressable representations, retrieval systems use collections of observations as external memory, and agentic systems actively construct and maintain persistent stores. These mechanisms occupy different points on a common spectrum rather than forming disconnected architectural families. Figure~\ref{fig:time_series_memory_evolution} summarizes this evolution, tracing the progression from internal state-based memory toward increasingly explicit, retrievable, and actively maintained forms of memory. They differ primarily in whether memory is internal or external, whether its capacity is fixed or can grow with accumulated data, how information is written and read, and how long the resulting state persists.

While memory architectures have been extensively studied in Natural Language Processing (NLP)~\cite{zhong2024memorybank, borgeaud2022improving, zhang2025survey}, they remain comparatively under-surveyed in the time-series domain. Crucially, memory is not simply a lens that can be seamlessly transferred from language modeling because continuous temporal data impose requirements that make memory a fundamentally distinct modeling problem. Information relevant to a prediction often lies far outside any computationally feasible lookback window. Winter peak energy demand, for example, may depend entirely on what happened during the previous winter and a newly deployed machine may have little to no historical data of its own. Simply extending the context window is not always a practical solution, as longer sequences exponentially increase the cost of attention while burying important and rare observations given large amounts of irrelevant history. Conversely, compressing the past into a fixed-size hidden state creates the opposite problem. It forces the model to discard information that may later become important, especially when the useful signal is weak, rare, or separated from the current observation by a long time gap. Consequently, the specific mechanisms by which temporal memory is retained, retrieved, updated, and discarded must be treated as a distinct and central challenge in time-series modeling.

This motivates our focus on external memory, which we define as information that a model can read from, and in some cases write to, a store outside its learned parameters. We use internal memory as the conceptual baseline for this distinction, connecting recurrence and state-based approaches to this baseline while deferring detailed architectural comparisons to existing surveys~\citep{wen2023transformers,hewamalage2021recurrent}. Our review therefore focuses on how memory becomes explicit, retrievable, and actively maintained, and how these mechanisms support downstream time-series tasks. We organize the literature around forecasting, classification, anomaly detection, and imputation, while also examining emerging reasoning and decision-making settings in which memory may retain not only historical signals but also evidence, actions, outcomes, and evolving context.

Rather than organizing methods by backbone or retrieval mechanism, we classify them according to what they retain and how that information is written and accessed. This perspective reveals common memory mechanisms across systems that use different terminology. For example, PatchTST~\citep{nie2023patchtst} retains historical patches as separately addressable entries, while PM-MemNet stores representative historical patterns and retrieves those relevant to the current window~\citep{lee2022learning}. Despite their different terminology, both retain past information as addressable entries and selectively access it based on the current input. Their difference lies primarily in what is stored and how the entries are constructed, rather than in the underlying role of memory.

This perspective also changes how we position existing surveys. Prior surveys have largely been organized around architectures or broader modeling paradigms. Reviews of recurrent, transformer, and graph-based models typically treat memory as an implicit property of the backbone. In contrast, surveys of foundation models and agents focus on broader modeling or system paradigms. These perspectives are valuable, but they cover different parts of the memory spectrum and rarely connect them within a common framework. As summarized in \Cref{tab:relatedsurveys}, existing surveys provide substantial coverage of parametric and architectural approaches, while explicit, retrieval-based, and agentic memory receive limited or fragmented focus. Our survey instead makes memory the organizing principle, providing a unified view of memory. This complements rather than replaces architecture-centric and agent-centric accounts by focusing on what information persists, how it is written and retrieved, and when it is revised or discarded.

\begin{table*}[t]
\centering\small
\caption{Positioning against representative time series and memory surveys. Coverage of the memory spectrum: \checkmark{} substantial, ($\sim$) partial, blank little/none.}
\label{tab:relatedsurveys}
\begin{tabularx}{\textwidth}{Xcccccc}
\toprule
\textbf{Survey} & \textbf{Param.} & \textbf{Explicit} & 
\textbf{Retrieval} & \textbf{Agentic} & \textbf{Problem view} \\
\midrule
\rowcolor{rowfill}
\citet{lim2021dlsurvey}      & \checkmark & ($\sim$) & & & \\
\citet{wen2023transformers}   & ($\sim$) & & & & \\
\rowcolor{rowfill}
\citet{jin2024gnnsurvey}      & ($\sim$) & ($\sim$) & & & \\
\citet{liang2024fmtutorial}   & ($\sim$) & & ($\sim$) & ($\sim$) & \\
\rowcolor{rowfill}
\citet{tsagentsurvey2026}   & & & ($\sim$) & \checkmark & ($\sim$) \\
\textbf{This survey}        & ($\sim$) & \checkmark & \checkmark & \checkmark & \checkmark \\
\bottomrule
\end{tabularx}
\end{table*}

Our contributions are as follows.

\begin{itemize}
\item \textbf{A formal framework for memory in time series.} We define memory in terms of what information is retained, how it is accessed, and how long it persists, and introduce four dimensions for comparing memory mechanisms: representation, capacity, access mechanism, and persistence.

\item \textbf{A unified taxonomy of memory mechanisms.} We organize existing approaches along a spectrum from implicit parametric memory to explicit, retrieval-based, and agentic memory, together with a problem-by-architecture matrix that highlights common design choices across methods.

\item \textbf{A systematic review of external memory.} We examine the three major classes of external memory under a common framework covering their mechanisms, subclasses, writing and integration strategies, and limitations. A formal read/write formulation places explicit, retrieval-based, and agentic memory within a common lifecycle.

\item \textbf{A task-centric view of memory.} We analyze memory mechanisms across forecasting, classification, anomaly detection, imputation, reasoning, and decision-making, highlighting which memory mechanisms are well studied and which remain largely unexplored for each task.

\item \textbf{A critical analysis of evaluation and resources.} We review existing datasets, benchmarks, and evaluation protocols and argue that current evaluations largely measure downstream task accuracy rather than memory operations themselves. We identify four measurements for evaluating memory retention, retrieval quality, temporal validity, memory management.

\item \textbf{Open research directions.} We identify seven concrete challenges for building reliable future time-series memory systems, focusing on how memory can remain useful, efficient, and trustworthy as data and environments evolve.
\end{itemize}

The remainder of the survey is organized as follows. \Cref{sec:prelim} formalizes memory and introduces the dimensions used throughout the survey. \Cref{sec:whymemory} discusses why memory is particularly important for time series, while \Cref{sec:taxonomy} presents the proposed taxonomy and problem-by-architecture matrix. \Cref{sec:parametric} provides background on internal memory, and \Crefrange{sec:explicit}{sec:agentic} review the three external memory classes. \Cref{sec:crosscutting} provides the task-centric view, followed by the resources and benchmarks in \Cref{sec:resources} and the open challenges in Section~\ref{sec:challenges}.

Terminology introduced throughout the survey, including the four memory classes, their subtypes, the four problem settings, and the evaluation measures of \Cref{sec:resources}, is collected in a glossary in Supplementary Material~\ref{sec:glossary}.

\section{Preliminaries}
\label{sec:prelim}


\subsection{A Generic Memory-Augmented Model}

Let $\mathbf{x}_{1:T}=(\mathbf{x}_1,\ldots,\mathbf{x}_T)$ denote a multivariate time series, where $\mathbf{x}_t\in\mathbb{R}^d$ and $d$ is the number of variables or channels. At an operational step $t$, standard models receive an observation context $\mathbf{o}_t$, e.g., an input window such as $\mathbf{o}_t=\mathbf{x}_{t-L+1:t}$, while memory-augmented models additionally access information retained from observations outside this immediate context. We denote this retained information by the memory state $\mathcal{M}_t$. A generic memory-augmented model can therefore be expressed as

\begin{equation}
\begin{aligned}
    \mathcal{M}_t
    &= \mathcal{W}_{\phi}(\mathcal{M}_{t-1}, \mathbf{o}_t, Q), \\
    \mathbf{r}_t
    &= \mathcal{R}_{\psi}(\mathcal{M}_t, \mathbf{o}_t, Q), \\
    \hat{\mathbf{y}}_t
    &= \mathcal{P}_{\theta}(\mathbf{o}_t, \mathbf{r}_t, Q).
\end{aligned}
\label{eq:generic-memory}
\end{equation}
where $\mathcal{W}_{\phi}$ writes information to memory, $\mathcal{R}_{\psi}$ reads information from memory, and $\mathcal{P}_{\theta}$ produces the task output. Here, $Q$ denotes an optional task or query specification. This factorization separates three roles that recur throughout the survey: what is retained, how it is accessed, and how the retrieved information is used.

\subsection{What is memory? A working definition}
We use \emph{time-series memory} to denote information that is structured and retained such that it remains conditionally accessible beyond the immediate input time-series context. Crucially, the difference between memory and standard parametric knowledge is drawn along the axis of \emph{representation and access} (refer to Definition~\ref{def:memory}). 

\begin{framed}
\begin{definition}[Memory of a model]
\label{def:memory}
The \emph{memory} of a time-series model refers to any mechanism that makes historical information available to the predictor beyond the information contained in its current prediction target. This may occur through direct access to a finite context window, a dynamically maintained state, or a separately maintained store. Memory mechanisms therefore differ in how historical information is represented, accessed, updated, and persisted, as well as in their memory capacity. Information implicitly encoded in static model parameters (which remain fixed after training) is not considered memory. In contrast, dynamically updated parameter stores—such as fast weights that adapt during inference to retain recent history—function as accessible information and fall within our definition. We refer to this broad construct throughout the survey as ``time-series memory.''
\end{definition}
\end{framed}
Under Definition~\ref{def:memory}, the trained parameters $\theta$ (such as the convolutional filters or projection matrices of a network) do not constitute memory. They encode statistical regularities learned during training rather than retaining specific historical information for subsequent access. For example, parameter optimization updates the model according to:
\begin{equation}
    \theta \leftarrow \theta-\eta\nabla_{\theta}\mathcal{L},
\end{equation}
thereby incorporating knowledge from the training data into the model's parameters. Memory, in contrast, makes historical information available to the predictor through a context window, dynamic state, or dedicated store, with different mechanisms providing different forms of access, persistence, and capacity.

\subsection{Task Instantiations}
To demonstrate the operational framework, governed by the write operator $\mathcal{W}_\phi$, read operator $\mathcal{R}_\psi$, and prediction operator $\mathcal{P}_\theta$, we show how it explicitly instantiates across downstream tasks. In each case, downstream computation relies on a composite pipeline: a query $Q$ prompts the readout operator $\mathcal{R}_\psi$ to extract a contextual representation $\mathbf{r}_t = \mathcal{R}_\psi(\mathcal{M}_t, \mathbf{o}, Q)$ from memory, which the predictor $\mathcal{P}_\theta$ subsequently processes alongside the current observation. For readability, we define the functional shorthand $f_\theta(\mathbf{o}, \mathcal{M}_t) \equiv \mathcal{P}_\theta(\mathbf{o}, \mathcal{R}_\psi(\mathcal{M}_t, \mathbf{o}, Q), Q)$, collapsing retrieval and prediction into a single mapping.

\textbf{Forecasting.} Given a historical lookback window of length $L$, the predictor $\mathcal{P}_\theta$ conditions on input observations $\mathbf{x}_{t-L+1:t}$ and memory readout $\mathbf{r}_t$ to project a future horizon of $H$ values:
\begin{equation}
\begin{aligned}
    \hat{\mathbf{x}}_{t+1:t+H}
    &= \mathcal{P}_\theta\Big(
        \mathbf{x}_{t-L+1:t},
        \mathcal{R}_\psi(\mathcal{M}_t, \mathbf{x}_{t-L+1:t}, Q),
        Q
    \Big) \\
    &\equiv f_{\theta}\big(\mathbf{x}_{t-L+1:t}, \mathcal{M}_t\big).
\end{aligned}
\end{equation}

\textbf{Classification.} Given a fully observed trajectory $\mathbf{x}_{1:T}$, the model queries the aggregated memory $\mathcal{M}$ to map the sequence to a discrete class distribution:
\begin{equation}
\begin{aligned}
    \hat{y}
    &= \mathcal{P}_\theta\Big(
        \mathbf{x}_{1:T},
        \mathcal{R}_\psi(\mathcal{M}_T, \mathbf{x}_{1:T}, Q),
        Q
    \Big) \\
    &\equiv f_{\theta}(\mathbf{x}_{1:T}, \mathcal{M}_T).
\end{aligned}
\end{equation}

\textbf{Anomaly Detection.} The model evaluates an incoming observation $\mathbf{o}_t$ against context retrieved from historical memory via $\mathcal{R}_\psi$, outputting a step-level or window-level anomaly score $s_t$:
\begin{equation}
\begin{aligned}
    s_t
    &= \mathcal{P}_\theta\Big(
        \mathbf{o}_t,
        \mathcal{R}_\psi(\mathcal{M}_t, \mathbf{o}_t, Q),
        Q
    \Big) \\
    &\equiv f_{\theta}(\mathbf{o}_t, \mathcal{M}_t).
\end{aligned}
\end{equation}

\textbf{Imputation.} Let $\Omega$ represent the indices of observed entries and $\Omega^c$ denote missing values. The predictor reconstructs $\hat{\mathbf{x}}_{\Omega^c}$ by leveraging memory representations retrieved conditionally on the observed subset $\mathbf{x}_{\Omega}$:
\begin{equation}
\begin{aligned}
    \hat{\mathbf{x}}_{\Omega^c}
    &= \mathcal{P}_\theta\Big(
        \mathbf{x}_{\Omega},
        \mathcal{R}_\psi(\mathcal{M}_{\Omega}, \mathbf{x}_{\Omega}, Q),
        Q
    \Big) \\
    &\equiv f_{\theta}(\mathbf{x}_{\Omega}, \mathcal{M}_{\Omega}).
\end{aligned}
\end{equation}

\textbf{Reasoning.} Given a series observation $\mathbf{o}_t$ and a query $Q$, the model retrieves relevant information from memory and uses it to derive an inference. Depending on the setting, the output may be a textual answer, explanation, reasoning trace, or intermediate inference:
\begin{equation}
\begin{aligned}
\mathbf{y}
&= \mathcal{P}_\theta\Big(
\mathbf{o}_t,
\mathcal{R}\psi(\mathcal{M}_t, \mathbf{o}_t, Q),
Q
\Big) \\\
&\equiv f_{\theta}(\mathbf{o}_t, \mathcal{M}_t),
\end{aligned}
\end{equation}
where $\mathbf{y}$ denotes the task-dependent reasoning output. Unlike forecasting or classification, the objective is not necessarily to predict a numerical value or class, but to interpret the observed series and produce an inference supported by the available evidence.

\textbf{Decision-Making.} Given an observation $\mathbf{o}_t$, the model retrieves relevant historical information and uses it to select an action $a_t$. The action may subsequently alter the environment and generate new observations, actions, and outcomes that can be incorporated into memory:
\begin{equation}
\begin{aligned}
a_t
&= \pi_\theta\Big(
\mathbf{o}_t,
\mathcal{R}\psi(\mathcal{M}_t, \mathbf{o}_t, Q),
Q
\Big) \\\
&\equiv \pi_\theta(\mathbf{o}_t, \mathcal{M}_t),
\end{aligned}
\end{equation}
where $a_t$ denotes the selected action. In contrast to reasoning, whose output is an inference or reasoning artifact, decision-making produces an action that directly determines the agent's subsequent interaction. 

Crucially, these distinct objective functions do not necessitate architecturally separate memory mechanisms. Instead, they demonstrate the versatility of a unified memory substrate in supporting disparate downstream computations. We examine how different memory mechanisms instantiate this common formulation
across tasks in Section~\ref{sec:crosscutting}.

\subsection{Four Axes of Memory}
\label{sec:4axes}
We characterize memory mechanisms along four axes: representation, capacity, persistence, and access. These properties describe the memory state and its lifecycle rather than the particular downstream task.

\textbf{Representation} specifies the form and granularity of retained information. Depending on the mechanism, memory may be represented as:
\begin{equation}
\begin{aligned}
    \mathcal{M}_t &\in \mathbb{R}^{d_h}, \quad \text{or} \\
    \mathcal{M}_t &\in \mathbb{R}^{N_t \times d_m}, \quad \text{or} \\
    \mathcal{M}_t &= \{\mathbf{x}_i\}_{i\in\mathcal{I}_t}
\end{aligned}
\end{equation}
Here, $\mathbb{R}^{d_h}$ denotes an implicit fixed-dimensional state,
$\mathbb{R}^{N_t \times d_m}$ denotes a collection of explicit vector
entries, and $\{\mathbf{x}_i\}_{i\in\mathcal{I}_t}$ denotes retained
observations or other raw inputs.

\textbf{Capacity} specifies how retained information scales with the available data horizon. Let $N_t$ denote the number of memory entries retained at time $t$. We categorize this scaling dynamic into three distinct regimes:

\emph{(1) Architecturally Bounded Capacity} imposes a strict, static upper limit on the memory footprint:
\begin{equation}
    N_t\leq N_{\max}, 
\end{equation} 
where $N_{\max}$ is independent of the amount of observed data. Once this limit is reached, additional historical information must be compressed, overwritten, or otherwise discarded. 

\emph{(2) Data-Scaled Capacity} can increase as more observations become available, without a fixed architectural upper limit:
\begin{equation}
    N_t \to \infty \quad \text{as} \quad t \to \infty.
\end{equation}
New observations can therefore be added to memory as the data horizon grows, allowing the memory footprint to expand with the amount of accumulated information. The capacity is consequently determined by the data retained rather than by a fixed architectural limit.

\emph{(3) Policy-Bounded Capacity} can grow with incoming data in principle, but its effective size is constrained by an explicit retention policy:
\begin{equation}
   N_t\leq N_{\mathrm{policy}}(t),  
\end{equation}
where $N_{\mathrm{policy}}(t)$ is determined by factors such as eviction rules, storage budgets, or computational constraints. Unlike architecturally bounded memory, the limit is not inherent to the model architecture; instead, the system decides how much information to retain as memory grows.

Collectively, the three regimes provide an additional memory perspective for understanding the interplay between structural limits, data accumulation, and active state management.

\textbf{Persistence} describes how long retained information remains available. For an entry $m$ written at time $t_{\mathrm{write}}$ and removed at $t_{\mathrm{discard}}$, its lifetime is

\begin{equation}
    \tau(m) = t_{\mathrm{discard}}-t_{\mathrm{write}}. 
\end{equation}
We distinguish:

\emph{(1) Per-sequence persistence:} memory is reset when a new sequence or interaction begins.

\emph{(2) Per-dataset persistence:} memory is constructed from a dataset and remains fixed during deployment.
    
\emph{(3) Online persistence:} memory continues to evolve across deployment or interaction steps.

Capacity and persistence are independent. A memory can have bounded capacity but persist throughout deployment, or have data-scaled capacity while aggressively evicting old entries.

\textbf{Access} specifies how information is written to and read from memory. Writing may involve recurrent state updates, gated updates, appending new entries, overwriting existing entries, or controller-selected operations. Reading may use dense attention, similarity search, nearest-neighbor retrieval, key-value lookup, or a learned controller. The access axis captures not only whether memory can be read, but also how selectively information is written, addressed, updated, and forgotten.

\begin{remark}
These axes provide complementary dimensions for characterizing memory mechanisms. The taxonomy in \Cref{sec:taxonomy} builds on them to organize time-series memory approaches by memory architecture design, studying memory approaches that range from implicit, fixed-capacity state representations to explicit stores with data-scaled or policy-bounded capacity and increasingly active mechanisms for writing, retrieval, and consolidation.
\end{remark}

\subsection{Internal versus External Memory}
Equipped with these axes, we can broadly divide time-series memory into two dominant paradigms. \emph{Internal memory} resides in fixed-size latent states, such as recurrent cell states, where the capacity is architecturally bounded at design time, and the representation is strictly implicit. \emph{External memory}—categorized in this survey into explicit memory banks, retrieval memory indices, and agentic memory systems—is defined by its discrete representation and distinct access mechanisms. While explicit memory relies on a fixed, architecturally bounded capacity, retrieval and agentic mechanisms expand this to data-scaled and policy-bounded capacities, allowing the memory footprint to grow dynamically. This decoupling of storage from model parameters allows external stores to incorporate information never seen in training without changing a single weight. This survey centers on these external memory paradigms, treating internal memory primarily as a point of contrast, though the boundary is increasingly permeable in modern hybrid architectures.

Taken together, these axes and paradigms provide a common language for describing how memory is represented, scaled, and accessed across time-series models. However, characterizing memory mechanisms alone does not explain why different forms of memory are needed in the first place. We therefore first examine the sources of memory demand in time-series problems.

\section{Why Time Series Need Memory}
\label{sec:whymemory}

\begin{figure}[h]
      \centering
      \includegraphics[width=0.5\linewidth]{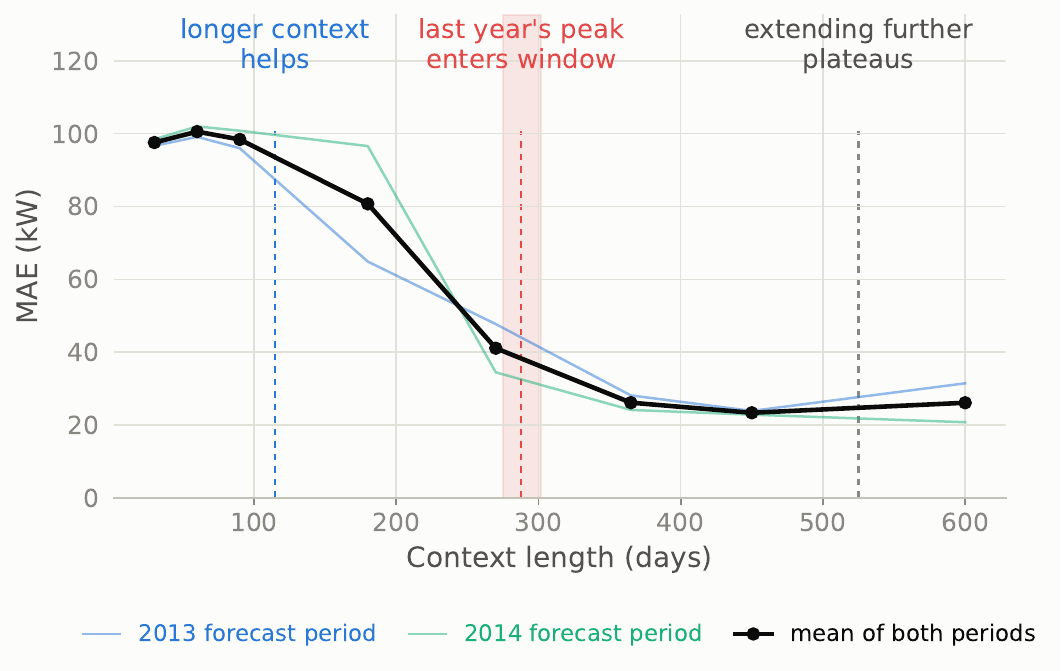}
      \caption{
           Zero-shot Chronos \citep{ansari2024chronos} error forecasting a real winter peak, vs.\ input
          context length (mean of two forecast periods; thin lines show each).
          Error stays flat under $\sim$90 days, then drops sharply once context
          crosses into the prior winter's peak ($\sim$275-302 days back), and
          plateaus past 450 days. The task needs long-range history, but a
          fixed-size window alone cannot fully exploit it.
      }
      \label{fig:context-length-sweep}
  \end{figure}

Before studying types of memory, we first need to understand why time-series models rely on them. Time-series models need memory when useful predictive information lies outside the immediate input window or is difficult to recover from a long history. Standard lookback windows work well when the past is recent and compact, but they fall short when key signals are distant, scattered across other series, or observed once and needed much later. Returning to the winter electricity example from \Cref{sec:intro}, forecasting a winter demand peak depends on operational data from the previous winter, which is long gone from a standard 90-day context window. As a result, the performance of forecasting models stalls until the context window stretches far enough, yet extending it further yields diminishing returns, demonstrating the need for better memory (see \Cref{fig:context-length-sweep}). Similarly, an anomaly detector deployed on a newly commissioned machine has no historical baseline of its own and must rely on logs from older equipment. In both scenarios, the model cannot simply look at its recent history; it must store, retrieve, or maintain past information over long horizons.

\begin{figure*}[h]
      \centering
      \includegraphics[width=\textwidth]{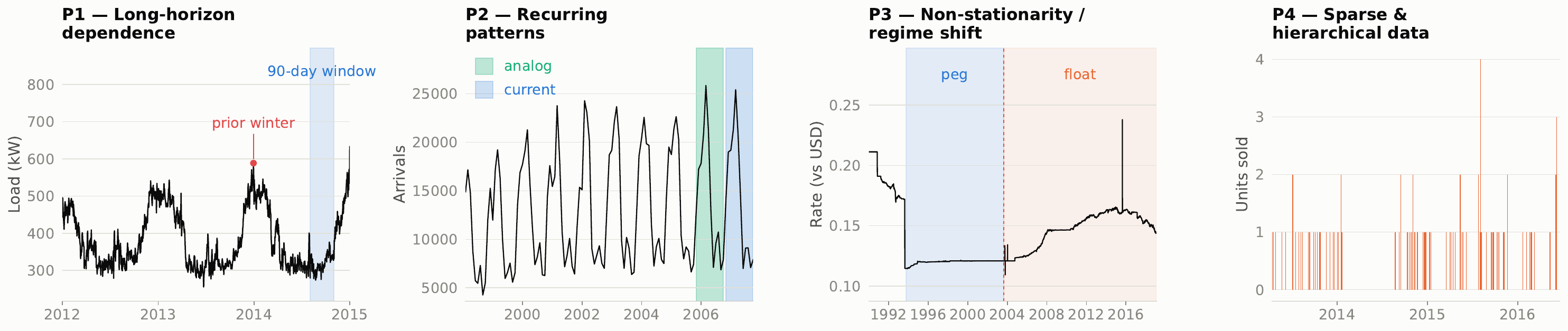}
      \caption{
Four real-world time-series examples motivating the problems in \Cref{sec:whymemory}.
  \textbf{P1:} winter-peaking electricity demand from the \emph{Electricity} dataset, where a 90-day autumn window misses the previous winter's peak.
  \textbf{P2:} monthly tourist arrivals from the \emph{Tourism} dataset exhibit a strong annual cycle, with the current window closely matching the same period one year earlier.
  \textbf{P3:} an exchange rate from the \emph{Exchange-Rate} dataset moving from a long peg to a floating regime, making old peg-era episodes misleading.
  \textbf{P4:} a product-store sales series from the \emph{M5} dataset with 93\% zero-sales days, requiring information from related series or higher aggregation levels.
      }
      \label{fig:why-memory}
  \end{figure*}

We group these challenges into four primary problem settings (see Figure \ref{fig:why-memory}), which define the structure of our taxonomy in \Cref{sec:taxonomy} and guide our analysis of specific mechanisms in \Cref{sec:explicit}--\Cref{sec:agentic}. These settings also highlight a fundamental architectural tradeoff: whether to compress historical context into a fixed, \emph{internal} state (such as updated weights or hidden vectors) or to store it in an \emph{external} memory that the model queries on demand.

\subsection{P1: Long-horizon dependence}
\label{sec:p1}

Time series frequently depend on events far back in their history. Unlike language, where context dependencies appear at arbitrary token distances, time-series dependencies are often periodic, driven by daily, weekly, or seasonal cycles. A retail forecaster, for example, relies on values recorded at exact annual offsets. Other exogenous drivers, such as macroeconomic shifts, climate trends, and regime changes, operate at slower frequencies and continuously influence the target variable over spans far exceeding practical window sizes.

Expanding the history window $L$ to capture these signals quickly becomes impractical. Standard self-attention scales quadratically with $L$, whereas linear models like DLinear~\citep{zeng2023dlinear} require larger input layers, diluting rare, informative points among irrelevant ones. Recurrent architectures and state-space models address this by compressing past information into a fixed hidden state. However, this internal compression is inherently lossy and often loses faint, distant signals. External memory architectures such as MTNet~\citep{chang2018memnet}, which stores historical blocks and retrieves relevant segments via attention, offer an alternative by maintaining distant context without continuous compression.

\subsection{P2: Recurring Patterns}
\label{sec:p2}

Many time series exhibit recurring behavior. When current conditions resemble a past event, the outcome of that past event provides strong predictive value. An unexpected demand spike is easier to forecast if the model can reference a similar spike from the previous year. Conversely, reconstruction-based anomaly detectors identify outliers by verifying that current inputs do not resemble any historical normal baseline~\citep{gong2019memorizing}.

Internal state compression is ill-suited for this task because it blends past episodes, destroying the distinct local dynamics needed for direct matching. Explicit, external memory solves this by storing past states as addressable entries. Models can either store raw historical windows to ground forecasts in historical precedents or maintain a bank of learned memory prototypes to evaluate incoming data. Beyond performance gains, explicit retrieval makes models far easier to audit, as predictions can be traced back to concrete historical examples.

\subsection{P3: Non-stationarity and regime shift}
\label{sec:p3}

Time series often undergo distribution shifts due to evolving market dynamics, sensor drift, or changing consumer habits. When the underlying data-generating process changes, historical data can become uninformative or actively misleading.

This creates a tension for memory design. On one hand, dynamic external stores allow models to adapt to distribution shifts without full retraining; new data points can be written to memory while outdated entries are removed. Fixed-state memory, by contrast, remains static after training. On the other hand, standard similarity-based retrieval assumes that past patterns remain valid indicators of future behavior. During a regime shift, retrieving a closely matching historical episode can lead to confident, yet erroneous predictions.

Handling non-stationary data requires active management policies rather than passive storage. Models must explicitly decide when to overwrite information, drop stale records, or ignore strong matches that belong to outdated regimes. This is a core motivation for the agentic memory write and update strategies detailed in \Cref{sec:agentic}.

\subsection{P4: Sparse and hierarchical data}
\label{sec:p4}

In many real-world applications, individual series lack sufficient local history to train a reliable model. For example, a newly launched product has no sales history, a newly deployed sensor lacks past baselines, and rare events offer few instances to learn from. In these cases, models must refer to related series or higher levels of aggregation. A new product can inherit trends from established items in the same category, while a regional sensor can leverage cross-series patterns across a spatial network. Unlike P1, which addresses temporal dependencies within a single series, P4 addresses spatial and structural dependencies across multiple series and hierarchical levels.

Memory mechanisms address data sparsity by maintaining a shared repository of global features across series. Frameworks like memory-augmented graph networks and hierarchical memory modules store cross-series patterns~\citep{liu2022magl}, allowing data-scarce nodes to pull context from the broader system. Shared external memory is naturally suited to this task, as a single series' internal state cannot easily transfer learned representations across distinct streams.

\begin{figure*}[t]
\centering
\includegraphics[width=\textwidth]{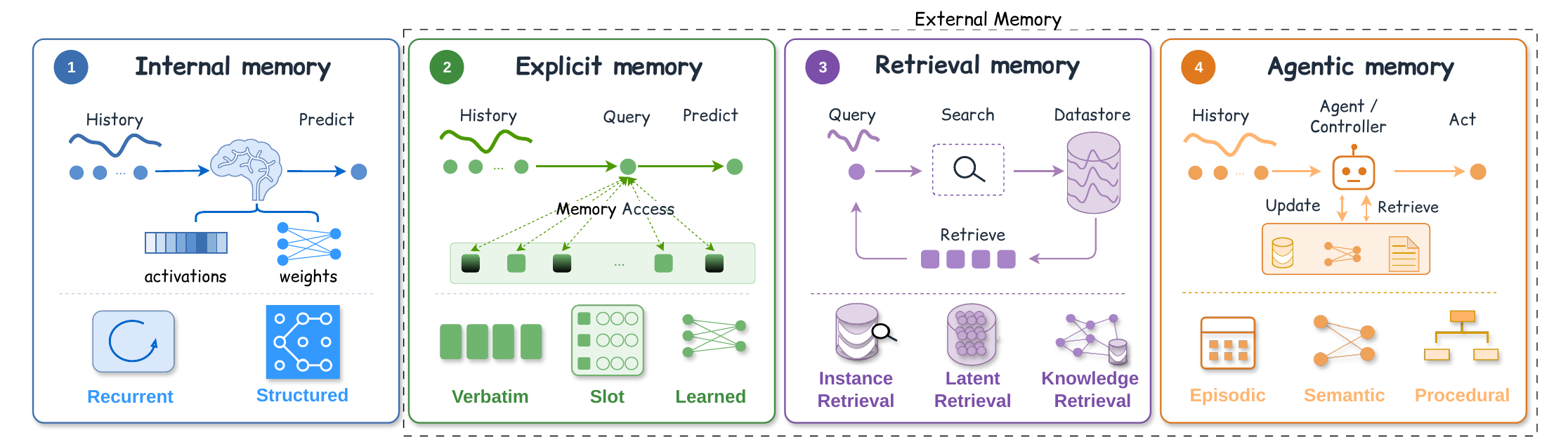}
\caption{Taxonomy of memory architectures for time-series models. (1) Internal memory retains historical information in dynamic activations, with learned parameters governing how the state is updated. It includes recurrent and structured state-space mechanisms. External memory includes: (2) explicit memory maintains dedicated addressable entries, (3) retrieval memory accesses an external collection, and (4) agentic memory actively manages what is stored and retrieved. The figure also shows the main subclasses within each memory architecture.}
\label{fig:taxonomy}
\end{figure*}

\subsection{Concerns shared across problems}

Two additional practical challenges influence memory selection across all four problem settings:

\begin{itemize}
    \item \textbf{Exogenous covariates:} Target series are frequently governed by exogenous variables such as news events, weather conditions, or macroeconomic indicators that lie outside the target time series itself. Memory architectures must be designed to store and index these external covariates alongside target histories.
    \item \textbf{Auditability:} In safety-critical domains like finance, healthcare, and industrial control, model predictions must be verifiable. External memory systems provide an explicit audit trail by identifying the exact past episodes or prototypes that triggered an output.
\end{itemize}

\section{Taxonomy of Time-series Memory}
\label{sec:taxonomy}

The problems in \Cref{sec:whymemory} necessitate memory mechanisms and architectures that preserve useful information from the past in an effective form for prediction. How this information is preserved varies substantially across time-series models. Some models compress the observed history into an internal state, while others keep individual observations or learned patterns explicitly available. External retrieval further allows models to draw on historical information, while more recent approaches actively manage what information should be stored, updated, or discarded.

We organize these approaches into four broad classes: \emph{internal memory}, \emph{explicit memory}, \emph{retrieval memory}, and \emph{agentic memory}. The difference is mainly architectural, based on where historical information is stored and how the model interacts with it. \Cref{fig:taxonomy} summarizes the four classes and their main subclasses. In the previous section, the problems describe \emph{why} memory is needed, while here the taxonomy describes \emph{how} that memory is implemented.

\subsection{Four memory architectures}
\label{sec:fourclasses}

\msec{Internal memory (\Cref{sec:parametric}).}
Internal memory retains past information within the model rather than in a separate memory store. We review two common forms. \emph{Recurrent-state memory}, as used in RNNs, LSTMs, and GRUs~\cite{elman1990finding,hochreiter1997lstm,cho2014learning}, carries a hidden state forward and updates it with each new observation. \emph{Structured-state memory}, including state-space models \cite{gu2023mamba}, summarizes a longer history into a structured, fixed-size representation that preserves useful temporal information over long sequences. In both cases, prediction depends on an internal state that has accumulated information from earlier observations.

The advantage is simplicity since memory is part of the model computation and requires no separate storage or retrieval step. However, historical information must be compressed into a bounded representation, so distant or infrequent events may be weakened or lost. This makes internal memory a useful baseline for P1, but less suitable when particular historical episodes must remain individually accessible.

\msec{Explicit memory (\Cref{sec:explicit}).}
Explicit memory is the simplest form of external memory. It preserves information as individually addressable entries rather than compressing the entire history into a single state. We distinguish three main mechanisms. \emph{Verbatim memory} keeps past observations directly accessible, as in Transformer attention~\cite{vaswani2017attention}. \emph{Slot memory} uses dedicated memory locations that store selected states or representations~\cite{graves2016hybrid,NEURIPS2018_e57c6b95}. \emph{Learned memory} stores compact patterns such as prototypes or shared representations. In time-series models, these memories make past observations or their representations explicitly addressable, rather than requiring historical information to be compressed into a hidden state. Their main limitation is thus capacity, which is tied to the context length, number of slots, or size of the learned memory.

\msec{Retrieval memory (\Cref{sec:retrieval}).}
Retrieval memory stores a larger collection outside the model and searches it when needed. Unlike explicit memory, the retrieved information can come from long historical archives, other series, or external knowledge. Retrieval memory has three main subclasses. \emph{Instance retrieval memory} stores concrete cases such as past windows or trajectories~\cite{Khandelwal2020Generalization}. \emph{Latent retrieval memory} stores compact representations such as embeddings. \emph{Knowledge retrieval memory} stores external information such as text, events, or structured domain knowledge~\cite{lewis2020rag,guu2020retrieval}. Their main advantage is scale and cross-series access, while the central challenge is retrieving information that is truly relevant rather than merely similar.

\msec{Agentic memory (\Cref{sec:agentic}).}
Agentic memory allows an agent to store and update information from past interactions. We categorize three main forms based on what is stored. \emph{Episodic memory} records specific past events, such as previous forecasting episodes and their outcomes. \emph{Semantic memory} stores more general knowledge learned from past observations, such as recurring regimes or relationships between variables. \emph{Procedural memory} stores knowledge about how to perform a task, such as which time-series tools or retrieval strategies to use in different situations~\cite{shinn2023reflexion}. Many agentic architectures combine two or more of these memory types.

Unlike retrieval memory, agentic memory can be changed based on new observations and outcomes. The agent can add new experiences, update its knowledge, or change how it handles similar situations in the future. This makes agentic memory useful when data or tasks change over time, but it also requires the agent to decide what information to keep, update, or remove.

\begin{table*}[h]
\centering\small
\renewcommand{\arraystretch}{1.35}
\caption{Relationship between the four time-series memory problems and memory architectures. \textbullet\textbullet{} dominant, \textbullet{} established but less prevalent, and ($\circ$) emerging or underexplored.}
\label{tab:matrix}
\setlength{\tabcolsep}{5pt}
\begin{tabularx}{\textwidth}{l >{\centering\arraybackslash}X >{\centering\arraybackslash}X >{\centering\arraybackslash}X >{\centering\arraybackslash}X}
\toprule
\textbf{Problem}
  & \multicolumn{1}{c}{\color{cInternal}\textbf{Internal}}
  & \multicolumn{1}{c}{\color{cExplicit}\textbf{Explicit}}
  & \multicolumn{1}{c}{\color{cRetrieval}\textbf{Retrieval}}
  & \multicolumn{1}{c}{\color{cAgentic}\textbf{Agentic}} \\
\midrule

\rowcolor{rowfill}
{P1--Long-horizon dependence}
  & \textbullet\textbullet & \textbullet & \textbullet\textbullet & \textbullet \\

{P2--Recurring patterns}
  & \textbullet & \textbullet\textbullet & \textbullet\textbullet & \textbullet \\

\rowcolor{rowfill}
{P3--Non-stationarity/drift}
  & ($\circ$) & ($\circ$) & \textbullet & \textbullet \\

{P4--Sparse/hierarchical data}
  & \textbullet & \textbullet\textbullet & \textbullet & ($\circ$) \\

\bottomrule
\end{tabularx}
\end{table*}

\subsection{Connecting problems and memory architectures}
\label{sec:matrixview}

The four architectures are not tied one-to-one to the four problems in \Cref{sec:whymemory}. A long-horizon dependency, for example, can be handled by maintaining a recurrent or state-space representation, explicitly storing historical segments, or retrieving the relevant period from an external archive. \Cref{tab:matrix} summarizes these relationships. Internal memory is widely used for long-horizon modeling (P1), particularly when historical information can be compressed into a useful latent state. Explicit and retrieval memory are natural choices for recurring patterns (P2), where retaining distinguishable historical examples enables direct matching. For non-stationarity (P3), retrieval allows the available context to change over time, while agentic memory additionally provides mechanisms for removing or updating stale information. Sparse and hierarchical settings (P4) have commonly used internal sharing and explicit cross-series memory to transfer information between related streams.

The matrix should therefore be read as a map of common architectural choices, not a strict assignment of problems to methods. Several architectures can address the same problem, and their suitability depends on the structure of the data and the type of historical information required.

\section{Internal Memory}
\label{sec:parametric}


\begin{framed}
\begin{definition}[Internal memory]
Internal memory refers to time-series models that encode historical observations through recurrently updated, fixed-dimensional latent states rather than explicitly storing the observed history. This paradigm enables models to incorporate increasingly long temporal histories by continuously compressing past observations into an internal hidden state. The state is updated according to learned dynamics, allowing it to capture temporal regularities, dependencies, and evolving patterns that are relevant to future predictions.
\end{definition}
\end{framed}

\subsection{Mechanism}
Internal memory can be viewed as compressing the observed history of a time series into a fixed-size dimensional representation. Given a time series $\mathbf{X} = \{\mathbf{x}_1, \mathbf{x}_2, \dots, \mathbf{x}_T\}$, where $\mathbf{X} \in \mathbb{R}^{T \times d}$ with $d$ denoting the variate number (number of channels) and $T$ the temporal sequence length, let $\mathbf{x}_t \in \mathbb{R}^d$ denote the input observation at time step $t$. The compression mechanism can be summarized as: 
\begin{equation}
\mathbf{h}_t = u_{\theta}(\mathbf{h}_{t-1}, \mathbf{x}_t) \in \mathbb{R}^{d_h},
\qquad
\widehat{\mathbf{y}}_t = f_{\theta}(\mathbf{o}_t, \mathbf{h}_t),
\label{eq:internal-memory}
\end{equation}
where $\mathbf{h}_t$ is the $d_h$-dimensional hidden state, $u_{\theta}$ updates it as new observations arrive, and $\mathbf{o}_t$ represents the local observation context available to the model at time $t$. This is the implicit-memory special case of the generic memory-augmented model in \Cref{eq:generic-memory}: the memory is carried by the hidden state, $\mathcal{M}_t \equiv \mathbf{h}_t$, while memory formation is performed by the recurrent update $u_\theta$. Retrieval is implicit, with $\mathcal{R}_\psi(\mathcal{M}_t,\mathbf{o}_t,Q) \equiv \mathbf{h}_t$, so the predictor evaluates both the local context and internal state via $f_\theta(\mathbf{o}_t, \mathbf{h}_t)$.

Repeated updates make $\mathbf{h}_t$ depend on the entire history, allowing it to retain information beyond the current input window. Its fixed dimension $d_h$ gives internal memory an architecturally bounded capacity (refer to \Cref{sec:4axes}). The compression of history into a fixed-size dimensional representation, however, makes internal memory difficult to inspect, address, or revise at the level of individual observations or past events.

\begin{figure}[t]
\centering
\includegraphics[width=0.6\columnwidth]{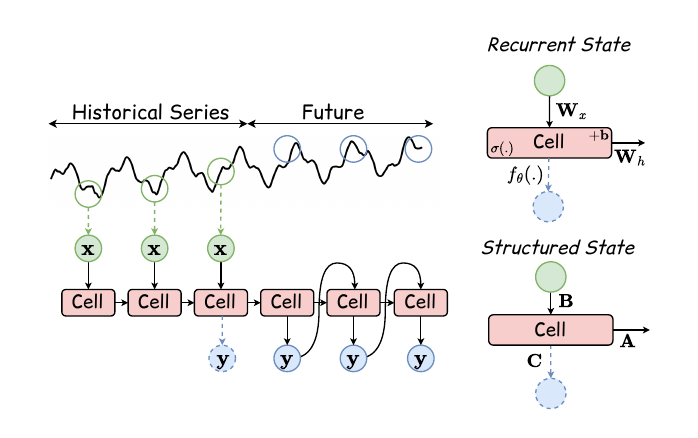}
\caption{Illustration of internal memory in time-series forecasting. The figure shows two approaches to compressing historical observations into a fixed-dimensional internal state: recurrent-state memory, where the state is updated through nonlinear recurrent transitions, and structured state memory, where temporal history is encoded through parameterized state-space dynamics.}
\label{fig:internal_state}
\end{figure}

We therefore view internal memory as the starting point of the time-series memory spectrum, where the emphasis is on learning compact summaries of temporal history before moving towards mechanisms that preserve past observations as persistent and addressable information. For readers seeking more details on the recurrent paradigm and its architectural variants in this domain, several dedicated surveys provide comprehensive reviews~\citep{hewamalage2021recurrent, wen2023transformers}. Figure~\ref{fig:internal_state} illustrates internal memory mechanisms through two representative approaches: recurrent-state memory and structured state memory.

\subsection{Types of Internal Memory}
\label{sec:internal-types}

\subsubsection{Recurrent state}
\label{sec:internal-recurrent}


RNNs~\cite{elman1990finding} established the fundamental formulation of recurrent internal temporal memory by maintaining an evolving hidden state that summarizes previous observations. Unlike models that process each observation independently, recurrent architectures update their internal state according to both the current input and the accumulated historical context, allowing temporal dependencies to be represented through sequential state evolution. In its generic form, a recurrent model instantiates \Cref{eq:internal-memory} with local observation context $\mathbf{o}_t = \mathbf{x}_{t-L+1:t}$ through a one-step update,
\begin{equation}
\begin{aligned}
\mathbf{h}_t &= \sigma\!\left(\mathbf{W}_{h}\mathbf{h}_{t-1} + \mathbf{W}_{x}\mathbf{x}_t + \mathbf{b}\right), \\
\widehat{\mathbf{y}}_t &= f_{\theta}(\mathbf{x}_{t-L+1:t}, \mathbf{h}_t),
\end{aligned}
\label{eq:recurrent}
\end{equation}
where $\sigma$ is an elementwise nonlinearity, $\mathbf{W}_h \in \mathbb{R}^{d_h \times d_h}$ and $\mathbf{W}_x \in \mathbb{R}^{d_h \times d}$ are learned parameters that govern state transitions and input encoding, respectively, and $\mathbf{b}$ is a bias. The hidden state $\mathbf{h}_t$, rather than these learned parameters, carries information from the observed history.


One of the difficulties is that early recurrent networks can struggle to preserve information over long horizons because repeated state updates caused gradients to vanish or explode. LSTMs~\cite{hochreiter1997lstm} address this limitation through gated memory updates, which control what information is retained, discarded, and exposed at each time step. GRUs~\cite{cho2014learning} provide a simpler gated alternative by combining the mechanisms for updating and resetting the recurrent state. These architectures become important building blocks for time series modeling because they provide a natural mechanism for representing dependencies.

A substantial body of time series forecasting work subsequently adapted recurrent memory to better exploit the structure of real-world temporal data. DeepAR~\cite{salinas2020deepar}, for example, uses an autoregressive recurrent state to capture temporal patterns shared across related series. DA-RNN~\cite{qin2017dual} enhances recurrent memory with dual-stage attention, allowing its recurrent encoder--decoder states to selectively incorporate informative variables and historical information when generating forecasts. TreNet~\cite{lin2017trenet} combines convolutional layers with LSTMs, using local temporal features to inform recurrent representations of longer-term dependencies, while LSTNet~\cite{lai2018modeling} extends this idea with a convolutional-recurrent architecture, a skip-RNN for long-range periodic patterns, and an autoregressive component for highly autocorrelated series. 

The same recurrent memory principle extends beyond forecasting. Multivariate LSTM-FCNs~\cite{karim2019multivariate} combine fully convolutional pathways with recurrent blocks for time series classification, demonstrating that the hidden state can encode discriminative temporal patterns. LSTM-based methods have likewise been used for anomaly detection, where the recurrent state provides a compact representation of expected temporal behavior against which deviations can be identified~\cite{Malhotra2015LongST}. Encoder--decoder variants further exploit recurrent states to model normal multi-sensor dynamics and detect anomalous sequences~\cite{malhotra2016lstm}. Recurrent memory has also been adapted to incomplete observations. GRU-D~\cite{che2018recurrent} modifies internal cell updates with trainable exponential decay mechanisms for missing intervals, while BRITS~\cite{3327757.3327783} uses twin bidirectional recurrent states to iteratively impute missing values in multivariate sequences.

In more specialized settings, SAnD~\cite{song2018attend} layers self-attention over LSTM-based representations to selectively identify informative variables and temporal patterns in clinical time series, while CloudLSTM~\cite{zhang2021cloudlstm} extends recurrent modeling to spatiotemporal point-cloud streams by introducing a Dynamic Point-cloud Convolution operator that extracts local spatial correlations from irregular point coordinates directly within the recurrent cell, demonstrating how recurrent states can encode evolving spatial as well as temporal dependencies. As we can see, combining recurrent memory with complementary paradigms, such as convolutions for local feature extraction, attention for selective routing, or linear components for scale, makes these architectures powerful, proving that the core principle of folding history into a compact hidden state remains versatile.

Interestingly, this recurrent view of memory has not disappeared with the rise of newer sequence architectures. P-sLSTM~\citep{kong2025unlocking} revisits LSTMs in the context of modern long-term forecasting and demonstrates that, with appropriate architectural and training choices, recurrent models can remain highly competitive against more recent methods.

\subsubsection{Structured state}
\label{sec:internal-structured}

State-space models approach the same problem from a different direction. Instead of designing increasingly sophisticated gates for a recurrent state, they ask whether the dynamics of the state itself can provide a more effective compression mechanism. Structured state-space models (SSMs)~\cite{gu2022s4} replace explicit gating operations with parameterized state transitions, where a fixed-dimensional hidden state continuously summarizes the input history. A general discrete-time state-space formulation mapping a time series observation $\mathbf{x}_t \in \mathbb{R}^d$ to a state $\mathbf{h}_t \in \mathbb{R}^{d_h}$ can be written as:
\begin{equation}
\mathbf{h}_t = \mathbf{A}\mathbf{h}_{t-1} + \mathbf{B}\mathbf{x}_t,
\qquad
\hat{\mathbf{y}}_t = \mathbf{C}\mathbf{h}_t,
\label{eq:ssm}
\end{equation}
which is a linear special case of \Cref{eq:recurrent}, obtained by
replacing the nonlinear transition $\sigma$ with the identity map and setting the transition matrix
$\mathbf{W}_{h}=\mathbf{A} \in \mathbb{R}^{d_h \times d_h}$, the input projection $\mathbf{W}_{x}=\mathbf{B} \in \mathbb{R}^{d_h \times d}$, and $\mathbf{b}=\mathbf{0}$. The readout is likewise specialized to the
linear state mapping $\widehat{\mathbf{y}}_t=\mathbf{C}\mathbf{h}_t$. Linearity makes the compression map available
in closed form,
\begin{equation}
\mathbf{h}_t
=
\mathbf{A}^{t}\mathbf{h}_0
+
\sum_{k=1}^{t}\mathbf{A}^{t-k}\mathbf{B}\mathbf{x}_k,
\label{eq:ssm-unrolled}
\end{equation}
so that $\mathbf{A}^{t-k}$ determines how the contribution of the observation at time $k$ propagates to the state at time $t$. This perspective has also been explored specifically for time series forecasting. FiLM~\cite{zhou2022film} introduces a frequency-enhanced Legendre memory model that combines structured state representations with frequency-domain information, illustrating how the capacity of a compact latent state can be improved by incorporating temporal structure beyond simple sequential recurrence.

More recent state-space models make this compression adaptive to the observed sequence. Mamba~\cite{gu2023mamba} introduces selective state spaces with input-dependent state transitions, allowing the model to selectively retain or discard information while maintaining efficient linear-time sequence processing. This idea has quickly been explored in time series forecasting, with TimeMachine~\cite{ahamed2024timemachine} investigating Mamba-based components for long-term forecasting and MambaTS~\cite{cai2024mambats} adapting selective state-space modelling specifically to long-term temporal dependencies. Architectural extensions have further refined these predictive capabilities; for instance, ms-Mamba~\cite{karadag2026ms} utilizes varying sampling rates to capture multi-scale temporal contexts, while Bi-Mamba+~\cite{liang2024bi} incorporates bidirectional processing to selectively integrate features over longer temporal ranges. The utility of selective state compression also extends well beyond forecasting. For time series classification, MambaSL~\cite{jung2026mambasl} demonstrates that an efficient single-layer Mamba architecture can achieve discriminative performance. Similarly, in anomaly detection, the MAAT~\cite{sellam2025mamba} integrates Mamba architectures with sparse attention mechanisms to effectively capture both short- and long-term temporal dependencies alongside association discrepancies. SST~\cite{patro2024sst} further combines Mamba and Transformer experts across temporal scales, reflecting the broader trend towards combining compressive state representations with mechanisms that can access richer contextual information. Attraos~\cite{hu2024attraos} further extends structured-state memory with a multi-resolution dynamic memory unit that captures historical dynamical structures across different temporal resolutions. At the same time, empirical analysis such as~\citet{wang2025smamba} suggests that selective state compression is not universally optimal, highlighting the importance of matching the memory mechanism to the structure of the underlying time series.

\subsection{Coverage and Limitations}

Despite the growing range of mechanisms proposed in the literature, the fundamental limitations of internal memory become clear when evaluated along our four axes. In terms of representation, whether history is accumulated through gated cell states, structured state transitions, or input-selective dynamics, the model ultimately distills the past into an implicit, fixed-dimensional vector. Because this state size is static, the memory's capacity is strictly architecturally bounded (refer to Section~\ref{sec:4axes}); it cannot grow as the data horizon expands, forcing constant compression and overwriting of old information. Consequently, the persistence of any single observation is highly volatile, typically lasting only until it is diluted by subsequent updates. Finally, the access mechanism is entirely implicit. A model may learn that an important event occurred, yet it cannot directly query or retrieve that event as a discrete stored record, nor selectively revise it after compression. This distinction separates internal memory from the external memory mechanisms considered next.

\begin{remark}
Internal memory provides an efficient mechanism for temporal modelling by encoding historical information in fixed-dimensional latent states, whose dynamics are governed by learned parameters. Its strength lies in compact representation and computational efficiency; however, its memory is constrained by fixed capacity, limited persistence, and implicit access. These limitations motivate the transition towards external memory mechanisms that support persistent storage, retrieval, and management of historical experiences 
(\Crefrange{sec:explicit}{sec:agentic}).
\end{remark}




\section{Explicit Memory}
\label{sec:explicit}

\begin{framed}
\begin{definition}[Explicit memory]
\label{def:explicit-memory}
Explicit memory refers to time-series models that hold information in a store of individually addressable entries (slots, vectors, or records), each written and read as a unit rather than entangled in weights or a dense state. At its core, this paradigm asks: \emph{how can a model keep specific pieces of the past available without folding them into a single shared representation?} The answer is to separate storage from computation: entries are created or revised individually and consulted through content-based addressing, so stored content can be inspected, edited, or expanded without changing the model that reads it.
\end{definition}
\end{framed}

\subsection{Mechanism}

Let the store be $\mathcal{M}_t = [\,\mathbf{m}_1,\dots,\mathbf{m}_{N_t}\,]^{\top} \in \mathbb{R}^{N_t \times d_m}$ with entry width $d_m$. For slot and learned stores, $N_t$ is a constant fixed before training. For verbatim stores, it grows as the window fills, but it never exceeds the context length $L$, and it resets once the model moves to the next sequence. Explicit memory is therefore bounded in both cases, which is different from the retrieval stores of \Cref{sec:retrieval}. An encoder maps the lookback window to a query $\mathbf{q}_t = g_{\phi}(\mathbf{x}_{t-L+1:t})$, and the \emph{read} resolves it against every entry by normalized similarity,
\begin{equation}
    w_{t,i} = \frac{\exp\!\big(s(\mathbf{q}_t,\mathbf{m}_i)\big)}{\sum_{j=1}^{N_t}\exp\!\big(s(\mathbf{q}_t,\mathbf{m}_j)\big)},
    \quad
    \mathbf{r}_t = \sum_{i=1}^{N_t} w_{t,i}\,\mathbf{m}_i,
\label{eq:explicit-read}
\end{equation}
with $s(\cdot,\cdot)$ typically cosine similarity or a scaled dot product. \Cref{eq:explicit-read} operates like an attention mechanism, sharpening the weights toward the most similar memory contents. The prediction conditions on the window to produce the readout, $\hat{\mathbf{y}}_t = \mathcal{P}_{\theta}(\mathbf{x}_{t-L+1:t}, \mathbf{r}_t)$, where $\hat{\mathbf{y}}_t$ is the task target. The \emph{write} takes one of two forms. Verbatim stores append, encoding a new element as a fresh entry and leaving the rest untouched, while slot stores revise existing entries across time steps:
\begin{equation}
    \mathbf{m}_i \leftarrow \mathbf{m}_i + \boldsymbol{\psi}_i \odot \Delta(\mathbf{m}_i, \{\mathbf{q}_t\}),
\label{eq:explicit-write}
\end{equation}
where $\Delta$ produces the candidate update for entry $i$, usually an aggregate of the queries that addressed it weighted by $w_{t,i}$, and the gate $\boldsymbol{\psi}_i$ controls how much of that update is admitted; controller-issued erase and add operations are the special case in which the model emits $\Delta$ and $\boldsymbol{\psi}_i$ itself. Explicit memory specializes \Cref{eq:generic-memory} by pairing append or gated write operations ($\mathcal{W}_\phi$, \Cref{eq:explicit-write}) with content-addressed read mechanisms ($\mathcal{R}_\psi$, \Cref{eq:explicit-read}). Unlike internal memory's identity read (\Cref{sec:parametric}), $\mathcal{R}_\psi$ performs selective retrieval over individually addressable entries $\mathbf{m}_i$, establishing its explicit structure. Persistence, the last axis of \Cref{sec:4axes}, refers to how long the store survives in memory. A verbatim store, for instance, holds only the entries built from the current input window. When the model moves on to the next sequence, those entries are discarded, and nothing carries over, so this memory is per-sequence.

Memory capacity is designed differently based on the store type. In slot and learned stores, the entries are parameters: $N_t$ is fixed before training, the contents of each $\mathbf{m}_i$ are set by gradient descent, and parameters stop changing when training stops. At inference, the read can therefore only return mixtures of patterns that training placed in the store. In verbatim, the entries are data, not parameters. Writing for these stores involves no changes in parameters: the trained encoder maps the new observation to a vector, and that vector is appended to the store as a fresh entry. Because encoding and appending are forward-pass operations, they apply to any input the model receives, including one it has never seen before. The store can thus hold new content, though how faithfully that content is represented depends on the encoder, which remains fixed. The read weights $w_{t,i}$ record how strongly the current window draws on each entry, which is what makes the read traceable, partially addressing the auditability problem raised in \Cref{sec:whymemory}. Methods in this family share the read of \Cref{eq:explicit-read}; they differ in what $\mathbf{m}_i$ contains and in how the write of \Cref{eq:explicit-write} is applied.

\begin{figure}[htbp]
\centering
\includegraphics[width=0.45\columnwidth]{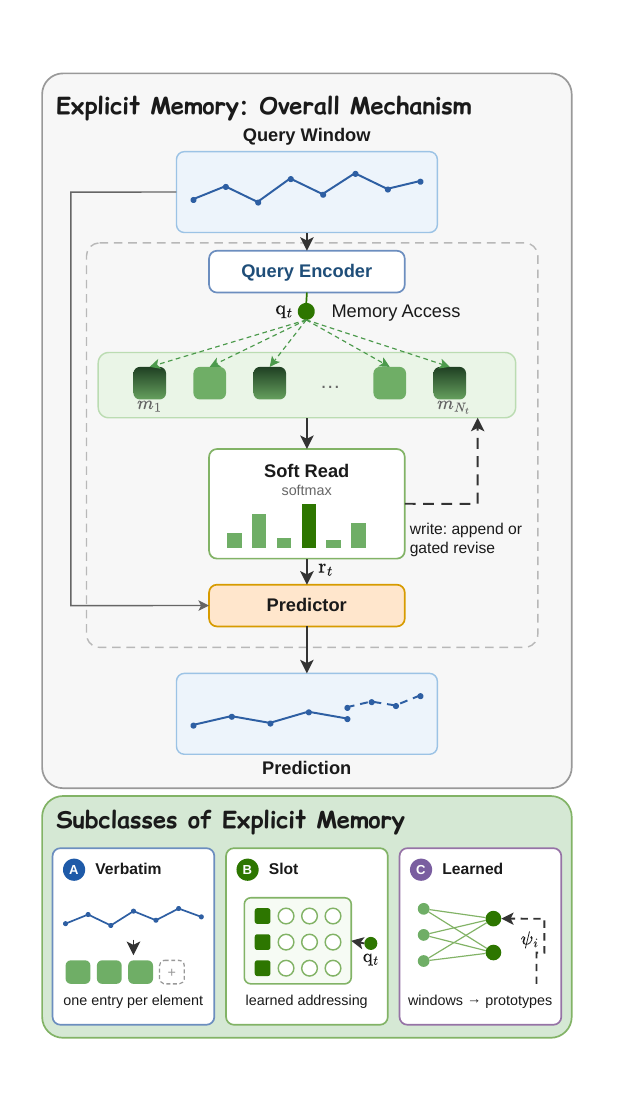}
\caption{Explicit memory for time series. \textbf{Top:} A query window is encoded as $\mathbf{q}_t$ and read against a store of individually addressable entries $\mathbf{m}_i$. The read operation weights entry by similarity, and the retrieved mixture $\mathbf{r}_t$ is integrated with the query window to produce the prediction. Writing takes one of two forms: appending a new entry, so that the store grows with the observed data, or revising existing entries through the gate $\boldsymbol{\psi}_i$, with capacity fixed in advance. \textbf{Bottom:} Illustration of the three explicit-memory forms considered in this section: verbatim, slot, and learned explicit memory.}
\label{fig:explicit}
\end{figure}

\subsection{Types of Explicit Memory}
\label{sec:explicit-types}

\subsubsection{Verbatim Memory}
\label{sec:explicit-raw}

Verbatim memory gives every observed timestep its own entry in the store. Entries are encoded on arrival, and the encoding is applied elementwise: no entry is merged with another or summarized into a coarser state, and forgetting occurs only when an entry is truncated from the store. A query is therefore scored against each retained element individually rather than against a learned summary of the past. The limiting case is the attention context of a Transformer~\cite{vaswani2017attention}, where the store is the input window itself, and every position remains addressable until the window advances past it.

Transformer-based forecasters instantiate this design directly. Within this literature, the word memory usually refers to verbatim retention. For example, LogTrans is introduced as breaking the memory bottleneck of self-attention on long series~\cite{li2019enhancing}, and the sparse or restructured reads of Informer~\cite{zhou2021informer}, Autoformer~\cite{wu2021autoformer}, and FEDformer~\cite{zhou2022fedformer} answer the same quadratic burden, thinning how the store is read while leaving its per-position entries intact. PatchTST~\cite{nie2023patchtst} coarsens the unit of storage from a timestep to a patch, shortening the store without consolidating across entries, and this patch-level verbatim store has become the default in time-series foundation models: TimesFM~\cite{das2023decoder} and Timer~\cite{liu2024timer} hold one key-value entry per patch in a decoder-only cache, Chronos~\cite{ansari2024chronos} quantizes values into a discrete vocabulary whose tokens are attended over individually, and Moirai~\cite{woo2024unified} and MOMENT~\cite{goswami2024moment} flatten patch sequences inside masked encoders. iTransformer~\cite{liu2024itransformer} marks the boundary of the category: it embeds the full history of each variate into a single token, so its store is verbatim across variates but consolidated over time.

Models that couple time series with language keep the same memory and extend what it already holds. Time-LLM~\cite{jin2024time} reprograms patch embeddings into the token space of a frozen language model, while ChatTS~\cite{xie2025chatts} and ITFormer~\cite{wang2025itformer} interleave encoded time-series tokens with text so that a single attention context serves as a joint store over both modalities. Verbatim memory keeps every element separately addressable, but this has a trade-off: the store grows linearly with the context length $L$, and attention over it grows quadratically. Because these models rely on a finite transformer context, their capacity is inherently architecturally bounded (see Section~\ref{sec:4axes}) limiting how far back in time they can look. As a result, much of this literature focuses on making the read operation cheaper, while leaving the contents of the store unchanged.

\subsubsection{Slot Memory}
\label{sec:explicit-slot}

Slot memory uses a fixed set of locations that hold model states rather than observations. The model learns how to read from and write to them. The formulation descends from the Neural Turing Machine (NTM) ~\cite{graves2014neural} and the Differentiable Neural Computer (DNC)~\cite{graves2016hybrid}, which pair a controller with an addressable matrix and learn the addressing itself. It also descends from the Variational Memory Encoder--Decoder~\cite{NEURIPS2018_e57c6b95}, which couples the store with a latent variable so that reads inject stochastic structure rather than a single deterministic vector. For temporal data, the design focuses not only on the addressing machinery, but also on what a memory slot is made to hold. PM-MemNet~\cite{lee2022learning} recasts forecasting as pattern matching: representative traffic patterns are clustered offline and stored as memory keys, and a query built from the input window retrieves the values that condition the prediction, allowing the model to respond to abrupt regime changes that a purely autoregressive encoder tends to smooth over. MegaCRN~\cite{jiang2023spatiotemporal} instead places a meta-node bank inside a graph convolutional recurrent encoder-decoder and uses the retrieved slots to generate the graph itself, so the memory parameterizes spatial structure rather than merely augmenting the hidden state. In anomaly detection, the mechanism is reversed: MemAE~\cite{gong2019memorizing} restricts the decoder to a sparse combination of memorized normal prototypes, so that the reconstruction error rises precisely for inputs that no slot can express. All of these designs fix the number of slots before training.

One of the central design questions in slot memory is about what to store to ensure reliable retrieval. The clearest example in time series is STanHop~\cite{wu2024stanhop}. It builds a forecaster from modern Hopfield layers, a content-addressable associative memory whose retrieval dynamics converge to a stored pattern, with optional external memory modules attachable at each resolution. The accompanying sparse Hopfield extension gives a tighter retrieval-error bound at no cost in capacity. STanHop~\cite{wu2024stanhop} is notable among the methods we review for explicitly evaluating store capacity and the reliability of pattern recovery.

\textit{Write} operations are less studied than reads in memory-based time-series models. Controller-issued, location-addressed updates in the NTM style have no counterpart on standard time-series benchmarks in the work we surveyed. Related works like STanHop populates its stores in advance, and the remaining methods collapse the write into the fixed or gated content-addressed update of \Cref{eq:explicit-write}. The nearest exceptions sit at the boundary of the field. Titans~\cite{behrouz2025titans} learns its write rule at test time and reports gains on forecasting benchmarks, but its store is a parametric fast-weight memory rather than an addressable slot matrix. In short, time-series models have learned how to read from slot memory. They decide where to look, and how much of what they find to change. However, they barely learned to write, lacking mechanisms to tell them what is worth storing, or where to put it.

\subsubsection{Learned Memory}
\label{sec:explicit-learned}

This memory stores compact representations learned from data, such as prototypes, centroids, or shared pattern banks. It is by far the dominant form in time series, and its logic inverts verbatim memory. Rather than keeping many concrete episodes and searching among them, the model keeps a few entries that summarize recurring structure, and explains a query as a mixture of them.

MEMTO~\cite{song2023memto} is the canonical time-series instance of this adaptation. It retains the reconstruction framing and the prototype bank but makes the update itself learned: a gate determines how intensively each entry absorbs new information, and $k$-means initialization stabilizes a two-phase procedure where entries are revised incrementally. Detection uses deviation in both input and latent space rather than reconstruction error alone. Recent work carries the gated bank into stronger backbones, with MOMEMTO~\cite{yoon2025momemto} attaching it to a patch-based foundation model and MemMambaAD~\cite{Li2025MemMambaADMS} pairing it with a state-space encoder.

A similar prototype-based logic applies to classification, in which the stored items serve as representatives of each class. TapNet~\cite{Zhang2020TapNetMT} learns an embedding together with a set of class prototypes and classifies by attentional distance. The bank is therefore the classifier, and a prediction is traceable to the representative it matched. Shapelet methods reach a similar structure by storing discriminative subsequences~\cite{li2021shapenet,tang2020interpretable}, and memory shapelet learning~\cite{10375830} maintains the store online for early classification of streaming series.

Forecasting relies on learned memory mostly when several series share the same structure. MegaCRN \cite{jiang2023spatiotemporal} attaches a Meta-Node Bank to a graph convolutional recurrent encoder-decoder. Node and time representations query the bank, and the retrieved entries generate the graph used for propagation. The bank therefore acts as an index of operating regimes rather than of episodes. Its reported disentanglement of road links and time slots with different behaviour is the clearest evidence that a bounded learned store can carry some non-stationary structure. MAGL~\cite{liu2022magl} records global historical features to mine spatial correlations, and HAMN~\cite{lee2025hamn} organizes the store by aggregation level, allowing sparse series to draw on their aggregates.

A more recent line of work uses learned memory to adapt frozen foundation models, and is notable because it argues explicitly against retrieval approaches in \Cref{sec:retrieval}. TS-Memory~\cite{lyu2026tsmemory} distills an offline nearest-neighbour teacher into a lightweight memory adapter, fused with the backbone at constant overhead so that no datastore is searched. MEMTS~\cite{yu2026memts} reaches a similar design from the same motivation. Overall, a bounded memory method is often chosen over an unbounded one not because it holds more, but because of its constant building cost.

Recent papers on memory also focus on imputation. MMNet~\cite{mmnet2025} pairs a missing-aware embedding with a memory-enhanced encoder, so a partially observed window is reconstructed from global similarity across the dataset rather than local context alone, and PRIME~\cite{yu2025prime} maintains a prototype memory of inter-series structure for irregularly sampled clinical series.

\subsection{Memory Writing and Integration}
\label{sec:explicit-write}

The \textit{read} operator reduces to \Cref{eq:explicit-read} in almost all of these methods. Yet there are major differences in how each method \textit{writes} with $\Delta$ and $\boldsymbol{\psi}_i$ in \Cref{eq:explicit-write}, and how the read result reaches the output. The most common approach is degenerate, in that the write is gradient update at training:
\begin{equation}
\begin{aligned}
    \Delta(\mathbf{m}_i, \{\mathbf{q}_t\}) &= -\nabla_{\mathbf{m}_i}\mathcal{L}, \\
    \boldsymbol{\psi}_i &=
    \begin{cases}
        \eta\,\mathbf{1} & \text{at training} \\
        \mathbf{0} & \text{at inference}
    \end{cases}
\end{aligned}
\label{eq:write-frozen}
\end{equation}
Here, memory entries are learned by gradient descent on the task loss $\mathcal{L}$ with learning rate $\eta$, then frozen. MemAE~\cite{gong2019memorizing}, TapNet~\cite{Zhang2020TapNetMT}, PM-MemNet~\cite{lee2022learning}, MegaCRN~\cite{jiang2023spatiotemporal}, STanHop's pre-populated modules~\cite{wu2024stanhop}, and the foundation-model adapters~\cite{lyu2026tsmemory,yu2026memts} all write this way. MNAD~\cite{park2020learning} makes the write input-dependent. Let $\mathcal{U}_i$ be the time steps that addressed entry $i$. Then:
\begin{equation}
    \Delta(\mathbf{m}_i, \{\mathbf{q}_t\}) = \sum_{t \in \mathcal{U}_i} w_{t,i}\,\mathbf{q}_t,
    \qquad
    \boldsymbol{\psi}_i = \mathbf{1}.
\label{eq:write-mnad}
\end{equation}
MEMTO~\cite{song2023memto} is the design in which the gate is learned. Let $\tilde{\mathbf{q}}_i$ be the aggregated queries addressing entry $i$, its update reads:
\begin{equation}
\begin{aligned}
    \mathbf{m}_i &\leftarrow (\mathbf{1}-\boldsymbol{\psi}_i) \odot \mathbf{m}_i + \boldsymbol{\psi}_i \odot \tilde{\mathbf{q}}_i, \\
    \boldsymbol{\psi}_i &= \sigma\!\big(\mathbf{W}[\,\mathbf{m}_i ; \tilde{\mathbf{q}}_i\,]\big),
\end{aligned}
\label{eq:write-memto}
\end{equation}
which is \Cref{eq:explicit-write} with $\Delta = \tilde{\mathbf{q}}_i - \mathbf{m}_i$. Here, $\sigma$ is the logistic function and $\mathbf{W}$ a learned projection, so each entry learns how much of the present it admits. MOMEMTO~\cite{yoon2025momemto} and MemMambaAD~\cite{Li2025MemMambaADMS} inherit this gate under stronger backbones, and online shapelet maintenance keeps the write active during deployment \cite{10375830}. 

The retrieved mixture then either \emph{replaces} the encoding before decoding, which is what forces reconstruction through stored content in the MemAE line, is \emph{fused} with the prediction as in MEMTO~\citep{song2023memto} and TS-Memory~\cite{lyu2026tsmemory}, or \emph{parameterizes} another component as when MegaCRN generates a graph. Replacing is the strongest option among the three. The decoder can only use what is in the store, so a poor reconstruction means no entry matched the input. This mechanism makes the error a useful anomaly score.

Two failure modes are specific to this class. \emph{Slot collapse} occurs when entries converge on similar content, leaving an effective capacity far below $N$; the compactness and separateness losses of MNAD~\cite{park2020learning} and the clustering-based initialization of MEMTO~\citep{song2023memto} aim to address this problem. \emph{Addressing diffusion} occurs when the weights $w_{t,i}$ spread across many entries, so that arbitrary inputs can be reconstructed from a mixture. Sparse addressing is a natural solution, and the sparse Hopfield analysis of STanHop characterizes its effect on retrieval error. Both modes indicate that in explicit memory, the diversity and sparsity of the store are part of the objective, not emergent properties of training.

\subsection{Coverage and Limitations}
\label{sec:explicit-positioning}

Explicit memory is concentrated in anomaly detection and classification. Anomaly detection is its strongest setting, because a bank of normality yields a detection criterion directly rather than as a by-product. Classification is the second, where prototype and shapelet stores align with matching a series against class representatives. Forecasting adoption is narrower than for other memory types, particularly retrieval memory. In addition, it is mostly cross-series or hierarchical, and imputation has only recently acquired dedicated instances.

The strengths and limitations of explicit memory follow directly from its position along our four memory axes. Because its capacity is architecturally bounded, the memory cost is known in advance, with no index to maintain and no latency that grows with the corpus. However, this means the size $N$ must be fixed before it is known how many distinct regimes exist in the data. The memory content is carefully checked, and the addressing weights account for each prediction. Its discrete representation ($\mathcal{M}_t \in \mathbb{R}^{N \times d_m}$) allows for interpretable read access: the memory content can be explicitly audited, and the addressing weights clearly account for each prediction. Yet, write access is notoriously unstable; joint training of reads and writes typically requires clustering initialization and diversity losses to prevent collapse. Furthermore, because persistence is typically per-dataset, memory entries lack temporal context; a prototype cannot be dated or deliberately retrieved when past conditions return. Second, unless the memory is specifically designed to update online, the store remains static after deployment and cannot learn patterns it never saw during training.

\begin{remark}
Explicit memory makes stored content addressable and auditable at a cost fixed by design, and its central design variable is not the read but the write, that is, whether and how selectively entries are revised. Its constraint is that capacity and content are both decided before deployment, so the store can index the regimes seen in training but cannot grow to accommodate new ones. Removing that constraint requires a store populated from data that expands independently of the model, which is the retrieval memory of \Cref{sec:retrieval}.
\end{remark}

\section{Retrieval-Augmented Memory}
\label{sec:retrieval}


\begin{framed}
\begin{definition}[Retrieval-augmented memory]
Retrieval-augmented memory refers to time-series systems that retain information in an external store and selectively recall a query-dependent subset to condition the current prediction. At its core, this paradigm asks: \emph{which information from a potentially large external memory should be recalled for the current query?} Rather than compressing all useful history into a bounded internal representation, retrieval memory preserves individually retrievable records whose capacity can grow independently of the predictor. Unlike agentic memory, however, the defining operation is selective recall rather than active management of what should be written, revised, or forgotten over time.
\end{definition}
\end{framed}

\begin{figure}[h]
    \centering
    \includegraphics[width=0.5\textwidth]{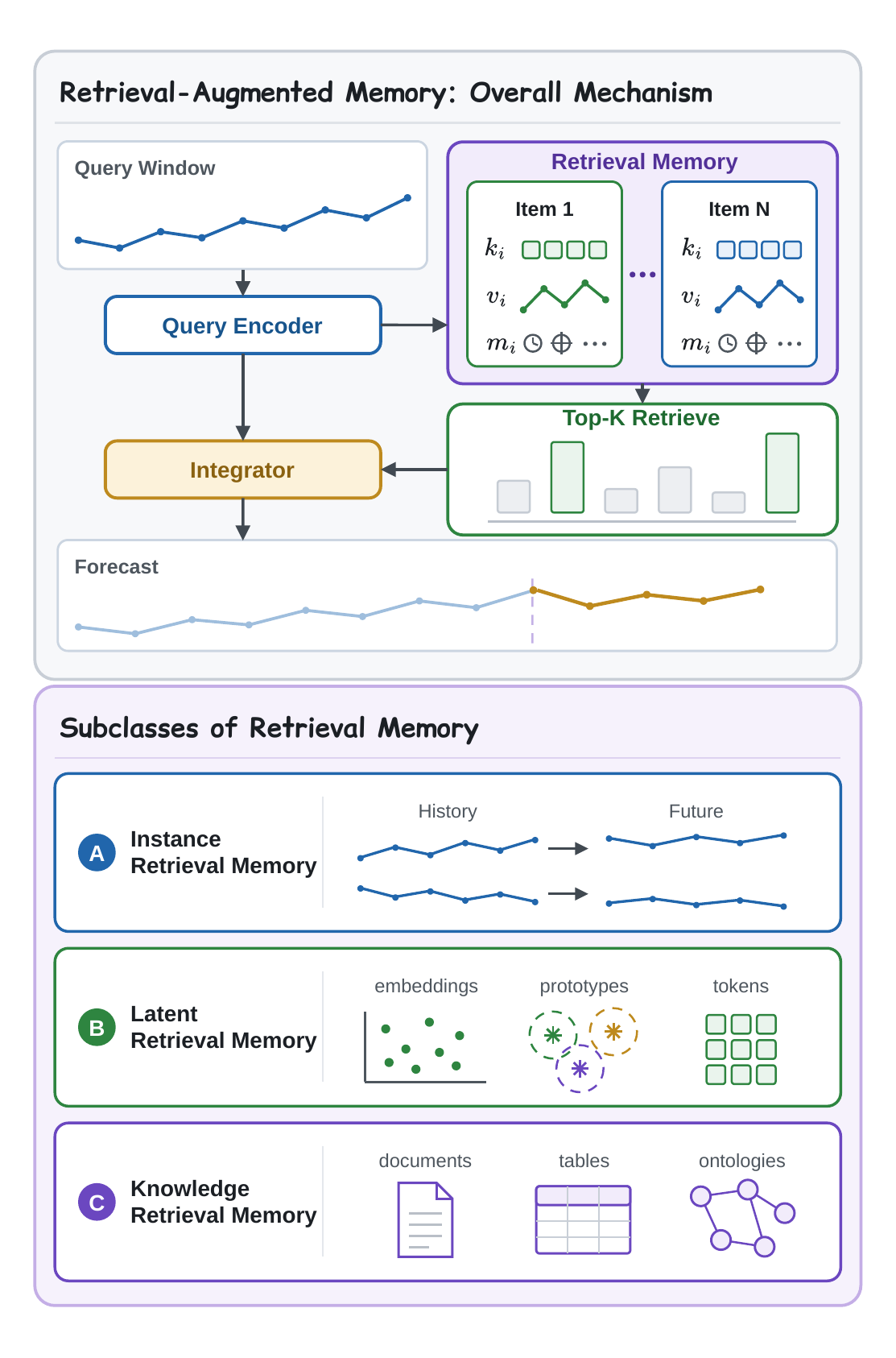}
    \caption{
Retrieval-augmented memory for time series. \textbf{Top:} A query window is encoded as $\mathbf{q}_t$ and matched against an external memory store whose entries contain a key $\mathbf{k}_i$, a value $\mathbf{v}_i$, and metadata $\mathbf{m}_i$. For the forecasting task, as an example, the top-$K$ relevant entries are retrieved, and their values are integrated with the query to produce the forecast. \textbf{Bottom:} Illustration of the three retrieval-memory forms considered in this section: instance, latent, and knowledge retrieval memory.
    }
    \label{fig:retrieval-memory}
\end{figure}

This shifts the memory problem from compressing all useful history into a bounded representation to determining what information to preserve externally and which records to recall for a particular query. An important design choice is therefore not only \emph{how} memory is retrieved, but \emph{what} is retained as memory. Time-series systems may retain concrete historical cases, compact latent representations, or external knowledge absent from the numerical series. We use this distinction as the primary organization of this section, categorizing retrieval-augmented memory into \emph{instance retrieval memory}, \emph{latent retrieval memory}, and \emph{knowledge retrieval memory}; these categories primarily describe memory \emph{representation}, while capacity, read/write, and persistence provide complementary characteristics. Figure~\ref{fig:retrieval-memory} summarizes this perspective. At a high level, a query window is encoded into a query representation $\mathbf{q}_t$ and matched against an external store containing keys $\mathbf{k}_i$, values $\mathbf{v}_i$, and optional metadata $\mathbf{m}_i$. The top-$K$ relevant entries are retrieved, and their values are provided to the predictor along with the query. 

\subsection{Mechanism}
\label{sec:retrieval-mechanism}

Let the external store be
\begin{equation}
    \mathcal{M}
    =
    \{(\mathbf{k}_i,\mathbf{v}_i,\mathbf{m}_i)\}_{i=1}^{N},
\end{equation}
where $\mathbf{k}_i$ is an indexable key, $\mathbf{v}_i$ is the stored memory content, and $\mathbf{m}_i$ denotes optional metadata such as timestamp, series identity, domain, sampling rate, or regime. For a query window $\mathbf{x}_{t-L+1:t}$, a query encoder produces $\mathbf{q}_t=g_{\phi}(\mathbf{x}_{t-L+1:t})$ and retrieves
\begin{equation}
    \begin{aligned}
        \mathcal{I}_t &= \operatorname{TopK}_{i}\, s(\mathbf{q}_t,\mathbf{k}_i;\mathbf{m}_i), 
        \\
        \widehat{\mathbf{y}} &= f_{\theta}\!\left(\mathbf{x}_{t-L+1:t}, \{\mathbf{v}_i\}_{i\in\mathcal{I}_t} \right).
    \end{aligned}
\label{eq:retrieval-memory}
\end{equation}
Here, $\operatorname{TopK}$ implements the read operator $\mathcal{R}$ via nearest-neighbor retrieval, and $s$ is the similarity measurement. We further characterize the retrieval-augmented memory by the \emph{store}, which determines the representation, capacity, and persistence of retained information; the \emph{retriever}, which implements the read operation; and the \emph{integration mechanism}, which determines how recalled information affects the output. These components need not operate over the same representation. A historical episode, for example, may be indexed by a learned or spectral embedding while the information returned to the predictor remains the original episode. Conversely, some systems retain only an encoded object and never return the underlying trajectory.

We therefore classify retrieval memory primarily by the content $\mathbf{v}_i$ that is persistently retained and made available for recall, rather than by the metric or backbone used to access it. Hybrid stores can combine more than one type, for example pairing a trajectory with textual context. The store may be built from the training set, observations accumulated before deployment, or a cross-series archive. In all cases, temporal validity is essential: keys must be computable from information available at query time, and any retrieved continuation or outcome must already have been observed to avoid leakage.

\subsection{Types of Retrieval Memory}
\label{sec:retrieval-types}

\subsubsection{Instance Retrieval Memory}
\label{sec:retrieval-instance}

This memory retains concrete historical cases such as windows, trajectories, context--future pairs, or examples from related series so that a current query can be grounded in specific precedents. In memory terms, these individually addressable cases form an \emph{explicit} store, whose capacity can scale with the amount of retained data. The store is typically maintained as a per-dataset index and accessed through similarity-based retrieval. This is the main form of retrieval memory in time-series forecasting and naturally targets recurring patterns, long-range dependence, and data-sparse series.


Early retrieval-based forecasting established the idea that historical cases can serve as external predictive evidence rather than merely as training examples. ReTime~\cite{jing2022retrievalbased} provides one of the earliest general formulations, motivated by the observation that forecasting uncertainty increases when only limited target history is available. It theoretically relates this uncertainty to prediction error and argues that conditioning on relevant reference series can reduce it. Operationally, ReTime separates \emph{relational retrieval}, which identifies related series using structural relations rather than relying only on potentially sparse query values, from \emph{content synthesis}, which combines the target and retrieved references to produce the prediction. Its extension to spatiotemporal forecasting and imputation further established retrieval as a reusable time-series mechanism rather than a forecasting-specific heuristic.

Subsequent work shifted from retrieving related series to retrieving concrete historical analogues from the forecasting corpus. RAFT~\cite{han2025raft} is a representative formulation: for a given query, it searches the training history for similar observed patterns and retrieves the corresponding realized futures, so that previously observed continuations become direct evidence for forecasting. This makes retrieval complementary to parametric learning: the predictor need not encode every recurring trajectory in its weights because relevant precedents can instead be supplied at inference time. Empirically, RAFT demonstrates that this relatively simple augmentation is effective across a broad set of forecasting benchmarks.

Other methods develop the same instance-memory principle through different retrieval and integration mechanisms. RATSF~\cite{wang2024ratsf} constructs a time-series knowledge base and introduces retrieval-augmented cross-attention to assimilate historical segments, targeting strongly non-stationary service-volume forecasting. RATD~\cite{liu2024ratd} transfers retrieval into probabilistic generation: an embedding-based retriever selects reference sequences, which then guide the denoising process of a diffusion forecaster rather than being directly fused with a deterministic prediction. These works therefore preserve the same memory object, such as historical examples, while changing how retrieved cases influence the predictor.

A second branch uses instance retrieval to adapt time-series foundation models without modifying the information stored in memory. RAF~\cite{tire2024raf} investigates retrieval specifically in the zero-shot TSFM setting and develops strategies for selecting related series and incorporating them into pretrained forecasters. TimeRAF~\cite{zhang2025timeraf} goes further by learning the retriever end-to-end and introducing channel prompting to integrate retrieved knowledge, while TS-RAG~\cite{ning2025tsrag} uses an Adaptive Retrieval Mixer to weight retrieved contexts and their associated future horizons before prediction. Across these methods, the progression is primarily in \emph{how retrieved instances are selected and fused}.

More specialized extensions broaden the same idea across predictors and domains. TimeRAG~\cite{yang2024timerag} converts retrieved sequences into context for an LLM-based forecaster, while RAST~\cite{ruan2026rast} adapts retrieval to traffic prediction through fine-grained spatiotemporal patterns and separate spatial--temporal query construction. Other work changes the access rule rather than the memory itself: SpecReTF~\cite{nguyen2026spectral} introduces frequency-aware retrieval with temporal recency, and Kang et al.'s CRAFT~\cite{ref58crm} extends spectral matching to multivariate channel structure. Instance retrieval has also begun to appear outside forecasting, including historical positive/negative cases for anomaly detection in LLMAD~\cite{liu2025llmad} and in-domain normal examples for test-time adaptation in RATFM~\cite{maru2025ratfm}.

\subsubsection{Latent Retrieval Memory}
\label{sec:retrieval-latent}

Latent retrieval memory retains compact, derived representations as first-class memory entries, such as embeddings, latent states, learned tokens, or priors. Although its contents are encoded, latent retrieval memory remains \emph{explicit} when the stored vectors or tokens are individually addressable; such banks can grow with available data, are typically read by embedding similarity, and are commonly retained across queries. This can reduce storage cost, make heterogeneous observations comparable in a shared space, and emphasize task-relevant structure that is difficult to express through a fixed distance over raw values.

ALER-TI~\cite{alerti2026} illustrates why retaining latent objects can be preferable to retrieving raw observations. In imputation, the incomplete query and a complete historical candidate are observed under different missingness patterns, so similarity measured directly in observation space can be poorly defined. ALER-TI addresses this mismatch through Latent Embedding Alignment, applying post-hoc masking in latent space so that complete historical embeddings can be pre-computed and cached while remaining comparable with corrupted queries. Retrieval therefore operates over a representation specifically constructed to preserve useful historical information under missingness rather than over the raw trajectory itself.

ReDiTT~\cite{reditt2026} provides a complementary motivation in asynchronous event sequences, where raw observations combine continuous inter-event times, discrete event types, variable sequence lengths, and padding. It encodes historical sequences into latent tokens, maintains these representations in a reference bank, and retrieves structurally similar latent sequences during both training and inference. The retrieved latent representations are then supplied through cross-attention to a conditional diffusion transformer, providing trajectory-specific guidance that stabilizes long-horizon generation. Here, the latent representation is not just an indexing key. Rather, it is itself the persistent object retrieved and consumed by the predictor, making ReDiTT a clear instance of latent retrieval memory under our definition.

GTR~\cite{cao2026enhancing} provides another example in multivariate time series forecasting, where a global temporal embedding represents long-range periodic patterns across the entire cycle. Given the current input and its position within the global cycle, GTR retrieves the corresponding latent segment and aligns it with the local multivariate sequence. Rather than retrieving a concrete historical trajectory, GTR retrieves a compact representation of global temporal structure, allowing long-range periodic information to be accessed without extending the input context.

The boundary with instance retrieval memory is determined by the role of the stored representation, not simply by the use of an encoder. If a method stores $\mathbf{k}_i=g(\mathbf{x}_i)$ only as a search key but returns the original historical case $(\mathbf{x}_i,\mathbf{y}_i)$, we regard it as \emph{instance retrieval memory with latent retrieval}. We reserve \emph{latent retrieval memory} for systems in which the encoded object itself is persistent and constitutes the principal recalled information. This distinction is particularly important for foundation-model retrieval, where learned encoders are often used for matching even though the retrieved value remains an identifiable historical segment. Latent retrieval memory gains compactness and abstraction, but can also discard rare details that later become predictive, making the representation objective part of the memory-design problem.

\subsubsection{Knowledge Retrieval Memory}
\label{sec:retrieval-knowledge}

Knowledge retrieval memory stores information that is not simply another realization of the target series, including text, semantic descriptions, exogenous events, metadata, and structured knowledge. These individually addressable knowledge records form an \emph{explicit} store that can grow with the available corpus, is read through semantic or structured retrieval, and is typically persistent across queries. Knowledge retrieval memory therefore provides context about the process rather than another trajectory to imitate.

TRACE~\cite{chen2025trace} represents a shift from retrieving numerically similar trajectories toward retrieving information grounded in the semantic context of a series. It learns a multimodal representation that aligns time-series channels with associated textual descriptions, using fine-grained channel-level alignment and hard-negative mining to make the embedding space semantically discriminative. The resulting retriever supports both time-series-to-text and text-to-time-series retrieval, allowing downstream models to access contextual evidence that numerical similarity alone may not recover. TRACE therefore contributes not only another retrieval metric, but also a mechanism for defining \emph{semantic relevance} across modalities.

SERAF~\cite{zhou2026seraf} addresses a related problem from a different direction. Rather than requiring an externally annotated text corpus, it derives textual descriptions of historical segments and performs two parallel retrievals: one according to numerical time-series similarity and another according to semantic similarity between the generated descriptions. Their retrieved futures are then selectively combined, allowing two histories that differ in local shape or scale but share higher-level characteristics such as trend or volatility to inform one another. This makes semantic retrieval complementary to instance-level numerical matching, particularly under non-stationarity where nearest trajectories are not necessarily the most informative precedents.

Other systems instantiate knowledge retrieval in domain-specific forms. Financial forecasting systems retrieve contemporaneous news or textual evidence alongside market histories, while healthcare models such as EMERGE~\cite{emerge2024} and REALM~\cite{zhu2024realm} connect longitudinal observations and clinical text to biomedical knowledge graphs and retrieve relevant entities or relations. These systems differ in knowledge source and downstream task, but share the same memory principle: the recalled value contributes information about the temporal process that is not represented simply as another historical trajectory.

Knowledge retrieval memory thus broadens retrieval from ``what happened in a similar historical case?'' to ``what external information is relevant to the current temporal state?'' It can also complement instance retrieval memory: a retrieved trajectory may be accompanied by text or metadata that helps determine whether the analogy is valid. The main difficulty is temporal and semantic alignment. Retrieved knowledge must match the correct entity and variable, but also remain valid at the relevant time and under the current regime; otherwise, semantically related but stale context can be as misleading as an outdated historical example.

\subsection{Memory Access and Integration}
\label{sec:retrieval-access}

Across the three memory types, reading is query-dependent, whereas writing typically consists of constructing or updating the external store and persistence is usually per-dataset. Instance memories are commonly searched with fixed similarities such as correlation, Euclidean distance, or dynamic time warping. Learned encoders replace hand-designed metrics with embedding similarity, while time-series-specific retrieval incorporates frequency, phase, recency, channel structure, calendar alignment, stationarity, or regime metadata. These choices define the \emph{read operation} over the store. In particular, a spectral or latent key should not be interpreted as latent retrieval memory when the returned value is still a concrete historical episode.

The central challenge is that \emph{similarity is not equivalent to predictive relevance}. The ideal retriever should identify the memory that most improves downstream inference, not merely the nearest item under a generic distance. This distinction becomes especially important under non-stationarity, where a highly similar example from an obsolete regime may be less useful than a moderately similar example from the current regime. Temporal metadata, structured similarity, and learned relevance functions provide different ways to address this mismatch.

Retrieved content is then incorporated through several recurring mechanisms. It may be appended to the input, fused through cross-attention or adaptive mixers, used as generative conditioning, or converted into prompt context. Input augmentation keeps retrieved evidence directly inspectable, whereas attention or latent fusion can select useful information more flexibly. Generative conditioning allows retrieved examples or priors to shape an entire predictive distribution, while prompting enables LLM-based systems to reason over retrieved evidence. These are \emph{access and integration strategies}, not additional memory types: the same instance store can support an MLP, diffusion model, TSFM, or LLM depending on how its entries are consumed.

\subsection{Coverage and Limitations}
\label{sec:retrieval-positioning}

Retrieval memory is heavily concentrated in forecasting, where instance retrieval memory dominates and increasingly interacts with foundation models, diffusion forecasters, and multimodal evidence. Outside forecasting, the literature is much thinner: retrieval-based anomaly detection has a strong recent example in LLMAD \citep{liu2025llmad}, while imputation, classification, reasoning, and decision-making remain relatively sparse. Latent and knowledge memories are particularly recent, suggesting that only a small subset of possible memory-type--task combinations has been explored. Section~\ref{sec:crosscutting} provides the task-centric comparison across memory architectures.

Across tasks, retrieval is characterized by explicit representations combined with a data-scaled capacity that grows seamlessly as the available corpus expands. It relies on query-dependent reads, limited or predominantly offline writes, and per-dataset persistence, while the external store can be updated without retraining the predictor. Instance retrieval memory additionally provides concrete precedents, latent retrieval memory offers compact reusable representations, and knowledge retrieval memory incorporates information unavailable in the numerical signal. Their corresponding limitations are retrieval and index-maintenance cost, dependence on the relevance function, loss of interpretability or detail in latent representations, and temporal/provenance issues for external knowledge. All three also provide little benefit when the store lacks an informative precedent. Finally, retrieval memory must be distinguished from the neighbouring classes in our taxonomy. Trainable prototype banks remain part of explicit memory (\Cref{sec:explicit}) when their capacity is architecturally bounded and learned jointly with the model, even if their read operation resembles retrieval. Conversely, the methods surveyed here largely use an unbounded external store with predefined or simply appended contents and focus on query-dependent reads of useful evidence. Once a model assumes active write access—using a controller or policy to decide what should be written, revised, promoted, or forgotten over time—the system shifts from data-scaled to policy-bounded capacity, moving beyond passive retrieval into the agentic memory considered in~\ref{sec:agentic}.

\begin{remark}
Retrieval-augmented memory is typically explicit in representation, unbounded in capacity, read through query-dependent retrieval, and persistent at the dataset level. Its strength lies in scalable access to concrete experiences, latent representations, or contextual knowledge without requiring all useful information to be compressed into model parameters. However, the store remains largely passive: retrieval determines what is recalled, but not what should continue to be remembered. Under non-stationarity, previously useful memories may become stale or misleading, motivating agentic memory mechanisms that actively govern writing, updating, and forgetting (Section~\ref{sec:agentic}).
\end{remark}
\begin{table*}[t]
\centering
\renewcommand{\arraystretch}{1.3}
\setlength{\tabcolsep}{5pt}
\caption{Functional memory types instantiated for time series. Decay refers to how quickly an entry loses validity under regime change.}
\label{tab:agentic-types}
\begin{tabularx}{\textwidth}{@{}>{\raggedright\arraybackslash}X >{\raggedright\arraybackslash}X l l l@{}}
\toprule
\textbf{Type} & \textbf{Stored content} & \textbf{Written} & \textbf{Read at} & \textbf{Decay} \\
\midrule
\rowcolor{rowfill}
\textbf{Episodic:} \emph{``What happened in similar cases before?''} & Context, forecast, and error triples & per instance & prediction, reflection & fast \\
\textbf{Semantic:} \emph{``What is generally true of this series?''} & Regimes, calendar rules, exogenous facts & once evidence accrues & prediction, scoring & medium \\
\rowcolor{rowfill}
\textbf{Procedural:} \emph{``How should I go about forecasting this?''} & Tool trajectories, model-selection playbooks & once evidence accrues & planning, action & slow \\
\bottomrule
\end{tabularx}
\end{table*}

\section{Agentic Memory}
\label{sec:agentic}

Sections~\ref{sec:explicit} and~\ref{sec:retrieval} described stores that a model \emph{reads}. In both cases, the memory store is essentially passive in one specific sense: the policy governing what is retained, retrieved, or revised is fixed before deployment. An explicit module may well write to its slots, but through a gate whose behaviour was set during training (Section~\ref{sec:explicit}-C), and a retrieval index is built once from a corpus and thereafter only queried.
Neither decides what is worth keeping in the first place, notices when a stored item has stopped being true, or removes it.
Agentic memory delegates those three decisions to a controller and remains an emerging, early-stage paradigm in time series.

For non-stationary time series, these decisions are crucial to ensure the quality of memory contents. Recalling last year's demand response to a promotion is wrong if the price regime has since changed, and the higher the similarity score, the more confidently the agent will be wrong. Static retrieval augmentation inherits this problem (Section~\ref{sec:retrieval}) because it can retrieve the most similar episode, but similarity alone cannot tell whether that episode remains valid.

Concurrent surveys cover the surrounding agent architecture, including reasoning and agentic systems in time series \cite{tsagentsurvey2026} and agent memory in general \cite{hu2026memoryagents}. We survey \emph{the store} rather than the whole agent, and we place it on the same spectrum as the explicit modules of Section~\ref{sec:explicit} and the retrieval indices of Section~\ref{sec:retrieval}, which makes the continuity across three classes visible and provides a lens complementary to those architecture-level accounts.

\subsection{Mechanism}
\label{sec:agentic-mechanism}
\begin{framed}
\begin{definition}[Agentic memory]
\label{def:agentic-mechanism}
An agentic memory is an external store $\mathcal{M}$ over which a controller $\pi$ governs three policies: a \emph{write} policy determining which artifacts of an interaction are persisted and in what abstracted form, a \emph{read} policy determining what is recalled and when, and a \emph{forget} policy determining what is revised, demoted, or discarded. The store evolves across interactions rather than being fixed per query.
\end{definition}
\end{framed}

\noindent Following the lifecycle formalization now standard in the agent-memory literature \cite{hu2026memoryagents}, an agent-–environment interaction at step $t$ emits artifacts $\phi_t$ that comprise a reasoning trace, the tools invoked, the agent's output, and, when available, feedback on the outcome.
This lifecycle directly specializes the generic memory formulation in Section~\ref{sec:prelim}, mapping generic observations to richer interaction artifacts ($\mathbf{o}_t \to \phi_t$) and factorizing the unified write operator $\mathcal{W}_\phi$ into distinct filtering/consolidation ($\mathcal{F}$) and encoding/structuring ($\mathcal{E}$) transformations.
Memory then evolves as
\begin{equation}
\begin{aligned}
\mathcal{M}_{t+1} &= \mathcal{E}\!\left(\mathcal{F}(\mathcal{M}_t,\phi_t)\right) \quad \Big[\equiv \mathcal{W}_\phi(\mathcal{M}_t, \phi_t)\Big], \\
r_t &= \mathcal{R}(\mathcal{M}_t, o_t, Q) \quad\quad\; \Big[\equiv \mathcal{R}_\psi(\mathcal{M}_t, o_t, Q)\Big].
\end{aligned}
\label{eq:agentic-lifecycle}
\end{equation}
where $\mathcal{F}$ forms memory candidates, $\mathcal{E}$ consolidates and prunes them, and $\mathcal{R}$ returns the signal on which the predictor conditions, with $o_t$ denoting the current observation or context derived from the input time series $\mathbf{x}_{1:t}$ and $Q$ the current task specification. 
All three operators are issued by the controller $\pi$ (Definition \ref{def:agentic-mechanism}) rather than fixed before deployment, which separates this class from Sections~\ref{sec:explicit} and~\ref{sec:retrieval}. For simplicity, we leave $\pi$ implicit in the notation.

Retrieval memory (Section~\ref{sec:retrieval}) can be viewed as the limiting case in which write-time filtering and post-deployment evolution are fixed or absent, so that only $\mathcal R$ remains active at inference. Internal memory (Section~\ref{sec:parametric}) instead is the case in which $\mathcal{F}$ is instantiated as the model's recurrent state-update function, while $\mathcal{E}$ and $\mathcal{R}$ require no separate memory operations.
In the agentic systems surveyed below, these operators are implemented through different mechanisms. For example,
\textsc{MemCast}~\cite{tao2026memcast} uses model-based extraction and reflection to form and refine memory entries, while \textsc{Argos}~\cite{gu2025argos} writes rules and revises them against feedback. For $\mathcal{R}$, \textsc{CastFlow}~\cite{pan2026castflow} selects by similarity, \textsc{MemCast} reweights by confidence, and \textsc{Cast-R1}~\cite{tao2026castr1} folds the retrieval decision into a learned policy.

\begin{figure}[t]
\centering
\includegraphics[width=0.7\columnwidth]{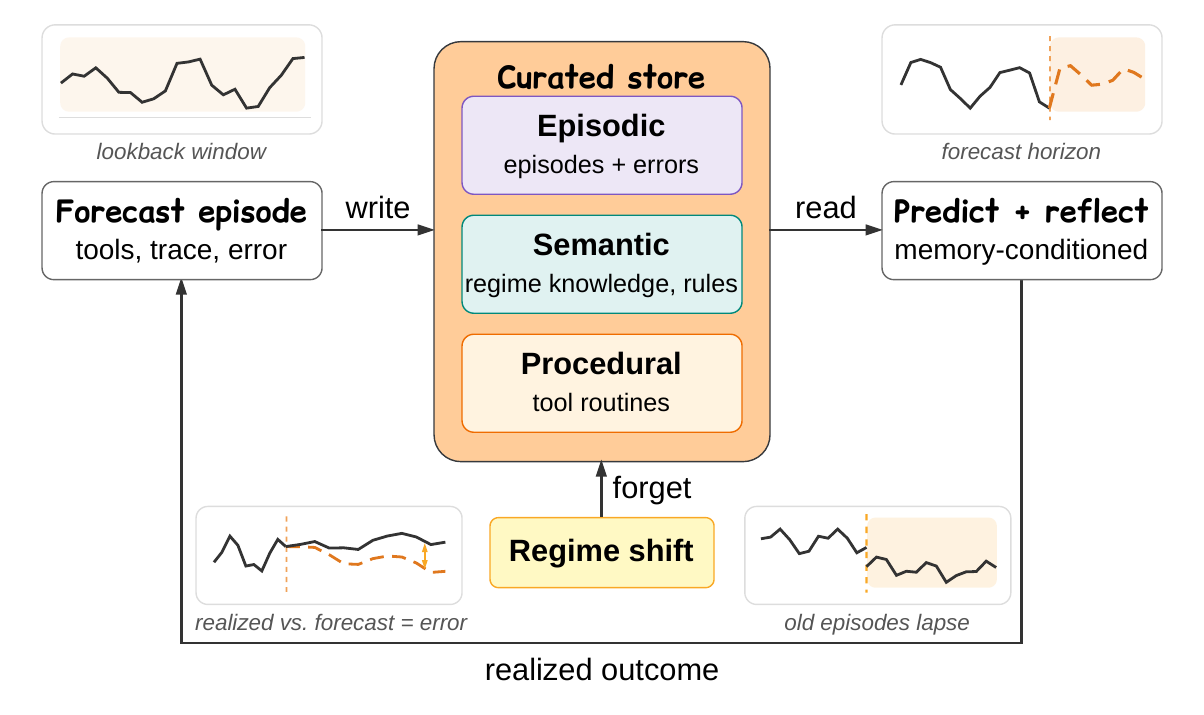}
\caption{The agentic memory loop for time series, illustrated on forecasting. A \emph{write} policy distills a forecasting episode into a store partitioned by memory type, a \emph{read} policy recalls entries to condition prediction and reflection, and the realized outcome is written back once the horizon elapses. The loop itself is task-agnostic; we use forecasting to make the \emph{forget} policy concrete because the horizon supplies a natural point at which an entry's validity can be checked. Beyond memory operations in Sections~\ref{sec:explicit} and~\ref{sec:retrieval}, an agentic \emph{forget} policy can demote entries whose validity has lapsed under regime change, which is the mechanism that non-stationarity necessitates.}
\label{fig:agentic-loop}
\end{figure}

Overall, agentic memory is \emph{explicit} in representation and, in principle, policy-bounded (refer to \Cref{sec:4axes}) in capacity, with the controller able to determine how much information is retained through its forgetting policy rather than through a fixed architectural limit. Read and write operations are likewise \emph{controller-issued}, allowing the memory contents and access patterns to adapt over time. In the fully online setting, the store persists across interactions and can be updated as new experience accumulates. However, as discussed below, most existing time-series systems do not yet operate in this fully online regime, instead constructing their stores offline and keeping them largely fixed at deployment. Figure~\ref{fig:agentic-loop} illustrates these components.


Classifying memory by function, as the agent literature does \cite{sumers2024cognitive,hu2026memoryagents}, reveals that most modules serve a small set of shared roles within a single store, despite differences in framework design and terminology. Table~\ref{tab:agentic-types} defines these three functional types by their target questions and stored contents, with Section~\ref{sec:agentic-design} expanding on the remaining columns.

\begin{table*}[http]
\centering
\setlength{\tabcolsep}{4pt}
\caption{Representative agentic memory systems for time series, organized by the store each maintains rather than by the agent architecture that surrounds it, which is what separates this account from agent-centric treatments \cite{cheng2026atsf,tsagentsurvey2026}.
\emph{Type}: E, S, P, and W denote episodic, semantic, procedural, and working (within-episode) memory. \emph{Scope}: per-sequence (resets when the interaction ends), per-dataset (distilled offline and frozen at test time), or online. \emph{Forget} indicates an explicit revision, demotion, or validity-decay policy. \emph{Task} names the canonical time-series tasks the store serves, where \emph{All four} denotes forecasting, anomaly detection, classification, and imputation.}
\label{tab:agentic-methods}
\begin{tabularx}{\textwidth}{@{}l l >{\raggedright\arraybackslash}X >{\raggedright\arraybackslash}X l l l@{}}
\toprule
\textbf{System} & \textbf{Type} & \textbf{Memory content} & \textbf{Write policy} & \textbf{Scope} & \textbf{Forget} & \textbf{Task} \\
\midrule
\rowcolor{rowfill}
MemCast \cite{tao2026memcast}   & E, S, P & patterns, wisdom, laws       & offline distillation + reflection & per-dataset  & conf.\ decay  & Forecasting \\
Cast-R1 \cite{tao2026castr1}    & W     & decision-relevant state      & RL-learned, multi-turn            & per-sequence & none          & Forecasting \\
\rowcolor{rowfill}
CastFlow \cite{pan2026castflow} & P     & tool-use trajectories        & sampled successful paths          & per-dataset  & none          & Forecasting \\
Nexus \cite{das2026nexus}       & E, S   & error signatures, guidelines & backtest calibration              & per-dataset  & none          & Forecasting \\
\rowcolor{rowfill}
AlphaCast \cite{zhang2025alphacast} & E, S & historical cases, domain knowledge, context
& context/case retrieval & per-sequence & none & Forecasting \\
Argos \cite{gu2025argos}        & S     & anomaly rules                & multi-agent generate and repair   & online       & rule revision & Anomaly det. \\
\rowcolor{rowfill}
AnomaMind \cite{tao2026anomamind} & S, P & anomaly patterns, domain knowledge & offline pattern mining       & per-dataset  & none          & Anomaly det. \\
Agentic-RAG \cite{ravuru2024agenticrag} & P & prompt pool of skills   & learned, hierarchical             & per-dataset  & none          & All four \\
\bottomrule
\end{tabularx}
\end{table*}

\subsection{Type of Agentic Memory}
\label{sec:agentic-methods}
A recent position paper recasts forecasting as an agentic process of perception, planning, action, reflection, and memory~\cite{cheng2026atsf}. The memory machinery itself is inherited from NLP agents, where verbal self-reflection \cite{shinn2023reflexion}, insight distillation \cite{zhao2024expel}, workflow abstraction~\cite{wang2024awm}, and agentic retrieval \cite{singh2025agenticrag} provide the write and read mechanisms that time-series systems now adopt.
Instead of treating memory as one module in an agent loop, we treat the memory itself as the object of study and ask what its write, read, and forget policies commit the system to. This shift allows an agentic memory to be compared with the explicit modules and retrieval indices on shared axes (Section~\ref{sec:explicit} and~\ref{sec:retrieval}). We organize the time-series agentic systems by the memory type they principally populate (Table~\ref{tab:agentic-methods}). 
Across these types, the same lifecycle operators are realized differently. The following sections will show how they form, evolve, and retrieve memory through model-based extraction, rule-based revision, similarity-based retrieval, and learned policies.

\subsubsection{Episodic and working stores.}
\textsc{MemCast} distills the training data into a hierarchical memory whose lowest tier summarizes prediction outcomes into \emph{historical patterns} and conditions inference on them, with a per-entry confidence score updated dynamically as the entry proves useful. This confidence adaptation lets the memory store evolve at inference time while explicitly avoiding test-data leakage, so the historical tier plays the role of an episodic memory that is refined over time. \textsc{Nexus} \cite{das2026nexus} maintains a domain-level calibration loop that scores past forecasts against ground truth across several historical splits and converts the resulting error signature into review guidelines for subsequent forecasts. Notably, a guideline is produced only if it improves accuracy beyond a fixed threshold on a held-out validation split, which makes the write itself selective based on actual performance.
\textsc{AlphaCast}~\cite{zhang2025alphacast} uses historical cases together with domain knowledge and contextual information to support iterative forecasting, but these resources are assembled for each forecasting interaction rather than maintained as a persistent, evolving store.

A further group operates one level down, retaining state only for the duration of a single analysis: \textsc{Cast-R1} carries decision-relevant evidence across the turns of a multi-step episode, \textsc{ReasonTSC} \cite{zhou2025enhancing} accumulates reasoning state before fusing it into a classification decision, and \textsc{TS-Reasoner} \cite{ye2026tsreasoner}, \textsc{TimeCopilot} \cite{garza2025timecopilot}, \textsc{TimeSeriesScientist} \cite{zhao2025tss}, and the multi-agent \textsc{ColaCare} \cite{wang2025colacare} retain intermediate findings and tool outputs within a single run. 
\textsc{Cast-R1} is a useful boundary case: its learned policy decides what evidence to retain across reasoning steps, but the resulting state remains confined to a single forecasting episode. It therefore illustrates agentic \emph{state management} without persistent memory across episodes.
This is working memory in the sense of \citet{hu2026memoryagents}. It is excluded from Table~\ref{tab:agentic-types} because it does not persist past the episode that created it, and so cannot decay under regime change.

\subsubsection{Semantic stores.}
While episodic stores retain experiences, semantic stores distill reusable knowledge, and they differ mainly in what that knowledge is asked to do. \textsc{MemCast} supports verification by inducing \emph{general laws} from extracted temporal features and applying them as criteria for reflective iteration. \textsc{AnomaMind} \cite{tao2026anomamind} instead supports grounding by providing anomaly-type descriptions and visual prototypes as external knowledge for its coarse-to-fine detection workflow. This knowledge is extracted offline from the training data and consulted alongside numerical diagnostic operators, so the store contributes contextual priors while evidence is gathered separately at inference. The core detection decisions are learned under verifiable feedback, so the semantic memory informs reasoning without serving as the final decision rule.

\textsc{Argos} \citep{gu2025argos} takes this idea a step further by turning stored knowledge into executable rules. These rules are generated, validated, and refined before deployment, making the repository both interpretable and directly testable. Each rule is implemented as a function that classifies a time window as normal or anomalous, so the stored knowledge is evaluated by running it on data.
Normality is captured through explicit anomaly \emph{rules} that collaborating agents write from observed patterns, repair when they throw syntax or runtime errors, and score on validation data before admitting them. A dedicated review agent checks each candidate against validation data and, when it detects performance regressions, derives revisions rather than discarding the rule outright, so the store is curated through an explicit generate-validate-repair loop. At deployment, the admitted rules run at low cost for online scoring and are combined with existing detectors, allowing the semantic store to serve a continuously monitored stream. Importantly, the rule repository remains inspectable, allowing operators to examine and modify the rules, which provides the auditability property discussed in Section~\ref{sec:whymemory}.

\subsubsection{Procedural stores.}
Beyond semantic knowledge, procedural stores capture how to act. \textsc{CastFlow} \citep{pan2026castflow} maintains a \emph{strategy memory} pairing each lookback window with its best-performing tool schedule. At planning time, the agent selects diagnostics by similarity to past windows, from change-point detection to spectral entropy, instead of scheduling tools zero-shot. The store is populated by expanding each training instance into several parallel exploration paths, scoring the resulting forecasts against ground truth, and archiving only the trajectory that performed best. Notably, the authors report a sweet spot in how many trajectories to recall: too few leave the agent under-guided, while too many inject redundant context that degrades the reasoning, which frames retrieval size as a design choice for a procedural store. Similarly, \textsc{MemCast}'s middle tier plays the same role under the name \emph{reasoning wisdom}, distilled from inference trajectories and used to select among candidate reasoning paths \citep{tao2026memcast}. 

\textsc{AnomaMind} \cite{tao2026anomamind} pairs its semantic store with a procedural component that governs how the coarse-to-fine detection workflow is applied, and \textsc{Agentic-RAG} \cite{ravuru2024agenticrag} organizes reusable skills as a hierarchical prompt pool. In the latter, a master agent routes a query to task-specific sub-agents, each drawing on its own prompt pool, and the procedural store is indexed by task. The memory content is constructed from successful runs. This success-only sourcing is also a limitation: the store records what worked, yet cannot capture the conditions under which a routine failed, so it cannot warn against reusing a strategy in a regime where it no longer applies.

\subsection{Memory Management and Integration}
\label{sec:agentic-design}
In this section, we discuss five dimensions that separate agentic memory systems, each of which raises an open question.

\textbf{What is written, and at what abstraction?} The formation operator $\mathcal F$ is implemented at different levels of abstraction. Raw episodes are cheap to write and transfer poorly, while abstracted laws and routines transfer but risk laundering a coincidence into a rule. \textsc{MemCast}'s three-tier design is comprehensive, keeping all three abstraction levels at once.

\textbf{How is memory integrated at inference?} The retrieval operator $\mathcal R$ determines how memory is integrated at inference. \textsc{CastFlow} reads at planning time and \textsc{AnomaMind} reads at every stage of a fixed coarse-to-fine workflow, letting the read point be built into the loop. \textsc{Argos} applies its whole rule repository to each window, discarding the need for a selective read mechanism. \textsc{MemCast} is the closest to a learned read, since a per-entry confidence score reweights what is recalled as entries prove useful, and Cast-R1 subsumes recall into an RL-learned decision policy. None of the systems we surveyed can operate without temporal context, so an uninformative store is queried at the same cost and shapes the output as much as an informative one.

\textbf{How are memory operations controlled?} Most systems use heuristic or prompt-level policies. \textsc{Cast-R1} instead learns the surrounding decision policy by supervised fine-tuning followed by multi-turn reinforcement learning, and \textsc{CastFlow} refines its workflow with verifiable rewards. The general trend in agent memory, which exposes store, retrieve, update, and discard as learned actions rather than fixed heuristics, has not yet been applied to memory operations for time series.

\textbf{When is the store allowed to change?} The evolution operator $\mathcal{E}$ governs when stored content is consolidated, revised, or persisted. With the partial exception of \textsc{Argos}, the memories surveyed here are \emph{distilled offline from the training split and frozen at inference}. \textsc{MemCast} adapts only per-entry confidence, and avoids leaking the test distribution, while \textsc{CastFlow} builds a case library per domain in advance, and \textsc{Nexus} calibrates on historical splits. Under our taxonomy, these systems are therefore closer to \emph{curated retrieval}, a store whose contents an agent chose but which is then read-only, than to the fully online memory the term ``agentic'' suggests. The true online regime, in which a deployed forecaster
writes and forgets as it goes, is largely underexplored.

\textbf{How should different memory types decay?} Once a store is allowed to change, the three types in Table~\ref{tab:agentic-types} should not be governed by a common rule. Procedural knowledge about which tool to use can stay relevant after a regime change, while specific past episodes become misleading the moment the regime shifts. A single forgetting rule across all three types therefore either discards routines that were still useful or keeps episodes that are already outdated. Our survey found limited evidence of type-differentiated decay, a gap Section~\ref{sec:challenges} develops further.

\subsection{Coverage and Limitations}
\label{sec:agentic-maturity}
As shown in the \emph{Scope} and \emph{Forget} columns of Table~\ref{tab:agentic-methods}, agentic memory mechanisms for time series are still emerging. Among surveyed methods, only one revises its store after deployment, while a second updates a confidence score without ever removing an entry. The rest build a store from the training split, freeze it, and read from it only. 
As Section~\ref{sec:agentic-design} noted, this is largely the curated retrieval regime. Most systems are close to Section~\ref{sec:retrieval}: an agent selects and structures the contents, but the store is still frozen at inference time. The literature has so far focused far more on \emph{write} than on \emph{forget}.
The same immaturity is also visible in task coverage. Agentic memory is concentrated almost entirely in forecasting, and the few anomaly detection systems populate semantic rather than episodic stores (\Cref{tab:taskmatrix}). This concentration leaves open whether the same memory mechanisms transfer to other time-series tasks.

Agentic memory is the class in this survey that most explicitly exposes memory revision as an operational mechanism, making it particularly suited to adaptation under drift. It accumulates experience without retraining, and its stores are inspectable enough to support audit. 
Against this, inference cost can be high and may grow with the size of the store, while write and forget policies are often hand-designed and rarely evaluated in isolation. Abstraction can also turn accidental correlations into spurious ``laws'', and an evolving store may amplify its own errors when incorrect predictions are written back as memory.
Moreover, no benchmark yet evaluates memory decisions rather than final forecast accuracy (Section~\ref{sec:resources}), leaving the benefit of curation over plain retrieval largely unmeasured.
\section{Cross-Cutting View: Memory by Task}
\label{sec:crosscutting}

The previous sections organized memory by the \emph{problem} it solves (\Cref{sec:whymemory}) and by the \emph{mechanism} that implements it (\Crefrange{sec:explicit}{sec:agentic}). This section focuses more on the question a practitioner actually starts from, which is not ``how does memory work?'' but ``I am building an anomaly detector, so what should I use?''. It reorganizes the same methods by \emph{task}, so the reader can look up a task and see which memory mechanisms have been tried and which method represents each.

The task view is very different from the problem view of \Cref{sec:whymemory}. A single task usually spans several problems at once; for instance, forecasting involves long-horizon dependence, recurring analogues, and drift together, while anomaly detection is built almost entirely around recognizing departures from stored normal patterns. Because of this, the task view exposes a different perspective. Instead of showing which \emph{problem} is under-served, it shows which \emph{application areas} have barely adopted a given class of memory at all. \Cref{tab:taskmatrix} places representative time-series memory methods into a task row and a mechanism column.

\begin{table*}[http]
\centering\small
\renewcommand{\arraystretch}{1.35}
\caption{Time-series memory methods by task and mechanism. The symbol $\varnothing$~\textit{open} marks task--mechanism combinations the reviewed literature has barely addressed.}
\label{tab:taskmatrix}
\setlength{\tabcolsep}{5pt}
\begin{tabularx}{\textwidth}{l >{\raggedright\arraybackslash}X >{\raggedright\arraybackslash}X >{\raggedright\arraybackslash}X >{\raggedright\arraybackslash}X}
\toprule
\textbf{Task}
  & \multicolumn{1}{c}{\color{cInternal}\textbf{Internal}}
  & \multicolumn{1}{c}{\color{cExplicit}\textbf{Explicit}}
  & \multicolumn{1}{c}{\color{cRetrieval}\textbf{Retrieval}}
  & \multicolumn{1}{c}{\color{cAgentic}\textbf{Agentic}} \\
\midrule
\rowcolor{rowfill}
\textbf{\faChartLine\ Forecasting}
  & DeepAR~\citep{salinas2020deepar}, DA-RNN~\citep{qin2017dual}, 
  TreNet~\citep{lin2017trenet}, LSTNet~\citep{lai2018modeling}, 
  SAnD~\citep{song2018attend}, Cloudlstm~\citep{zhang2021cloudlstm},
  P-sLSTM~\citep{kong2025unlocking}, FiLM~\citep{zhou2022film}, 
  TimeMachine~\citep{ahamed2024timemachine}, 
  MambaTS~\citep{cai2024mambats},
  ms-Mamba~\citep{karadag2026ms},
  Bi-Mamba+~\citep{liang2024bi},
  SST~\citep{patro2024sst}
  Attraos~\citep{hu2024attraos}
  & MTNet~\citep{chang2018memnet},  MAGL~\citep{liu2022magl},  PM-MemNet~\cite{lee2022learning}, MegaCRN~\citep{jiang2023spatiotemporal}, STanHop~\citep{wu2024stanhop},  HAMN~\citep{lee2025hamn}, MEMTS~\citep{yu2026memts}, TS-Memory~\citep{lyu2026tsmemory} 
  & RATD~\citep{liu2024ratd}, RAFT~\citep{han2025raft},  MQ-ReTCNN~\citep{yang2022mqretnn}, RATSF~\citep{wang2024ratsf},  RAF~\citep{tire2024raf}, TimeRAG~\citep{yang2024timerag}, RAST~\citep{ruan2026rast}, CRAFT~\citep{ref58crm}, ReDiTT~\citep{reditt2026}, SERAF~\citep{zhou2026seraf}, TRACE~\citep{chen2025trace}, ReTime~\citep{jing2022retrievalbased}, TimeRAF~\citep{zhang2025timeraf}, TS-RAG~\citep{ning2025tsrag},   SpecReTF~\citep{nguyen2026spectral}
  GTR~\citep{cao2026enhancing}
  & Cast-R1~\citep{tao2026castr1}, MemCast~\citep{tao2026memcast}, CastFlow~\citep{pan2026castflow}, Nexus~\citep{das2026nexus},   AlphaCast~\citep{zhang2025alphacast}, Agentic-RAG~\citep{ravuru2024agenticrag}, TimeSeriesScientist~\citep{zhao2025tss} \\
\textbf{\faTags\ Classification}
  & LSTM-FCN~\citep{karim2019multivariate},
  MambaSL~\citep{jung2026mambasl}
  & DPSN~\citep{tang2020interpretable}, ShapeNet~\citep{li2021shapenet}, TapNet~\citep{Zhang2020TapNetMT}, Memory Shapelet~\citep{10375830}
  & TRACE~\citep{chen2025trace}
  & Agentic-RAG~\citep{ravuru2024agenticrag}, ReasonTSC~\citep{zhou2025enhancing}, ColaCare~\citep{wang2025colacare}\\
\rowcolor{rowfill}
\textbf{\faExclamationTriangle\ Anomaly detection}
  & LSTM-AD~\citep{Malhotra2015LongST}, EncDec-AD~\citep{malhotra2016lstm}, MAAT~\citep{sellam2025mamba}
  & MEMTO~\citep{song2023memto}, MOMEMTO~\citep{yoon2025momemto}, MemMambaAD~\citep{Li2025MemMambaADMS}, Anomaly Trans.~\citep{xu2021anomaly}
  & LLMAD~\citep{liu2025llmad}, RATFM~\citep{maru2025ratfm}
  & Argos~\citep{gu2025argos}, AnomaMind~\citep{tao2026anomamind}, Agentic-RAG~\citep{ravuru2024agenticrag} \\
\textbf{\faPuzzlePiece\ Imputation}
  & GRU-D~\citep{che2018recurrent}, BRITS~\citep{3327757.3327783}
  & MMNet~\citep{mmnet2025}, PRIME~\citep{yu2025prime}
  & ALER-TI~\citep{alerti2026}
  & Agentic-RAG~\citep{ravuru2024agenticrag} \\
\rowcolor{rowfill}
\textbf{\faBrain\ Reasoning}
  & \open
  & ChatTS~\citep{xie2025chatts}, ITFormer~\citep{wang2025itformer}
  & EMERGE~\citep{emerge2024}, REALM~\citep{zhu2024realm}
  & TS-Reasoner~\citep{ye2026tsreasoner}, TimeCopilot~\citep{garza2025timecopilot}, TimeSeriesScientist~\citep{zhao2025tss} \\
\textbf{\faSitemap\ Decision-making}
  & LSTM-DRL~\citep{zou2024novel}
  & \open
  & \open
  & Cast-R1~\citep{tao2026castr1}, MemCast~\citep{tao2026memcast}, CastFlow~\citep{pan2026castflow} \\
\bottomrule
\end{tabularx}
\end{table*}
 
\msec{Forecasting.}
Forecasting predicts future values of a series from its past. It is the only task where all four memory types are already well represented, because forecasting benefits from several forms of history such as compressed temporal state for recent dynamics, explicit memory for recurring motifs or cross-series structure, and retrieval for rare or distant analogues. Retrieval is particularly strong here because a forecast often depends on a small number of relevant past episodes rather than the entire history~\citep{han2025raft,liu2024ratd}. Agentic memory is newer, but becomes useful when the model must decide what historical evidence to retain or revise as conditions change~\citep{tao2026memcast}.

\msec{Classification.}
Classification assigns a whole series to one of a set of classes. The task naturally favors explicit memory because class identity is often determined by matching a series against representative shapes, prototypes, or discriminative subsequences~\citep{tang2020interpretable,Zhang2020TapNetMT}. Retrieval, however, is almost absent. This suggests a possible gap, as retrieving similar labelled series could provide both stronger evidence and a direct explanation for the assigned class, especially when classes are rare or heterogeneous.

\msec{Anomaly detection.}
Anomaly detection flags inputs that depart from normal behaviour. Explicit memory fits this task particularly well because normal patterns can be stored as a reference set and anomalies identified by their failure to match that memory~\citep{song2023memto}. Retrieval provides a complementary view by comparing an input with relevant historical examples rather than a fixed learned prototype. Agentic memory is also promising under drift, where the definition of ``normal'' changes and stale memory must be updated rather than blindly reused.

\msec{Imputation.}
Imputation fills in missing values within a series from the observed context around them. Most existing methods still rely on internal and explicit memory to propagate information from nearby observations. Yet the task is naturally suited to retrieval where missing segments can often be reconstructed from analogous subsequences elsewhere in the same series or in related series, as explored by ALER-TI~\citep{alerti2026}. The relatively sparse literature suggests that memory has so far been used mainly to preserve local context rather than to exploit broader historical analogues.

\msec{Reasoning.}
This newer task asks a model to interpret a series and answer questions or explain what it shows, usually with a language model. Here, raw Transformer context already provides explicit memory over the observed series, but this alone limits reasoning to what is in the current window. Retrieval extends the evidence available to the model, while agentic memory supports multi-step analysis by retaining intermediate findings, tool outputs, and previous conclusions~\citep{ye2026tsreasoner,garza2025timecopilot}. The task therefore shifts the role of memory from preserving signal history toward preserving and organizing evidence for reasoning.

\msec{Decision-making.}
Here, a model acts on a series over time, taking decisions that serve two distinct purposes: executing an immediate action on the series, and generating outcomes that feed back into the system. Because the model must remember not only what happened, but also what it did and what followed, agentic memory is the natural candidate to accumulate these observations and revise later decisions~\citep{tao2026castr1,pan2026castflow}. Explicit and retrieval memory remain largely open, despite the potential value of recalling similar past decision episodes before acting.

Overall, \Cref{tab:taskmatrix} suggests that memory requirements are strongly task-dependent. Explicit memory is most useful when stored patterns directly support the task, as in classification and anomaly detection, while retrieval is strongest when relevant historical analogues matter, as in forecasting. Agentic memory is a relatively recent direction, but its general-purpose design allows it to support a broader range of downstream tasks. Existing approaches already span forecasting, anomaly detection, and other time-series tasks, although the literature remains concentrated primarily on forecasting. This broader flexibility stems from the ability of an agent to combine internal context, explicit stores, retrieval, and task-specific tools within a single workflow. In contrast, the other memory types provide more specific mechanisms for storing or accessing information, making them well suited to some tasks but less natural for others. The open cells in the table therefore highlight memory capabilities that remain underexplored for particular tasks.

\section{Benchmarks, Evaluation, and Resources}
\label{sec:resources}

We organize representative resources from conventional numerical benchmarks through foundation-model, multimodal, reasoning, and agentic settings. These families are not mutually exclusive; for example, a benchmark may be both multivariate and multimodal, but the organization exposes how evaluation requirements expand as models move outward along the memory spectrum. Table~\ref{tab:benchmark-landscape} summarizes these benchmark families and highlights how each stresses a different aspect of memory, from retaining temporal information and sharing cross-series structure to retrieving external context and maintaining information across multi-step interactions.

\begin{table*}[http]
\caption{Representative benchmark families relevant to time-series memory. Archival conference/journal papers are prioritized; $^{\dagger}$ denotes an emerging non-archival or workshop benchmark included because no mature archival equivalent currently exists.}
\label{tab:benchmark-landscape}
\centering
\footnotesize
\setlength{\tabcolsep}{4pt}
\begin{tabularx}{\textwidth}{p{0.13\textwidth} p{0.28\textwidth} X}
\hline
\textbf{Family} & \textbf{Representative resources} & \textbf{Primary evaluation role} \\
\hline
\rowcolor{rowfill}
General &
TFB~\cite{qiu2024tfb};
Monash Archive~\cite{godahewa2021monash};
UCR~\cite{dau2019ucr};
TSB-AD~\cite{liu2024tsbad} &
Standard forecasting, classification, and anomaly-detection evaluation across heterogeneous datasets. \\

Multivariate &
BasicTS+~\cite{shao2025basicts} &
Multivariate forecasting across diverse temporal and cross-variable dependencies; useful for testing shared and cross-variable memory. \\

\rowcolor{rowfill}
Spatiotemporal &
LargeST~\cite{liu2023largest} &
Large-scale traffic forecasting with long temporal coverage, many sensors, and metadata; useful for testing memory across many spatially related entities. \\

Hierarchical &
M5~\cite{makridakis2022m5} &
Forecasting across sparse product--store series and multiple aggregation levels. \\

\rowcolor{rowfill}
Foundation model &
ProbTS~\cite{zhang2024probts};
Chronos~\cite{ansari2024chronos};
BOOM~\cite{cohen2025boom} &
Broad-horizon, held-out/zero-shot, and large-scale evaluation of universal forecasters. \\

Multimodal &
Time-MMD~\cite{liu2024timemmd};
CiK~\cite{williams2025context};
Time-IMM~\cite{chang2025timeimm};
TRACE~\cite{chen2025trace} &
Time series paired with text, context, irregular modalities, or cross-modal retrieval targets. \\

\rowcolor{rowfill}
Reasoning &
TimeSeriesExamAgent~\cite{gwiazda2026timeseriesexam};
EngineMT-QA~\cite{wang2025itformer};
TSQA~\cite{kong2025timemqa};
ChatTS~\cite{xie2025chatts} &
Pattern understanding, question answering, explanation, and multi-step temporal reasoning. \\

Agentic &
MAFS~\cite{huang2025manyminds};
TimeSeriesGym$^{\dagger}$~\cite{li2025timeseriesgym} &
Agent cooperation and end-to-end time-series ML workflows; persistent-memory evaluation remains immature. \\

\rowcolor{rowfill}
Memory-specific &
SynTSBench~\cite{tan2025syntsbench};
TS-Haystack$^{\dagger}$~\cite{zumarraga2026tshaystack} &
Controlled temporal capability tests and long-context retrieval under increasing context length. \\
\hline
\end{tabularx}
\end{table*}

\subsection{Standard Time-Series Benchmarks}

\subsubsection{General Time Series}
Established task benchmarks remain necessary because a memory mechanism is useful only if it improves a well-defined downstream problem. TFB~\cite{qiu2024tfb} is the most comprehensive forecasting resource in this group. It is introduced to address three sources of unreliable comparison: limited domain coverage, biased model selection, and inconsistent evaluation pipelines, and evaluates both univariate and multivariate methods under a unified protocol across heterogeneous domains. For memory research, this breadth is useful for separating a mechanism that genuinely provides reusable historical information from one whose gains are specific to a small collection of frequently reused forecasting datasets.

Other archives provide complementary task coverage. The Monash Forecasting Archive~\cite{godahewa2021monash} exposes heterogeneous frequencies, lengths, and missing-value characteristics, while UCR~\cite{dau2019ucr} remains the standard archive for classification. Anomaly detection requires particular care because conclusions can change substantially with the dataset and metric. TSB-AD~\cite{liu2024tsbad} addresses this issue directly by curating 1,070 series from 40 datasets, benchmarking a broad range of detectors, and analysing evaluation-measure reliability rather than treating the metric as fixed. This is especially relevant to memory-based anomaly detectors, whose apparent gains from matching against stored normal patterns should not be conflated with biases in permissive evaluation metrics.

Imputation benchmarking is less consolidated. Recent archival work has emphasized that evaluation should reflect realistic MCAR, MAR, and NMAR missingness rather than only random deletion~\cite{toye2025imputation}. Such protocols are important for memory-based imputers because the usefulness of a stored prototype or retrieved analogue depends on how the missingness process alters the observable evidence; however, there is not yet a shared imputation benchmark designed specifically around memory mechanisms.

Existing results suggest that memory can help across these conventional tasks, but that its value depends on whether the store contains reusable evidence. Representative examples span several memory forms: RAFT~\cite{han2025raft} uses retrieved historical analogues for forecasting, MEMTO~\cite{song2023memto} uses prototype-style explicit memory for anomaly detection and classification. These results establish that memory is useful across multiple standard time-series tasks, but current evidence remains fragmented across task-specific benchmarks. A broader evaluation is still needed to determine whether particular memory mechanisms generalize across the heterogeneous regimes represented by TFB~\cite{qiu2024tfb}, and UCR~\cite{dau2019ucr}.

\subsubsection{Multivariate Time Series}
Multivariate benchmarks are particularly relevant to memory because predictive information may be distributed across variables as well as across time. A memory mechanism may therefore retain not only past states of individual channels, but also recurring multivariate configurations and dependencies that reappear across different temporal contexts.

BasicTS+~\cite{shao2025basicts} is the main benchmark for this setting. It is motivated by inconsistent conclusions across multivariate forecasting studies and standardizes training and evaluation while explicitly characterizing datasets by their temporal and cross-variable heterogeneity. Its large comparison across heterogeneous datasets shows why a single ``best'' multivariate architecture is difficult to identify: the usefulness of temporal modelling and cross-variable interaction depends strongly on the structure of the dataset. For memory-based models, this heterogeneity provides a natural stress test. Datasets with strong cross-variable structure can test whether shared or explicit memory captures reusable dependencies among channels, whereas more heterogeneous datasets test whether the model can selectively access useful channel relationships rather than indiscriminately sharing stored information. BasicTS+ is not itself a memory benchmark, but it provides the controlled diversity needed to determine when multivariate memory is actually useful.

Existing retrieval results on datasets covered by the BasicTS+ ecosystem make this heterogeneity concrete. RAFT~\cite{han2025raft}, Zhang et al.'s CRAFT~\cite{kang2026craft}, and SpecReTF~\cite{nguyen2026spectral} provide representative examples of instance, channel-aware, and frequency-aware retrieval on commonly used multivariate forecasting datasets. Across these approaches, memory gains are not uniform: retrieval is most useful when past multivariate configurations provide reusable future precedents, whereas the benefit shrinks or can become detrimental when similarity in the observed history does not imply similar future dynamics. The resulting insight is that multivariate memory should not be judged only by average accuracy; its value depends on the temporal and cross-variable structure that determines whether historical experience is actually reusable. A controlled comparison of these memory approaches under the exact BasicTS+ protocol remains missing.

\subsubsection{Spatiotemporal Time Series}
Spatiotemporal benchmarks extend the multivariate setting by attaching observations to a structured set of locations or nodes. This is particularly relevant to memory systems because useful historical information may come from another entity whose dynamics are spatially related to the query, making the size and structure of the shared memory part of the evaluation problem.

LargeST~\cite{liu2023largest} was introduced to move traffic forecasting beyond the relatively small California Performance Measurement System (PeMS) benchmarks~\cite{caltrans_pems}. It contains five years of measurements from up to 8,600 sensors together with sensor metadata, substantially increasing both the temporal archive and the number of entities that a model must handle. For memory-based forecasting, this scale tests whether a shared or retrieved store remains useful when candidate historical evidence is distributed across thousands of correlated streams, rather than only whether a model can exploit a small fixed graph. The commonly used PeMSD3, PeMSD4, PeMSD7, and PeMSD8~\cite{song2020spatial} datasets remain useful controlled baselines, while LargeST provides the stronger scalability test.

Current traffic results provide comparatively strong evidence for memory. PM-MemNet~\cite{lee2022learning}, MegaCRN~\cite{jiang2023spatiotemporal}, and MAGL~\cite{liu2022magl} provide representative explicit or learned-memory approaches that reuse recurring traffic patterns and shared spatial structure, while retrieval-based RAST~\cite{ruan2026rast} shows that recalling historical spatiotemporal precedents contributes beyond the forecasting backbone in controlled ablations. Taken together, these approaches indicate that traffic is a favorable regime for memory because useful patterns recur across both time and related sensors, and because cross-sensor structure provides additional reusable evidence. What remains less established is whether the same advantage persists when the searchable history expands to the multi-year, thousands-of-sensors scale represented by LargeST~\cite{liu2023largest}.

\subsubsection{Hierarchical Time Series}
Hierarchical benchmarks stress a different form of memory sharing: information can be reused across aggregation levels rather than only across neighbouring channels or nodes. The M5 Accuracy Competition~\cite{makridakis2022m5} is the canonical example, with 42,840 retail sales series organized across product, store, state, and aggregate levels. Many bottom-level series are sparse or intermittent, so their own local histories may provide little evidence for forecasting. This makes M5 a natural testbed for the P4 setting in Section~III: memory can be evaluated by whether information retained from related products, stores, or higher-level aggregates improves a data-sparse target rather than by simply extending that target's lookback window.

Current hierarchical-memory results support this form of cross-level sharing. HAMN~\cite{lee2025hamn}, for example, allows sparse lower-level series to draw on memory formed from richer aggregate series and improves forecasting across multiple hierarchical datasets. The main empirical insight is therefore that memory can compensate for weak local history when related levels contain reusable structure. However, this benefit has not yet been established at the substantially larger hierarchy and sparsity represented by M5 itself.

\subsection{Foundation-Model Benchmarks}
Foundation models change the evaluation question from performance on a fixed training distribution to generalization across datasets, frequencies, and domains. Chronos~\cite{ansari2024chronos} helped establish this setting by pretraining tokenized time-series models on a large mixture of public and synthetic data and evaluating them on a broad collection of datasets, including zero-shot targets not used for task-specific fitting. For memory research, this creates an important confound: an external memory may appear useful because it supplies new evidence, but some of that information may already be encoded parametrically during pretraining. Evaluation should therefore distinguish gains from retrieved or explicit memory from gains attributable to pretraining overlap.

Current foundation-model results indicate that parametric pretraining and additional memory are complementary rather than substitutes. Retrieval-based approaches such as TS-RAG~\cite{ning2025tsrag} and TimeRAF~\cite{zhang2025timeraf} show that query-specific historical evidence can still improve strong pretrained forecasters, while TS-Memory~\cite{lyu2026tsmemory} and MEMTS~\cite{yu2026memts} show a complementary direction in which reusable knowledge is internalized into compact memory modules to avoid datastore retrieval at inference. Together, these approaches suggest that broad pretraining does not eliminate the value of task- or domain-specific memory; instead, the useful memory form depends on the trade-off between adaptable external evidence and bounded inference cost. Their gains also make provenance increasingly important, since memory benefits can otherwise be confounded with information already encountered during pretraining.

\subsection{Multimodal Time-Series Benchmarks}
Multimodal benchmarks are the natural counterpart to the knowledge-memory setting of Section~\ref{sec:retrieval}, where useful evidence may be textual, contextual, or structured rather than another numerical trajectory. The key evaluation question is therefore stronger than whether an additional modality improves average accuracy: the benchmark should reveal whether the model can identify and use the specific external information relevant to the current temporal state.

Context-is-Key (CiK)~\cite{williams2025context} is particularly informative in this respect. Each forecasting instance pairs numerical history with carefully constructed textual context, and the tasks are designed so that the context is necessary for solving the instance correctly. This makes CiK closer to a memory stress test than a conventional multimodal dataset: a model that ignores the supplied contextual evidence should fail even if its numerical forecasting backbone is strong. For retrieval or knowledge-memory systems, it therefore provides a way to test whether recalled context changes the forecast in the direction implied by the evidence rather than merely being fused without effect.

TRACE~\cite{chen2025trace} targets the access side of knowledge memory more directly. It aligns time-series channels with textual descriptions in a shared representation and evaluates both text-to-series and series-to-text retrieval, so retrieval quality can be measured independently of a downstream forecaster. This is valuable for the decomposition in Section~\ref{sec:retrieval}: TRACE evaluates whether the correct cross-modal memory can be found before asking whether a predictor can use it.

Other resources broaden the setting. Time-MMD~\cite{liu2024timemmd} provides aligned numerical and textual series across nine domains together with MM-TSFlib for multimodal forecasting, while Time-IMM~\cite{chang2025timeimm} introduces asynchronous sampling, missingness, and multiple causes of irregularity. These extensions are useful for testing whether knowledge memory remains accessible when modality alignment is imperfect, or evidence arrives at different times.

Existing multimodal results suggest that memory is most useful when the recalled evidence contributes information that is genuinely absent from the numerical history. TRACE~\cite{chen2025trace}, SERAF~\cite{zhou2026seraf}, and EMERGE provide representative approaches that retrieve semantically aligned descriptions, combine numerical and semantic precedents, or ground temporal observations in external clinical knowledge. Across these examples, the gain comes not merely from adding another modality, but from retrieving complementary evidence that changes what can be inferred from the time series alone. The benchmark implication is therefore that multimodal memory should be judged by whether it finds temporally and semantically aligned evidence, while robustness to asynchronous, missing, or misaligned modalities remains much less characterized.

\subsection{Reasoning Benchmarks}
Reasoning benchmarks evaluate capabilities that conventional forecasting errors cannot express. TimeSeriesExamAgent~\cite{gwiazda2026timeseriesexam} is an important recent example because it treats benchmark construction itself as a scalable problem. It first defines controlled reasoning categories, including pattern recognition, noise understanding, similarity, anomaly detection, and causality, and then uses an agentic generation pipeline to extend these tests to real-world domains. From a memory perspective, these tasks expose whether a model can retain and compare evidence across separated observations, although the benchmark still scores the resulting reasoning answer rather than the underlying memory operations.

EngineMT-QA~\cite{wang2025itformer}, TSQA from Time-MQA \cite{kong2025timemqa}, and ChatTS~\cite{xie2025chatts} extend evaluation to temporal--textual question answering, numerical and open-ended questions, and multivariate reasoning. Collectively, they provide increasingly rich tasks in which intermediate evidence may need to remain available across several reasoning steps. What remains missing is an explicit intervention on memory itself, for example, controlling which earlier observation must be retained, when an intermediate conclusion should be stored, or whether the model can recover that evidence after a long reasoning trajectory.

Current results reveal an important distinction between long context and effective memory. ChatTS~\cite{xie2025chatts} represents explicit within-context memory, while TS-Reasoner~\cite{ye2026tsreasoner} and TimeCopilot~\cite{garza2025timecopilot} retain intermediate findings or tool outputs as working memory during multi-step analysis. Yet TS-Haystack~\cite{zumarraga2026tshaystack} shows that access to relevant evidence still deteriorates when informative events are buried within increasingly long temporal histories, whereas explicit retrieval remains substantially more reliable. Taken together, these approaches suggest that simply exposing a model to more history, or retaining intermediate reasoning state, does not guarantee that the required distant evidence remains recoverable. Most reasoning benchmarks still lack matched with-memory and without-memory variants, so the specific contribution of persistent memory remains difficult to isolate.

\subsection{Agentic Evaluation}
Agentic evaluation is substantially less mature because most existing systems are still judged by the quality of their final downstream prediction. MAFS~\cite{huang2025manyminds}, for example, evaluates a multi-agent forecasting system across multiple conventional forecasting datasets; this establishes end-task effectiveness but does not isolate whether persistent memory contributes to the result.

TimeSeriesGym~\cite{li2025timeseriesgym} moves evaluation closer to the agent level. Rather than restricting the task to producing a forecast, it evaluates agents on practical time-series machine-learning engineering challenges involving data handling, repository understanding, code, models, and multiple generated artifacts. This broader interaction horizon creates opportunities for an agent to reuse earlier findings and decisions, but persistent memory is still not an explicit evaluation target. Neither TimeSeriesGym~\cite{li2025timeseriesgym} nor current forecasting-agent benchmarks directly measure the central operations of Section~\ref{sec:agentic}: whether the controller writes a useful experience, updates it when evidence changes, recalls it in a later episode, or removes it when it becomes stale. The absence of an archival standardized benchmark for these operations is therefore itself a finding of this survey.

Memory-enabled forecasting agents such as MemCast~\cite{tao2026memcast}, CastFlow~\cite{pan2026castflow}, and Nexus~\cite{das2026nexus} provide representative episodic, procedural, and error-derived memory stores, and their end-task results suggest that reusing accumulated experience can improve agentic forecasting workflows. However, current agentic evaluations usually compare the complete system rather than an otherwise identical agent with and without persistent memory. MAFS~\cite{huang2025manyminds}, for example, shows clear gains from specialization and inter-agent information exchange, but does not isolate a persistent-memory effect. The current evidence therefore supports the effectiveness of memory-enabled agentic systems, but not yet the causal contribution of their write, recall, update, and forget operations.

\subsection{Memory-Specific Evaluation}
Standard task accuracy conflates the store, the access mechanism, and the predictor. A memory-specific benchmark should instead control what information is needed, where it is located, and how long it must remain recoverable.

TS-Haystack~\cite{zumarraga2026tshaystack} is the closest existing resource to this objective. It inserts known short events into longer sensor streams and evaluates direct retrieval, temporal reasoning, multi-step reasoning, and contextual anomaly tasks as context length increases. Because the location of the inserted event is controlled, performance can be interpreted as an ability to preserve and recover temporally localized information rather than only as aggregate task accuracy. Its reported comparison between compression and retrieval is particularly relevant to the memory spectrum: representations that remain sufficient for classification can still lose the local detail required to retrieve a specific event. TS-Haystack is currently a workshop benchmark, rather than a mature field standard.

SynTSBench~\cite{tan2025syntsbench} provides a complementary methodological precedent. Its programmable synthetic generation isolates temporal patterns, irregularities, and robustness factors, allowing model capabilities to be varied independently rather than inherited from a fixed real-world dataset. It is not designed specifically around memory, but the same controlled-generation principle could place informative events at known temporal distances or recurrence frequencies and thereby quantify how memory fidelity degrades.

Building on these directions, we recommend that time-series memory evaluation report at least four quantities separately from final task error. First, \emph{retention} should measure whether a known informative event remains recoverable as its temporal distance from the query increases. Second, \emph{retrieval quality} should measure whether the relevant record appears among the retrieved items, with random, no-retrieval, and, where possible, oracle controls. Third, \emph{temporal validity} should test whether a model rejects or down-weights memories from obsolete regimes. Fourth, \emph{memory management} should score write, update, and forget decisions over repeated episodes under a fixed storage budget. These measurements would turn memory from an architectural label into an independently testable capability and connect directly to the effective temporal memory capacity discussed in Section~XI.

\subsection{Open-Source Resources}
Benchmarking is also supported by mature software ecosystems. GluonTS provides reference implementations and evaluation tools for probabilistic forecasting and anomaly detection \cite{alexandrov2020gluonts}; aeon unifies forecasting, classification, regression, clustering, similarity search, and related time-series utilities \cite{middlehurst2024aeon}; and Merlion provides forecasting and anomaly-detection pipelines with deployment-oriented evaluation for univariate and multivariate series \cite{bhatnagar2023merlion}. More specialized resources such as TFB, BasicTS+, LargeST, MM-TSFlib, and TSB-AD supply task-specific datasets and reproducible protocols. Complementing these general-purpose resources, the accompanying collection for this survey is organized by the memory spectrum (explicit, retrieval, and agentic) and records the stored representation, read/write semantics, target task, venue, and available code for each method. 
\section{Open Challenges and Future Directions}
\label{sec:challenges}


The memory spectrum presented in this survey suggests that the central challenge in time series modelling is not simply how to access longer histories, but how to determine \emph{which parts of history should remain available, in what form, and at what cost}. Internal memory addresses long histories through compression, external memory through persistent storage and retrieval, and agentic memory further introduces explicit decisions about what to store, retrieve, update, and forget. We identify seven interconnected challenges that are likely to shape future time-series memory systems. These challenges can be summarized by seven guiding research questions, which provide an overview of the key issues discussed in the following subsections (Table~\ref{tab:future-rqs}).


\begin{table*}[http]
\centering
\small
\caption{Guiding research questions for future time-series memory systems.}
\label{tab:future-rqs}
\setlength{\tabcolsep}{4pt}
\begin{tabularx}{\textwidth}{p{0.13\textwidth} p{0.28\textwidth} X}
\toprule
\textbf{Section} & \textbf{Direction} & \textbf{Guiding research question} \\
\midrule

\rowcolor{rowfill}
\ref{sect: chall1}
& Memory Allocation and Effective Temporal Capacity
& \emph{How can architectures selectively retain the most informative time-series data, and how should the effective capacity of this recoverable information be quantified?} \\


\ref{sect: chall2}
& From Information Retrieval to Temporal Retrieval
& \emph{What makes a historical experience useful beyond surface-level morphological similarity?} \\

\rowcolor{rowfill}
\ref{sect: chall4}
& Memory Validity under Distribution Shift
& \emph{{How should time-series architectures continuously assess whether stored knowledge remains valid under changing conditions?}} \\

\ref{sect: chall5}
& Hierarchical Multimodal Memory
& \emph{{How should time-series architectures organize multimodal information across memory representations while preserving complementary information across modalities?}} \\

\rowcolor{rowfill}
\ref{sect: chall6}
& Beyond LLM-Centric Agentic Systems
& \emph{How should heterogeneous memory representations be coordinated?} \\

\ref{sect: chall7}
& Memory from Privileged Information
& \emph{Can privileged information during training shape memory under inference-time constraints?} \\

\rowcolor{rowfill}
\ref{sect: chall8}
& Causal Memory for Temporal Reasoning
& \emph{How can temporal memory preserve causal structure that enables models to simulate how a system evolved under different interventions?} \\


\bottomrule
\end{tabularx}
\end{table*}

\subsection{Memory Allocation and Effective Temporal Capacity}
\label{sect: chall1}
Early neural memory research established a fundamental distinction between compressing history into internal model representations and retaining information in an explicit, addressable external store~\cite{graves2014neural, weston2014memory, sukhbaatar2015end}. Complementary directions sought to extend usable history through temporal segment caching~\cite{dai2019transformer, rae2019compressive} or large-scale external retrieval~\cite{guu2020retrieval, borgeaud2022improving}. However, directly porting these capacity-scaling paradigms from natural language to time series exposes a fundamental misalignment. In NLP, information density is relatively uniform; in time series, it is radically asymmetrical. A sensor stream may contain millions of routine observations, yet only a handful of critical causal precursors (e.g., an anomalous vibration prior to failure). Consequently, two systems with identical memory budgets can retain vastly different predictive capabilities if they do not carefully manage \emph{what} they choose to remember. 

As highlighted by continual learning research, memory is not merely a storage problem, but a ``selection problem''~\cite{rolnick2019experience, parisi2019continual}. Maximizing raw storage capacity introduces redundancy, increases retrieval cost, and obscures critical experiences beneath routine noise—much like human cognition, which preferentially consolidates salient events over highly predictable background stimuli~\cite{labar2006cognitive, kensinger2020retrieval}. Because the value of a time-series observation may only become apparent months after it occurs, the goal of temporal memory is to determine the \emph{future predictive utility} of an event at the time of observation. This is especially relevant in long-horizon forecasting, where autoregressive rollout compounds errors step by step, so small early inaccuracies amplify over the horizon \citep{nguyenreviving}. Retaining the right distant precursors, not just recent forecasts, is crucial for preventing long-horizon predictions from drifting.

Resolving this requires architectures that actively counteract temporal smoothing by decoupling the memory write rate from the data sampling rate. Rather than uniformly compressing histories at fixed intervals, future time-series memory must become \emph{event-driven} and \emph{information-aware}. A promising solution is a two-track memory system. Routine, predictable patterns (like normal daily seasonality) can be heavily compressed into the model's standard hidden states. However, unexpected events, detected when the model makes a large prediction error or sees a sudden shift in the data, should be routed to a separate, permanent external memory. By triggering memory updates based on ``surprise'' rather than just the passage of time, the system can ensure that rare but critical clues are kept intact for when they are needed again.

To validate such systems, we argue for evaluating \emph{effective temporal memory capacity}, which measures how much temporally distant information a model can selectively retain, recover, and use to condition future predictions. Effective temporal memory capacity should be assessed by measuring how reliably task-relevant information remains recoverable as its temporal distance from the prediction target increases. While recent evaluation suites (discussed further in Section~\ref{sec:resources}) have made significant strides in stress-testing temporal capabilities—such as BasicTS+~\citep{shao2025basicts} and LargeST~\citep{liu2023largest} for scaling cross-variable and spatiotemporal dependencies, or SynTSBench~\citep{tan2025syntsbench} and TS-Haystack~\citep{zumarraga2026tshaystack} for controlled long-context retrieval—they still exhibit critical gaps. Specifically, these benchmarks largely evaluate retrieval from predefined, static context windows rather than testing a model's ability to autonomously manage an evolving memory constraint over unbounded streams. They do not explicitly evaluate the dynamic \emph{selection} problem. Future benchmarks must evolve to place informative causal precursors at highly variable temporal distances from the target window across continuous data streams. Measuring whether those specific events remain recoverable will provide a principled way to compare latent states, compressed memories, and persistent external stores, answering two core questions: \emph{First, by what mechanisms should an architecture selectively retain the most informative events from time-series histories? Second, how should we benchmark the effective capacity of the historical information that remains recoverable?}

\subsection{From information retrieval to temporal retrieval.}
\label{sect: chall2}
Retrieval-augmented time series forecasting has begun to move beyond simply extending the input context by allowing models to access relevant historical experiences. Initial efforts in retrieval-augmented forecasting, such as RATD~\cite{liu2024ratd} and RAFT~\cite{han2025raft}, demonstrated the utility of fetching external historical trajectories based primarily on time-domain similarity. However, as recently highlighted by~\citet{nguyen2026spectral}, anchoring retrieval strictly in the time domain exposes critical structural vulnerabilities. Firstly, time-domain distance metrics inherently ignore how energy is distributed across frequency bands, causing models to severely misidentify underlying periodic patterns when phase shifts occur~\cite{nguyen2026spectral}. Furthermore, these retrieval mechanisms often suffer from a lack of temporal recency weighting. All historical observations are treated equally, ignoring the reality that under non-stationary regime shifts, recent data often carries substantially stronger predictive power than distant matches. Table~\ref{tab:taskmatrix} shows that retrieval-based memory is particularly
well represented in forecasting, while comparatively fewer approaches have
explored retrieval for classification, anomaly detection, and imputation, and
retrieval-based memory remains largely unexplored for decision-making.

While incorporating spectral and recency constraints represents a necessary progression toward dynamics-aware retrieval, a deeper mechanical bottleneck remains. Retrieval paradigms, even when operating across both time and frequency domains, still rely fundamentally on \emph{morphological similarity} (e.g., Euclidean distance or cosine similarity in latent spaces). Morphological similarity is not necessarily equivalent to predictive relevance. Two historical sequences may exhibit virtually identical electrical demand profiles, yet possess completely divergent subsequent dynamics due to unobserved confounders (e.g., a standard summer day versus a latent heatwave). Conversely, a true historical causal precursor may exhibit high phase misalignment or amplitude distortion relative to the current window, causing it to be erroneously discarded by strictly symmetric matching criteria.

Beyond finding the right data, there is a bottleneck in how retrieved memories are actually integrated into the model. Current systems treat retrieval as an isolated step and lack a principled way to blend the retrieved historical data with current observations. Simply stitching a non-adjacent retrieved sequence next to the current one creates unnatural structural jumps at the boundaries, disrupting the continuous temporal manifold~\cite{nie2023patchtst, zhang2023crossformer}. On the other hand, standard attention provides a flexible mechanism for blending the two sources, but may also dilute distinctive temporal patterns when retrieved and current signals are combined indiscriminately. For retrieved memories, the broader challenge is therefore to integrate historical evidence without washing out the distinctive temporal patterns that made the memory useful in the first place.

Moving forward, time-series retrieval methods must transition from purely \emph{similarity-based matching} towards \emph{task- and dynamics-aware temporal retrieval}. A promising solution involves learning asymmetric similarity functions, where queries and keys are explicitly optimized to match current observations with historical precursors that led to similar \emph{future} outcomes, rather than just similar pasts.  Furthermore, to resolve the fusion bottleneck \citep{nguyenspectral}, these retrieved trajectories should be integrated via uncertainty-aware gating mechanisms that dynamically adjust for phase, scale, and the specific downstream task (e.g., forecasting vs. anomaly detection).  Ultimately, future architectures must move beyond surface-level similarity to answer a single core question: \emph{What makes a historical experience useful beyond surface-level morphological similarity?}

\subsection{Memory Validity under Distribution Shift}
\label{sect: chall4}
Most memory systems treat stored information as persistent evidence that can be reused whenever it appears relevant. However, as the underlying data-generating process evolves, the relationship between past observations and future outcomes may change. A historical pattern can therefore remain retrievable while becoming unreliable for the current process. The central challenge is consequently not simply to retrieve relevant history, but to determine whether that history remains trustworthy under the current conditions.

This problem is particularly acute in time series, where changes in consumption behaviour, infrastructure, operating conditions, or component ageing can alter the underlying dynamics. Historical observations may therefore remain useful, become outdated, or become useful again when similar conditions return. Recent time-series models increasingly address this problem through drift-aware adaptation, state disentanglement, and retention or retrieval of historical concepts~\cite{cai2025disentangling,chen2025learning,zhan2025continuous}. However, these approaches do not explicitly formulate memory maintenance as a continuous assessment of predictive validity that determines whether a stored pattern should remain active, be temporarily suppressed, reactivated, or permanently retired.

Addressing this challenge requires reconceptualizing memory as a dynamic system with mechanisms for continuous validation and controlled decay. Such a system must be condition-aware, distinguishing between \emph{permanent structural breaks} and \emph{transient regime shifts}, because a loss of reliability does not necessarily mean that a memory has become obsolete; it may instead reflect a temporary departure from the regime in which that memory is useful. If a process undergoes a fundamental shift, such as a factory replacing a core component, the memory system should progressively discount, replace, or retire memories associated with the obsolete regime. Conversely, if the shift is cyclical or transient, such as a seasonal change or temporary economic shock, the relevant memories should not be destroyed, but temporarily deactivated until similar conditions return.

Such condition-aware memory would complement temporal retrieval with a second mechanism for assessing whether retrieved knowledge remains reliable under the current regime. A memory would therefore not be reused solely because it resembles the present context; its continued use would depend on whether the knowledge it encodes remains consistent with ongoing observations and outcomes. This creates a memory lifecycle in which knowledge can remain active, become uncertain and temporarily suppressed, recover when its reliability is restored, or be retired when evidence indicates a persistent structural break. The challenge is therefore to move from passive storage to active memory maintenance, raising a critical question: \emph{How should time-series architectures continuously assess whether stored knowledge remains valid under changing conditions, and determine when memories should remain active?}

\subsection{Hierarchical Multimodal Memory}
\label{sect: chall5}
As discussed throughout this survey, memory mechanisms differ in how they represent, retain, and access historical information. Internal states provide efficient access to recent or recurring information, compressed representations extend temporal coverage at reduced fidelity, while external memory enables persistent retrieval of selected experiences. Recent time-series foundation models further broaden this design space by incorporating information beyond the numerical time series itself~\citep{das2023decoder,xie2025chatts,shi2025time}. For example, ChatTS maps time-series patches and text tokens into a shared embedding space through a lightweight projection layer before jointly processing them with an LLM~\citep{xie2025chatts}, while Time-VLM separately encodes temporal, visual, and textual information and then fuses these representations through cross-modal attention~\citep{zhong2025timeVLM}. These studies demonstrate that heterogeneous information can be jointly represented and accessed, but largely treat the representation and storage configuration as fixed design choices rather than a problem of adaptive memory allocation. The distribution in Table~\ref{tab:taskmatrix} similarly shows that existing
memory mechanisms are primarily distinguished by how information is stored and
accessed, while the organization of heterogeneous information across these
mechanisms remains largely unaddressed.

The challenge is to organize heterogeneous modalities across a hierarchy of memory representations. Different modalities provide complementary information at different temporal scales and levels of abstraction: continuous sensor streams may capture fine-grained dynamics that are efficiently represented in compact latent states, while textual or visual observations may preserve semantic information that is difficult to recover from the time series alone. A hierarchical memory could therefore maintain information at multiple levels, from shared low-dimensional representations for common information to modality-specific high-fidelity representations for complementary details. The key difficulty is determining when modalities should be jointly encoded, separately preserved, or promoted to richer representations while avoiding redundant storage and unnecessary memory cost.

An important research question is \emph{how should time-series architectures dynamically organize multimodal information across shared and modality-specific memory representations while preserving complementary information across modalities?} Such a framework would move beyond simply combining multimodal inputs toward learning how heterogeneous information should be represented at different levels of abstraction and when it should be shared or preserved separately.

\subsection{Beyond LLM-Centric Agentic Systems}
\label{sect: chall6}

Recent agentic time-series systems increasingly use LLMs as the central controller for reasoning, memory management, and tool use (see Section~\ref{sec:agentic}). These systems demonstrate that language models can organize historical information~\citep{tao2026memcast,tao2026anomamind}, generate summaries of complex workflows~\citep{zhang2025alphacast,zhao2025tss}, retrieve relevant experiences~\citep{ravuru2024agenticrag,tao2026memcast}, and coordinate multiple sources of information~\citep{das2026nexus}. Table \Cref{tab:taskmatrix} reveals a strong concentration of agentic memory
around LLM-centric architectures, particularly in forecasting, reasoning,
and decision-making. This concentration suggests that the emerging agentic
paradigm has largely inherited language-model-centric architectures rather
than developing memory and control mechanisms specialized for continuous
numerical dynamics. While effective for semantic reasoning and instruction following, positioning an LLM as the universal cognitive engine creates a fundamental architectural bottleneck when managing complex, continuous temporal dynamics.

In adjacent AI domains, early agent architectures were predominantly language-centric, but researchers quickly realized that forcing all perception through a textual bottleneck degrades performance. Consequently, recent systems natively incorporate visual and other modalities to capture complex information, a shift clearly demonstrated by the emergence of Vision-Language Models (VLMs). Time-series agents have reached a parallel inflection point. An LLM is primarily trained to model language, whereas time series data are numerical, continuous, high-dimensional, and governed by precise temporal relationships. Converting a long sensor stream into textual descriptions, or asking a language model to reason directly over numerical values, may therefore introduce information loss. At the same time, textual information can provide contextual information that is difficult to infer from the numerical signal alone.

A promising future direction is therefore to move towards heterogeneous multi-model agent architectures, where the agent's perception, memory, and control loop are co-governed by multiple specialized foundation models rather than centralized in a single LLM. Numerical encoders could preserve fine-grained temporal patterns, language models could represent semantic events and domain knowledge, and multi-modal models could connect temporal observations with text, images, and other contextual information. This leads to an open question of \emph{how should heterogeneous memory representations be coordinated so that language-based reasoning complements the numerical and temporal structure of the underlying data?}

\subsection{Memory from Privileged Information}
\label{sect: chall7}
Another promising direction for time-series memory is to use privileged information. Under the Learning Using Privileged Information (LUPI) paradigm, models exploit information available during training but unavailable at inference, often transferring this knowledge through teacher--student learning or generalized distillation~\cite{vapnik2015learning}. This setting is particularly natural for time series, where offline training provides richer temporal context than is available to an online model. For example, Karlsson et al.~\cite{karlsson2022privileged} study time-series privileged information in which intermediate observations between the prediction time and the target outcome are available during training but not at test time. Similarly, recent privileged knowledge distillation approaches have explored richer representations for multivariate time-series forecasting~\cite{liu2025timekd}. However, these methods primarily use privileged information to improve prediction, while its potential for directly supervising \emph{memory formation} remains largely unexplored.

The key challenge is that a memory mechanism must decide what to retain before knowing what will matter. A bounded-memory model must continuously determine which historical information to store, compress, or discard, even though the importance of an observation may only become apparent much later. Privileged temporal context provides a natural training signal for this problem. A teacher with access to the complete trajectory can identify which parts of the observed history were important for subsequent dynamics and transfer this knowledge to a causal student that operates only on information available at inference.

Such privileged memory distillation could supervise what a memory system retains, compresses, overwrites, and retrieves. A privileged teacher could use future observations, complete trajectories, additional sensor channels, delayed labels, or expert annotations to identify which historical information is most relevant, while a causal student learns to reproduce this memory using only information available at inference. This is particularly valuable in long-horizon and streaming settings, where limited memory capacity requires selective retention. The central question is therefore: \emph{Can privileged information during training shape memory under inference-time constraints?}

\subsection{Causal Memory for Temporal Reasoning}
\label{sect: chall8}

Internal memory offers an efficient means of compressing long temporal histories, but existing memory mechanisms are largely optimized for predictive accuracy rather than reasoning over the dynamics they encode. Meanwhile, causal time-series research has developed methods for identifying temporal dependencies and causal relationships~\cite{assaad2022survey,gong2025causal}, with recent work incorporating causal representations into forecasting~\cite{cai2025causal}. Yet causal structure is typically treated as a means to improve prediction, rather than as information that should be explicitly retained within memory for downstream reasoning. Consequently, a latent memory may capture recurring associations between events without preserving the mechanisms needed to explain why they occur or how they would respond to interventions.

Integrating causal structure into temporal memory could instead enable models to reason over past experience. A causal memory could retain temporal dependencies, delayed effects, regime-specific mechanisms, and intervention outcomes, allowing an agent to reason about the factors underlying an observed trajectory. This is particularly important for counterfactual reasoning, where the goal is not simply to predict what will happen, but to infer how a trajectory would have evolved under an alternative intervention. For example, rather than remembering only that increased load preceded a temperature rise, a causal memory could retain the relationship between load, temperature, and system state, enabling the model to reason about the consequences of reducing the load.

The key challenge is therefore to make causal structure a \emph{first-class component of temporal memory} that can be updated, retrieved, and reasoned over, rather than implicitly compressed into a predictive state. This leads to a fundamental question: \emph{How can temporal memory preserve and update causal mechanisms to support reasoning about why events occurred and how trajectories would change under alternative interventions?}

\section{Conclusion}
\label{sec:conclusion}


We recast a decade of time-series methods through a unified spectrum of memory, spanning implicit parametric state, explicit and compressed representations, persistent retrieval, and agent-curated stores. This perspective reveals a continuity often obscured by architecture-specific accounts, where different approaches vary not only in how much history they retain, but also in how historical information is represented, accessed, updated, and forgotten. It also makes several gaps in time-series memory research more visible. Long-term retrieval remains concentrated in forecasting, memory-specific evaluation is still limited, and agentic memory for time series remains at an early stage. More broadly, the survey suggests that the central challenge is shifting from simply extending temporal context towards learning what historical information is worth retaining, how it should be represented, where it should reside, and when it should still be trusted. We hope that the memory spectrum, taxonomy, and accompanying online repository
provide a durable framework for organizing this emerging literature and for situating future methods by the role that memory plays in temporal learning rather than by architectural era.

\FloatBarrier

\bibliographystyle{IEEEtranN}
\bibliography{references}

@article{elman1990finding,
  title={Finding structure in time},
  author={Elman, Jeffrey L},
  journal={Cognitive science},
  volume={14},
  number={2},
  pages={179--211},
  year={1990},
  publisher={Wiley Online Library}
}

@article{hochreiter1997lstm,
  title={Long short-term memory},
  author={Hochreiter, Sepp and Schmidhuber, J{\"u}rgen},
  journal={Neural Computation},
  volume={9}, number={8}, pages={1735--1780}, year={1997}
}

@inproceedings{cho2014learning,
  title={Learning phrase representations using RNN encoder--decoder for statistical machine translation},
  author={Cho, Kyunghyun and Van Merri{\"e}nboer, Bart and Gul{\c{c}}ehre, {\c{C}}a{\u{g}}lar and Bahdanau, Dzmitry and Bougares, Fethi and Schwenk, Holger and Bengio, Yoshua},
  booktitle={Proceedings of the 2014 conference on empirical methods in natural language processing (EMNLP)},
  pages={1724--1734},
  year={2014}
}

@inproceedings{vaswani2017attention,
  title={Attention is all you need},
  author={Vaswani, Ashish and Shazeer, Noam and Parmar, Niki and Uszkoreit, Jakob and Jones, Llion and Gomez, Aidan N and Kaiser, Lukasz and Polosukhin, Illia},
  booktitle={Advances in Neural Information Processing Systems (NeurIPS)},
  volume={30}, year={2017}
}

@inproceedings{wen2023transformers,
  title={Transformers in time series: A survey},
  author={Wen, Qingsong and Zhou, Tian and Zhang, Chaoli and Chen, Weiqi and Ma, Ziqing and Yan, Junchi and Sun, Liang},
  booktitle={International Joint Conference on Artificial Intelligence (IJCAI)},
  pages={6778--6786}, year={2023}
}

@article{lim2021dlsurvey,
  title={Time-series forecasting with deep learning: a survey},
  author={Lim, Bryan and Zohren, Stefan},
  journal={Philosophical Transactions of the Royal Society A},
  volume={379}, number={2194}, pages={20200209}, year={2021}
}

@inproceedings{liang2024fmtutorial,
  title={Foundation models for time series analysis: A tutorial and survey},
  author={Liang, Yuxuan and Wen, Haomin and Nie, Yuqi and Jiang, Yushan and Jin, Ming and Song, Dongjin and Pan, Shirui and Wen, Qingsong},
  booktitle={Proceedings of the 30th ACM SIGKDD Conference on Knowledge Discovery and Data Mining},
  pages={6555--6565}, year={2024}
}

@inproceedings{jin2024gnnsurvey,
  title={A survey on graph neural networks for time series: Forecasting, classification, imputation, and anomaly detection},
  author={Jin, Ming and Koh, Huan Yee and Wen, Qingsong and Zambon, Daniele and Alippi, Cesare and Webb, Geoffrey I and King, Irwin and Pan, Shirui},
  booktitle={IEEE Transactions on Pattern Analysis and Machine Intelligence (TPAMI)},
  year={2024}
}

@inproceedings{lai2018modeling,
  title={Modeling long-and short-term temporal patterns with deep neural networks},
  author={Lai, Guokun and Chang, Wei-Cheng and Yang, Yiming and Liu, Hanxiao},
  booktitle={The 41st international ACM SIGIR conference on research \& development in information retrieval},
  pages={95--104},
  year={2018}
}

@inproceedings{qin2017dual,
author = {Qin, Yao and Song, Dongjin and Cheng, Haifeng and Cheng, Wei and Jiang, Guofei and Cottrell, Garrison W.},
title = {A dual-stage attention-based recurrent neural network for time series prediction},
year = {2017},
publisher = {AAAI Press},
booktitle = {Proceedings of the 26th International Joint Conference on Artificial Intelligence},
pages = {2627–2633},
numpages = {7},
}

@inproceedings{kong2025unlocking,
  title={Unlocking the power of lstm for long term time series forecasting},
  author={Kong, Yaxuan and Wang, Zepu and Nie, Yuqi and Zhou, Tian and Zohren, Stefan and Liang, Yuxuan and Sun, Peng and Wen, Qingsong},
  booktitle={Proceedings of the AAAI Conference on Artificial Intelligence},
  volume={39},
  number={11},
  pages={11968--11976},
  year={2025}
}

@inproceedings{gu2022s4,
  title={Efficiently modeling long sequences with structured state spaces},
  author={Gu, Albert and Goel, Karan and R{\'e}, Christopher},
booktitle={Proceedings of the International Conference on Learning Representations (ICLR)}, year={2022}
}

@article{gu2023mamba,
  title={Mamba: Linear-time sequence modeling with selective state spaces},
  author={Gu, Albert and Dao, Tri},
  journal={arXiv preprint arXiv:2312.00752}, year={2023}
}

@inproceedings{zhou2022film,
  title={{FiLM}: Frequency improved {Legendre} memory model for long-term time series forecasting},
  author={Zhou, Tian and Ma, Ziqing and Wang, Xue and Wen, Qingsong and Sun, Liang and Yao, Tao and Yin, Wotao and Jin, Rong},
  booktitle={Advances in Neural Information Processing Systems (NeurIPS)}, year={2022}
}

@inproceedings{hu2024attraos,
  title={Attractor memory for long-term time series forecasting: A chaos perspective},
  author={Hu, Jiaxi and Hu, Yuehong and Chen, Wei and Jin, Ming and Pan, Shirui and Wen, Qingsong and Liang, Yuxuan},
  booktitle={Advances in Neural Information Processing Systems (NeurIPS)},
  year={2024}
}

@article{wang2025smamba,
  title={Is {Mamba} effective for time series forecasting?},
  author={Wang, Zihan and others},
  journal={Neurocomputing}, year={2025}
}

@article{karadag2026ms,
  title={ms-mamba: Multi-scale mamba for time-series forecasting},
  author={Karadag, Yusuf Meric and Talaz, Ismail and Dino, Ipek Gursel and Kalkan, Sinan},
  journal={Neurocomputing},
  pages={133226},
  year={2026},
  publisher={Elsevier}
}

@article{liang2024bi,
  title={Bi-mamba+: Bidirectional mamba for time series forecasting},
  author={Liang, Aobo and Jiang, Xingguo and Sun, Yan and Shi, Xiaohou and Li, Ke},
  journal={arXiv preprint arXiv:2404.15772},
  year={2024}
}

@article{jung2026mambasl,
  title={MambaSL: Exploring single-layer Mamba for time series classification},
  author={Jung, Yoo-Min and Kim, Leekyung},
  journal={arXiv preprint arXiv:2604.15174},
  year={2026}
}

@article{sellam2025mamba,
  title={Mamba Adaptive Anomaly Transformer with association discrepancy for time series},
  author={Sellam, Abdellah Zakaria and Benaissa, Ilyes and Taleb-Ahmed, Abdelmalik and Patrono, Luigi and Distante, Cosimo},
  journal={Engineering Applications of Artificial Intelligence},
  volume={160},
  pages={111685},
  year={2025},
  publisher={Elsevier}
}

@article{cai2024mambats,
  title={{MambaTS}: Improved selective state space models for long-term time series forecasting},
  author={Cai, Xiuding and others},
  journal={arXiv preprint arXiv:2405.16440}, year={2024}
}

@article{ahamed2024timemachine,
  title={{TimeMachine}: A time series is worth 4 mambas for long-term forecasting},
  author={Ahamed, Md Atik and Cheng, Qiang},
  journal={arXiv preprint arXiv:2403.09898}, year={2024}
}

@article{patro2024sst,
  title={{SST}: Multi-scale hybrid {Mamba}-{Transformer} experts for time series forecasting},
  author={Patro, Badri N and others},
  journal={arXiv preprint arXiv:2404.14757}, year={2024}
}

@article{salinas2020deepar,
  title={DeepAR: Probabilistic forecasting with autoregressive recurrent networks},
  author={Salinas, David and Flunkert, Valentin and Gasthaus, Jan and Januschowski, Tim},
  journal={International journal of forecasting},
  volume={36},
  number={3},
  pages={1181--1191},
  year={2020},
  publisher={Elsevier}
}

@inproceedings{lin2017trenet,
  author    = {Tao Lin and Tian Guo and Karl Aberer},
  title     = {Hybrid Neural Networks for Learning the Trend in Time Series},
  booktitle = {Proceedings of the Twenty-Sixth International Joint Conference on
               Artificial Intelligence, {IJCAI-17}},
  pages     = {2273--2279},
  year      = {2017},
  doi       = {10.24963/ijcai.2017/316},
  url       = {https://doi.org/10.24963/ijcai.2017/316},
}

@inproceedings{zhang2021cloudlstm,
  title={Cloudlstm: A recurrent neural model for spatiotemporal point-cloud stream forecasting},
  author={Zhang, Chaoyun and Fiore, Marco and Murray, Iain and Patras, Paul},
  booktitle={Proceedings of the AAAI Conference on Artificial Intelligence},
  volume={35},
  number={12},
  pages={10851--10858},
  year={2021}
}

@inproceedings{song2018attend,
  title={Attend and diagnose: Clinical time series analysis using attention models},
  author={Song, Huan and Rajan, Deepta and Thiagarajan, Jayaraman and Spanias, Andreas},
  booktitle={Proceedings of the AAAI Conference on Artificial Intelligence},
  volume={32},
  number={1},
  year={2018}
}

@inproceedings{liu2022magl,
  title={Memory augmented graph learning networks for multivariate time series forecasting},
  author={Liu, Xiangyue and Lyu, Xinqi and Zhang, Xiangchi and Gao, Jianliang and Chen, Jiamin},
  booktitle={Proceedings of the 31st ACM International Conference on Information \& Knowledge Management (CIKM)},
  pages={4254--4258}, year={2022}
}

@article{chang2018memnet,
  title={A memory-network based solution for multivariate time-series forecasting},
  author={Chang, Yen-Yu and others},
  journal={arXiv preprint arXiv:1809.02105}, year={2018}
}

@inproceedings{liu2024ratd,
  title={Retrieval-augmented diffusion models for time series forecasting},
  author={Liu, Jingwei and Yang, Ling and Li, Hongyan and Hong, Shenda},
  booktitle={Advances in Neural Information Processing Systems (NeurIPS)}, year={2024}
}

@inproceedings{shi2025time,
  title={Time-moe: Billion-scale time series foundation models with mixture of experts},
  author={Shi, Xiaoming and Wang, Shiyu and Nie, Yuqi and Li, Dianqi and Ye, Zhou and Wen, Qingsong and Jin, Ming},
booktitle={Proceedings of the International Conference on Learning Representations (ICLR)},   volume={2025},
  pages={34635--34667},
  year={2025}
}

@inproceedings{li2019enhancing, 
title={Enhancing the Locality and Breaking the Memory Bottleneck of Transformer on Time Series Forecasting}, 
author={Li, Shiyang and Jin, Xiaoyong and Xuan, Yao and Zhou, Xiyou and Chen, Wenhu and Wang, Yu-Xiang and Yan, Xifeng}, 
  booktitle = {Advances in Neural Information Processing Systems (NeurIPS)}, year={2019}}

@inproceedings{liu2024timer, 
title={Timer: Generative Pre-trained Transformers Are Large Time Series Models}, 
author={Liu, Yong and Zhang, Haoran and Li, Chenyu and Huang, Xiangdong and Wang, Jianmin and Long, Mingsheng}, 
booktitle={Proceedings of the International Conference on Machine Learning (ICML)}, 
year={2024}}

@inproceedings{jin2024time, 
title={Time-{LLM}: Time Series Forecasting by Reprogramming Large Language Models}, 
author={Jin, Ming and Wang, Shiyu and Ma, Lintao and Chu, Zhixuan and Zhang, James Y and Shi, Xiaoming and Chen, Pin-Yu and Liang, Yuxuan and Li, Yuan-Fang and Pan, Shirui and Wen, Qingsong}, 
booktitle={Proceedings of the International Conference on Learning Representations (ICLR)}, year={2024}}

@inproceedings{woo2024unified, 
title={Unified Training of Universal Time Series Forecasting Transformers}, author={Woo, Gerald and Liu, Chenghao and Kumar, Akshat and Xiong, Caiming and Savarese, Silvio and Sahoo, Doyen}, 
booktitle={Proceedings of the International Conference on Machine Learning (ICML)}, year={2024}}

@inproceedings{song2023memto, 
title={{MEMTO}: Memory-guided Transformer for Multivariate Time Series Anomaly Detection}, 
author={Song, Junho and Kim, Keonwoo and Oh, Jeonglyul and Cho, Sungzoon}, 
  booktitle = {Advances in Neural Information Processing Systems (NeurIPS)},
  year={2023}}

@article{behrouz2025titans,
  title   = {Titans: Learning to Memorize at Test Time},
  author  = {Behrouz, Ali and Zhong, Peilin and Mirrokni, Vahab},
  journal = {arXiv preprint arXiv:2501.00663},
  year    = {2025}
}

@article{garza2023timegpt,
  title={{TimeGPT-1}},
  author={Garza, Azul and Mergenthaler-Canseco, Max},
  journal={arXiv preprint arXiv:2310.03589}, year={2023}
}

@inproceedings{goswami2024moment,
  title={{MOMENT}: A family of open time-series foundation models},
  author={Goswami, Mononito and Szafer, Konrad and Choudhry, Arjun and Cai, Yifu and Li, Shuo and Dubrawski, Artur},
  booktitle={Proceedings of the International Conference on Machine Learning (ICML)}, year={2024}
}

@inproceedings{das2023decoder,
  title={A decoder-only foundation model for time-series forecasting},
  author={Das, Abhimanyu and Kong, Weihao and Sen, Rajat and Zhou, Yichen},
  booktitle={Proceedings of the International Conference on Machine Learning (ICML)}, year={2024}
}

@article{singh2025agenticrag,
  title={Agentic retrieval-augmented generation: A survey on agentic {RAG}},
  author={Singh, Aditi and Ehtesham, Abul and Kumar, Saket and Khoei, Tala Talaei},
  journal={arXiv preprint arXiv:2501.09136}, year={2025}
}

@inproceedings{lewis2020rag,
  title={Retrieval-augmented generation for knowledge-intensive {NLP} tasks},
  author={Lewis, Patrick and Perez, Ethan and Piktus, Aleksandra and others},
  booktitle={Advances in Neural Information Processing Systems (NeurIPS)}, year={2020}
}

@inproceedings{nie2023patchtst,
  title={A time series is worth 64 words: Long-term forecasting with transformers},
  author={Nie, Yuqi and Nguyen, Nam H and Sinthong, Phanwadee and Kalagnanam, Jayant},
booktitle={Proceedings of the International Conference on Learning Representations (ICLR)}, year={2023}
}

@inproceedings{zhou2022fedformer,
  title={Fedformer: Frequency enhanced decomposed transformer for long-term series forecasting},
  author={Zhou, Tian and Ma, Ziqing and Wen, Qingsong and Wang, Xue and Sun, Liang and Jin, Rong},
   booktitle={Proceedings of the International Conference on Machine Learning (ICML)},
  pages={27268--27286},
  year={2022},
  organization={PMLR}
}

@inproceedings{liu2024itransformer,
  title={{iTransformer}: Inverted transformers are effective for time series forecasting},
  author={Liu, Yong and Hu, Tengge and Zhang, Haoran and Wu, Haixu and Wang, Shiyu and Ma, Lintao and Long, Mingsheng},
booktitle={Proceedings of the International Conference on Learning Representations (ICLR)}, year={2024}
}

@inproceedings{zhang2023crossformer,
  title={Crossformer: Transformer utilizing cross-dimension dependency for multivariate time series forecasting},
  author={Zhang, Yunhao and Yan, Junchi},
booktitle={Proceedings of the International Conference on Learning Representations (ICLR)},   year={2023}
}

@inproceedings{wu2021autoformer,
  title={Autoformer: Decomposition transformers with auto-correlation for long-term series forecasting},
  author={Wu, Haixu and Xu, Jiehui and Wang, Jianmin and Long, Mingsheng},
  booktitle={Advances in Neural Information Processing Systems (NeurIPS)},
  volume={34}, pages={22419--22430}, year={2021}
}

@inproceedings{zhou2021informer,
  title={Informer: Beyond efficient transformer for long sequence time-series forecasting},
  author={Zhou, Haoyi and Zhang, Shanghang and Peng, Jieqi and Zhang, Shuai and Li, Jianxin and Xiong, Hui and Zhang, Wancai},
  booktitle={Proceedings of the AAAI Conference on Artificial Intelligence},
  volume={35},
  number={12},
  pages={11106--11115},
  year={2021}
}

@inproceedings{zeng2023dlinear,
  title={Are transformers effective for time series forecasting?},
  author={Zeng, Ailing and Chen, Muxi and Zhang, Lei and Xu, Qiang},
  booktitle={Proceedings of the AAAI Conference on Artificial Intelligence},
  volume={37}, pages={11121--11128}, year={2023}
}

@inproceedings{xu2021anomaly,
  title={Anomaly transformer: Time series anomaly detection with association discrepancy},
  author={Xu, Jiehui and Wu, Haixu and Wang, Jianmin and Long, Mingsheng},
booktitle={Proceedings of the International Conference on Learning Representations (ICLR)}, year={2021}
}

@article{cheng2026atsf,
  title   = {Position: Beyond Model-Centric Prediction---Agentic Time Series Forecasting},
  author  = {Cheng, Mingyue and Tao, Xiaoyu and Liu, Qi and Guo, Ze and Chen, Enhong},
  journal = {arXiv preprint arXiv:2602.01776},
  year    = {2026}
}

@article{tao2026memcast,
  title   = {{MemCast}: Memory-Driven Time Series Forecasting with Experience-Conditioned Reasoning},
  author  = {Tao, Xiaoyu and Cheng, Mingyue and Guo, Ze and Yu, Shuo and Liu, Yaguo and Liu, Qi and Wang, Shijin},
  journal = {arXiv preprint arXiv:2602.03164},
  year    = {2026}
}

@article{tao2026castr1,
  title   = {{Cast-R1}: Learning Tool-Augmented Sequential Decision Policies for Time Series Forecasting},
  author  = {Tao, Xiaoyu and Cheng, Mingyue and Jiang, Chuang and Gao, Tian and Zhang, Huanjian and Liu, Yaguo and Liu, Qi},
  journal = {arXiv preprint arXiv:2602.13802},
  year    = {2026}
}

@article{pan2026castflow,
  title   = {{CastFlow}: Learning Role-Specialized Agentic Workflows for Time Series Forecasting},
  author  = {Pan, Bokai and Cheng, Mingyue and Liu, Zhiding and Yu, Shuo and Tao, Xiaoyu and Wu, Yuchong and Liu, Qi and Lian, Defu and Chen, Enhong},
  journal = {arXiv preprint arXiv:2604.27840},
  year    = {2026}
}

@article{das2026nexus,
  title   = {{Nexus}: An Agentic Framework for Time Series Forecasting},
  author  = {Das, Sarkar Snigdha Sarathi and Goyal, Palash and Parmar, Mihir and Peng, Nanyun and Tirumalashetty, Vishy and Li, Chun-Liang and Zhang, Rui and Yoon, Jinsung and Pfister, Tomas},
  journal = {arXiv preprint arXiv:2605.14389},
  year    = {2026}
}

@article{zhang2025alphacast,
  title   = {{AlphaCast}: A Human Wisdom--LLM Intelligence Co-Reasoning Framework for Interactive Time Series Forecasting},
  author  = {Zhang, Xiaohan and Gao, Tian and Cheng, Mingyue and Pan, Bokai and Guo, Ze and Liu, Yaguo and Tao, Xiaoyu and Liu, Qi},
  journal = {arXiv preprint arXiv:2511.08947},
  year    = {2025}
}

@article{gu2025argos,
  title   = {{Argos}: Agentic Time-Series Anomaly Detection with Autonomous Rule Generation via Large Language Models},
  author  = {Gu, Yile and Xiong, Yifan and Mace, Jonathan and Jiang, Yuting and Hu, Yigong and Kasikci, Baris and Cheng, Peng},
  journal = {arXiv preprint arXiv:2501.14170},
  year    = {2025}
}

@inproceedings{
garza2025timecopilot,
title={TimeCopilot},
author={Azul Garza and Renee Rosillo Garcia},
booktitle={Recent Advances in Time Series Foundation Models Have We Reached the 'BERT Moment'?},
year={2025},
url={https://openreview.net/forum?id=A11xyQvSmT}
}

@article{zhao2025tss,                                   
  title   = {{TimeSeriesScientist}: A General-Purpose AI Agent for Time Series Analysis},
  author  = {Zhao, Haokun and Zhang, Xiang and Wei, Jiaqi and Xu, Yiwei and He, Yuting and Sun, Siqi and You, Chenyu},
  journal = {arXiv preprint arXiv:2510.01538},
  year    = {2025}
}

@article{ravuru2024agenticrag,                          
  title   = {Agentic Retrieval-Augmented Generation for Time Series Analysis},
  author  = {Ravuru, Chidaksh and Sakhinana, Sagar Srinivas and Runkana, Venkataramana},
  journal = {arXiv preprint arXiv:2408.14484},
  year    = {2024}
}

@article{hu2026memoryagents,
  title   = {Memory in the Age of {AI} Agents: A Survey of Forms, Functions and Dynamics},
  author  = {Hu, Yuyang and Liu, Shichun and Yue, Yanwei and Zhang, Guibin and others},
  journal = {arXiv preprint arXiv:2512.13564},
  year    = {2026}
}

@article{tsagentsurvey2026,                             
  title   = {A Survey of Reasoning and Agentic Systems in Time Series with Large Language Models},
  author  = {Chang, Ching and Shi, Yidan and Cao, Defu and Yang, Wei and Hwang, Jeehyun and Wang, Haixin and Pang, Jiacheng and Wang, Wei and Liu, Yan and Peng, Wen-Chih and others},
  journal = {Transactions on Machine Learning Research},
  year    = {2025}
}

@article{sumers2024cognitive,
  title   = {Cognitive Architectures for Language Agents},
  author  = {Sumers, Theodore R. and Yao, Shunyu and Narasimhan, Karthik and Griffiths, Thomas L.},
  journal = {Transactions on Machine Learning Research},
  year    = {2024}
}

@inproceedings{shinn2023reflexion,
  title     = {Reflexion: Language Agents with Verbal Reinforcement Learning},
  author    = {Shinn, Noah and Cassano, Federico and Gopinath, Ashwin and Narasimhan, Karthik and Yao, Shunyu},
  booktitle = {Advances in Neural Information Processing Systems (NeurIPS)},
  year      = {2023}
}

@inproceedings{zhao2024expel,
  title     = {{ExpeL}: {LLM} Agents Are Experiential Learners},
  author    = {Zhao, Andrew and Huang, Daniel and Xu, Quentin and Lin, Matthieu and Liu, Yong-Jin and Huang, Gao},
  booktitle = {Proceedings of the AAAI Conference on Artificial Intelligence},
  year      = {2024}
}

@article{wang2024awm,
  title   = {Agent Workflow Memory},
  author  = {Wang, Zora Zhiruo and Mao, Jiayuan and Fried, Daniel and Neubig, Graham},
  journal = {arXiv preprint arXiv:2409.07429},
  year    = {2024}
}

@article{tao2026anomamind,
    title={AnomaMind: Agentic Time Series Anomaly Detection with Tool-Augmented Reasoning},
    author={Tao, Xiaoyu and Wu, Yuchong and Cheng, Mingyue and Guo, Ze and Gao, Tian},
    journal={arXiv preprint arXiv:2602.13807},
    year={2026}
}

@inproceedings{gong2019memorizing,
  title={Memorizing normality to detect anomaly: Memory-augmented deep autoencoder for unsupervised anomaly detection},
  author={Gong, Dong and Liu, Lingqiao and Le, Vuong and Saha, Budhaditya and Mansour, Moussa Reda and Venkatesh, Svetha and Hengel, Anton van den},
  booktitle={Proceedings of the IEEE/CVF international conference on computer vision},
  pages={1705--1714},
  year={2019}
}

@article{graves2016hybrid,
  title={Hybrid computing using a neural network with dynamic external memory},
  author={Graves, Alex and Wayne, Greg and Reynolds, Malcolm and Harley, Tim and Danihelka, Ivo and Grabska-Barwi{\'n}ska, Agnieszka and Colmenarejo, Sergio G{\'o}mez and Grefenstette, Edward and Ramalho, Tiago and Agapiou, John and others},
  journal={Nature},
  volume={538},
  number={7626},
  pages={471--476},
  year={2016},
  publisher={Nature Publishing Group}
}

@inproceedings{NEURIPS2018_e57c6b95,
 author = {Le, Hung and Tran, Truyen and Nguyen, Thin and Venkatesh, Svetha},
  booktitle = {Advances in Neural Information Processing Systems (NeurIPS)},
  editor = {S. Bengio and H. Wallach and H. Larochelle and K. Grauman and N. Cesa-Bianchi and R. Garnett},
 pages = {},
 publisher = {Curran Associates, Inc.},
 title = {Variational Memory Encoder-Decoder},
 url = {https://proceedings.neurips.cc/paper_files/paper/2018/file/e57c6b956a6521b28495f2886ca0977a-Paper.pdf},
 volume = {31},
 year = {2018}
}

@inproceedings{
Khandelwal2020Generalization,
title={Generalization through Memorization: Nearest Neighbor Language Models},
author={Urvashi Khandelwal and Omer Levy and Dan Jurafsky and Luke Zettlemoyer and Mike Lewis},
booktitle={Proceedings of the International Conference on Learning Representations (ICLR)}, year={2020},
url={https://openreview.net/forum?id=HklBjCEKvH}
}

@inproceedings{guu2020retrieval,
  title={Retrieval augmented language model pre-training},
  author={Guu, Kelvin and Lee, Kenton and Tung, Zora and Pasupat, Panupong and Chang, Mingwei},
  booktitle={Proceedings of the International Conference on Machine Learning (ICML)},
  pages={3929--3938},
  year={2020},
  organization={PMLR}
}

@article{graves2014neural,
  title={Neural turing machines},
  author={Graves, Alex and Wayne, Greg and Danihelka, Ivo},
  journal={arXiv preprint arXiv:1410.5401},
  year={2014}
}

@inproceedings{sukhbaatar2015end,
  title={End-to-end memory networks},
  author={Sukhbaatar, Sainbayar and Weston, Jason and Fergus, Rob and others},
  booktitle = {Advances in Neural Information Processing Systems (NeurIPS)},  volume={28},
  year={2015}
}

@article{weston2014memory,
  title={Memory networks},
  author={Weston, Jason and Chopra, Sumit and Bordes, Antoine},
  journal={arXiv preprint arXiv:1410.3916},
  year={2014}
}

@inproceedings{dai2019transformer,
  title={Transformer-xl: Attentive language models beyond a fixed-length context},
  author={Dai, Zihang and Yang, Zhilin and Yang, Yiming and Carbonell, Jaime G and Le, Quoc and Salakhutdinov, Ruslan},
  booktitle={Proceedings of the 57th annual meeting of the association for computational linguistics},
  pages={2978--2988},
  year={2019}
}

@inproceedings{rae2019compressive,
  title={Compressive transformers for long-range sequence modelling},
  author={Rae, Jack W and Potapenko, Anna and Jayakumar, Siddhant M and Lillicrap, Timothy P},
booktitle={Proceedings of the International Conference on Learning Representations (ICLR)}, year={2019}
}

@inproceedings{borgeaud2022improving,
  title={Improving language models by retrieving from trillions of tokens},
  author={Borgeaud, Sebastian and Mensch, Arthur and Hoffmann, Jordan and Cai, Trevor and Rutherford, Eliza and Millican, Katie and Van Den Driessche, George Bm and Lespiau, Jean-Baptiste and Damoc, Bogdan and Clark, Aidan and others},
  booktitle={Proceedings of the International Conference on Machine Learning (ICML)},
  pages={2206--2240},
  year={2022},
  organization={PMLR}
}

@article{nguyen2026spectral,
  title={Spectral Retrieval-Augmented Time-Series Forecasting},
  author={Nguyen, Huu Hiep and Nguyen, Minh Hoang and Nguyen, Dung and Le, Hung},
  journal={arXiv preprint arXiv:2606.19412},
  year={2026}
}

@inproceedings{rolnick2019experience,
  title={Experience replay for continual learning},
  author={Rolnick, David and Ahuja, Arun and Schwarz, Jonathan and Lillicrap, Timothy and Wayne, Gregory},
  booktitle = {Advances in Neural Information Processing Systems (NeurIPS)},  volume={32},
  year={2019}
}

@article{parisi2019continual,
  title={Continual lifelong learning with neural networks: A review},
  author={Parisi, German I and Kemker, Ronald and Part, Jose L and Kanan, Christopher and Wermter, Stefan},
  journal={Neural networks},
  volume={113},
  pages={54--71},
  year={2019},
  publisher={Elsevier}
}

@inproceedings{yang2022mqretnn,
  author    = {Yang, Sitan and Eisenach, Carson and Madeka, Dhruv},
  title     = {{MQ-ReTCNN}: Multi-Horizon Time Series Forecasting with Retrieval-Augmentation},
  booktitle = {KDD 2022 Workshop on Mining and Learning from Time Series -- Deep Forecasting: Models, Interpretability, and Applications},
  year      = {2022},
}

@inproceedings{jing2022retrievalbased,
  author    = {Jing, Baoyu and Zhang, Si and Zhu, Yada and Peng, Bin and Guan, Kaiyu and Margenot, Andrew and Tong, Hanghang},
  title     = {Retrieval Based Time Series Forecasting},
  booktitle = {CIKM'22 Workshop on Applied Machine Learning Methods for Time Series Forecasting (AMLTS)},
  year      = {2022},
  address   = {Atlanta, GA, USA},
  note      = {Workshop paper; arXiv:2209.13525},
  eprint    = {2209.13525},
  archivePrefix = {arXiv},
  primaryClass  = {cs.LG},
  url       = {https://arxiv.org/abs/2209.13525}
}

@inproceedings{han2025raft,
  title     = {Retrieval Augmented Time Series Forecasting},
  author    = {Han, Sungwon and Lee, Seungeon and Cha, Meeyoung and Arik, Sercan O and Yoon, Jinsung},
 booktitle={Proceedings of the International Conference on Machine Learning (ICML)},
  pages     = {21774--21797},
  year      = {2025},
  editor    = {Singh, Aarti and Fazel, Maryam and Hsu, Daniel and Lacoste-Julien, Simon and Berkenkamp, Felix and Maharaj, Tegan and Wagstaff, Kiri and Zhu, Jerry},
  volume    = {267},
  series    = {Proceedings of Machine Learning Research},
  month     = {13--19 Jul},
  publisher = {PMLR},
  url       = {https://proceedings.mlr.press/v267/han25d.html}
}

@misc{wang2024ratsf,
  author        = {Wang, Tianfeng and Cui, Gaojie},
  title         = {{RATSF}: Empowering Customer Service Volume Management through Retrieval-Augmented Time-Series Forecasting},
  year          = {2024},
  eprint        = {2403.04180},
  archivePrefix = {arXiv},
  primaryClass  = {cs.LG},
  doi           = {10.48550/arXiv.2403.04180},
  url           = {https://arxiv.org/abs/2403.04180}
}

@misc{tire2024raf,
  author        = {Tire, Kutay and Taga, Ege Onur and Ildiz, Muhammed Emrullah and Oymak, Samet},
  title         = {Retrieval Augmented Time Series Forecasting},
  year          = {2024},
  eprint        = {2411.08249},
  archivePrefix = {arXiv},
  primaryClass  = {cs.LG},
  doi           = {10.48550/arXiv.2411.08249},
  url           = {https://arxiv.org/abs/2411.08249}
}

@inproceedings{ning2025tsrag,
  author    = {Ning, Kanghui and Pan, Zijie and Liu, Yu and Jiang, Yushan and Zhang, James and Rasul, Kashif and Schneider, Anderson and Ma, Lintao and Nevmyvaka, Yuriy and Song, Dongjin},
  title     = {{TS-RAG}: Retrieval-Augmented Generation based Time Series Foundation Models are Stronger Zero-Shot Forecaster},
  booktitle = {Advances in Neural Information Processing Systems (NeurIPS)},  volume    = {38},
  year      = {2025},
  publisher = {Curran Associates, Inc.},
  doi       = {10.52202/085713-5448},
}

@inproceedings{yang2024timerag,
  author    = {Yang, Silin and Wang, Dong and Zheng, Haoqi and Jin, Ruochun},
  title     = {{TimeRAG}: Boosting {LLM} Time Series Forecasting via Retrieval-Augmented Generation},
  booktitle = {ICASSP 2025 -- 2025 IEEE International Conference on Acoustics, Speech and Signal Processing (ICASSP)},
  year      = {2025},
  publisher = {IEEE},
  address   = {Hyderabad, India},
  doi       = {10.1109/ICASSP49660.2025.10889933},
  url       = {https://doi.org/10.1109/ICASSP49660.2025.10889933}
}

@article{ruan2026rast,
  author  = {Ruan, Weilin and Dang, Xilin and Zhou, Ziyu and Lyu, Sisuo and Liang, Yuxuan},
  title   = {A Retrieval Augmented Spatio-Temporal Framework for Traffic Prediction},
  journal = {Proceedings of the AAAI Conference on Artificial Intelligence},
  year    = {2026},
  volume  = {40},
  number  = {46},
  pages   = {39163--39172},
  doi     = {10.1609/aaai.v40i46.41264},
  url     = {https://ojs.aaai.org/index.php/AAAI/article/view/41264}
}

@inproceedings{ref58crm,
  author    = {Kang, Junhyeok and Seo, Jun and Park, Soyeon and Han, Sangjun and Bae, Seohui and Choe, Hyeokjun and Lee, Soonyoung},
  title     = {Channel-Wise Retrieval for Multivariate Time Series Forecasting},
  booktitle = {ICASSP 2026 -- 2026 IEEE International Conference on Acoustics, Speech and Signal Processing (ICASSP)},
  year      = {2026},
  publisher = {IEEE},
  address   = {Barcelona, Spain},
  doi       = {10.1109/ICASSP55912.2026.11463178},
  url       = {https://doi.org/10.1109/ICASSP55912.2026.11463178}
}

@inproceedings{liu2025llmad,
  author    = {Liu, Jun and Zhang, Chaoyun and Qian, Jiaxu and Ma, Minghua and Qin, Si and Bansal, Chetan and Lin, Qingwei and Rajmohan, Saravan and Zhang, Dongmei},
  title     = {Large Language Models can Deliver Accurate and Interpretable Time Series Anomaly Detection},
  booktitle = {Proceedings of the 31st ACM SIGKDD Conference on Knowledge Discovery and Data Mining V.2},
  year      = {2025},
  pages     = {4623--4634},
  publisher = {Association for Computing Machinery},
  address   = {New York, NY, USA},
  doi       = {10.1145/3711896.3737239},
  url       = {https://doi.org/10.1145/3711896.3737239}
}

@misc{maru2025ratfm,
  author        = {Maru, Chihiro and Sato, Shoetsu},
  title         = {{RATFM}: Retrieval-Augmented Time Series Foundation Model for Anomaly Detection},
  year          = {2025},
  eprint        = {2506.02081},
  archivePrefix = {arXiv},
  primaryClass  = {cs.LG},
  doi           = {10.48550/arXiv.2506.02081},
  url           = {https://arxiv.org/abs/2506.02081}
}

@misc{alerti2026,
  author        = {Truong, Xuan-Thong and Le, Trung-Kien and Kieu, Tung and Nguyen, Thi-Thu and Nguyen, Nhat-Hai},
  title         = {{ALER-TI}: Aligned Latent Embedding Retrieval for Time Series Imputation},
  year          = {2026},
  eprint        = {2607.07640},
  archivePrefix = {arXiv},
  primaryClass  = {cs.LG},
  doi           = {10.48550/arXiv.2607.07640},
  url           = {https://arxiv.org/abs/2607.07640}
}

@misc{reditt2026,
  author        = {Lyu, Saiyue and Zhang, Zhitian and Deng, Ruizhi and Durand, Thibaut},
  title         = {{ReDiTT}: Retrieval Augmented Conditional Diffusion Transformers for Asynchronous Time Series},
  year          = {2026},
  eprint        = {2607.12391},
  archivePrefix = {arXiv},
  primaryClass  = {cs.LG},
  doi           = {10.48550/arXiv.2607.12391},
  url           = {https://arxiv.org/abs/2607.12391}
}

@inproceedings{shi2026kronos,
  title={Kronos: A foundation model for the language of financial markets},
  author={Shi, Yu and Fu, Zongliang and Chen, Shuo and Zhao, Bohan and Xu, Wei and Zhang, Changshui and Li, Jian},
  booktitle={Proceedings of the AAAI Conference on Artificial Intelligence},
  volume={40},
  number={30},
  pages={25366--25373},
  year={2026}
}

@inproceedings{emerge2024,
  author    = {Zhu, Yinghao and Ren, Changyu and Wang, Zixiang and Zheng, Xiaochen and Xie, Shiyun and Feng, Junlan and Zhu, Xi and Li, Zhoujun and Ma, Liantao and Pan, Chengwei},
  title     = {{EMERGE}: Enhancing Multimodal Electronic Health Records Predictive Modeling with Retrieval-Augmented Generation},
  booktitle = {Proceedings of the 33rd ACM International Conference on Information and Knowledge Management},
  year      = {2024},
  pages     = {3549--3559},
  publisher = {Association for Computing Machinery},
  address   = {New York, NY, USA},
  doi       = {10.1145/3627673.3679582},
  url       = {https://doi.org/10.1145/3627673.3679582}
}

@inproceedings{Malhotra2015LongST,
  title={Long Short Term Memory Networks for Anomaly Detection in Time Series},
  author={Pankaj Malhotra and Lovekesh Vig and Gautam M. Shroff and Puneet Agarwal},
  booktitle={The European Symposium on Artificial Neural Networks},
  year={2015},
  url={https://api.semanticscholar.org/CorpusID:43680425}
}

@article{malhotra2016lstm,
  title={LSTM-based encoder-decoder for multi-sensor anomaly detection},
  author={Malhotra, Pankaj and Ramakrishnan, Anusha and Anand, Gaurangi and Vig, Lovekesh and Agarwal, Puneet and Shroff, Gautam},
  journal={arXiv preprint arXiv:1607.00148},
  year={2016}
}

@article{zou2024novel,
  title={A novel deep reinforcement learning based automated stock trading system using cascaded lstm networks},
  author={Zou, Jie and Lou, Jiashu and Wang, Baohua and Liu, Sixue},
  journal={Expert Systems with Applications},
  volume={242},
  pages={122801},
  year={2024},
  publisher={Elsevier}
}

@article{karim2019multivariate,
  title={Multivariate LSTM-FCNs for time series classification},
  author={Karim, Fazle and Majumdar, Somshubra and Darabi, Houshang and Harford, Samuel},
  journal={Neural networks},
  volume={116},
  pages={237--245},
  year={2019},
  publisher={Elsevier}
}

@inproceedings{li2021shapenet,
  title     = {ShapeNet: A Shapelet-Neural Network Approach for Multivariate Time Series Classification},
  author    = {Li, Guozhong and Choi, Byron and Xu, Jianliang and Bhowmick, Sourav S. and Chun, Kwok-Pan and Wong, Grace Lai-Hung},
  booktitle = {Proceedings of the AAAI Conference on Artificial Intelligence},
  volume    = {35},
  number    = {9},
  pages     = {8375--8383},
  year      = {2021},
  doi       = {10.1609/aaai.v35i9.17018}
}

@inproceedings{Zhang2020TapNetMT,
  title={TapNet: Multivariate Time Series Classification with Attentional Prototypical Network},
  author={Xuchao Zhang and Yifeng Gao and Jessica Lin and Chang-Tien Lu},
  booktitle={Proceedings of the AAAI Conference on Artificial Intelligence},
  year={2020},
  url={https://api.semanticscholar.org/CorpusID:210703726}
}

@inproceedings{tang2020interpretable,
  title={Interpretable time-series classification on few-shot samples},
  author={Tang, Wensi and Liu, Lu and Long, Guodong},
  booktitle={2020 international joint conference on neural networks (IJCNN)},
  pages={1--8},
  year={2020},
  organization={IEEE}
}

@ARTICLE{10375830,
  author={Wan, Xiaoxue and Cen, Lihui and Chen, Xiaofang and Xie, Yongfang and Gui, Weihua},
  journal={IEEE Transactions on Cybernetics}, 
  title={Memory Shapelet Learning for Early Classification of Streaming Time Series}, 
  year={2024},
  volume={54},
  number={5},
  pages={2757-2770},
  doi={10.1109/TCYB.2023.3337550}}

@article{zhou2025enhancing,
  title={Enhancing llm reasoning for time series classification by tailored thinking and fused decision},
  author={Zhou, Jiahui and Li, Dan and Li, Lin and Chen, Zhuomin and Wu, Shunyu and Ye, Haozheng and Lou, Jian and Spanos, Costas J},
  journal={arXiv preprint arXiv:2506.00807},
  year={2025}
}

@inproceedings{wang2025colacare,
  title={Colacare: Enhancing electronic health record modeling through large language model-driven multi-agent collaboration},
  author={Wang, Zixiang and Zhu, Yinghao and Zhao, Huiya and Zheng, Xiaochen and Sui, Dehao and Wang, Tianlong and Tang, Wen and Wang, Yasha and Harrison, Ewen and Pan, Chengwei and others},
  booktitle={Proceedings of the ACM on Web Conference 2025},
  pages={2250--2261},
  year={2025}
}

@article{yoon2025momemto,
  title={Momemto: Patch-based memory gate model in time series foundation model},
  author={Yoon, Samuel and Kim, Jongwon and Ha, Juyoung and Ko, Young Myoung},
  journal={arXiv preprint arXiv:2509.18751},
  year={2025}
}

@article{Li2025MemMambaADMS,
  title={MemMambaAD: Memory-augmented state space model for multivariate time series anomaly detection},
  author={Gang Li and Mingchao Ge and Jin Wan and Delong Han and Min Li and Mingle Zhou},
  journal={Eng. Appl. Artif. Intell.},
  year={2025},
  volume={158},
  pages={111308},
  url={https://api.semanticscholar.org/CorpusID:279357027}
}

@article{che2018recurrent,
  title={Recurrent neural networks for multivariate time series with missing values},
  author={Che, Zhengping and Purushotham, Sanjay and Cho, Kyunghyun and Sontag, David and Liu, Yan},
  journal={Scientific reports},
  volume={8},
  number={1},
  pages={6085},
  year={2018},
  publisher={Nature Publishing Group UK London}
}

@inproceedings{3327757.3327783,
author = {Cao, Wei and Wang, Dong and Li, Jian and Zhou, Hao and Li, Yitan and Li, Lei},
title = {BRITS: bidirectional recurrent imputation for time series},
year = {2018},
publisher = {Curran Associates Inc.},
address = {Red Hook, NY, USA},
booktitle = {Proceedings of the 32nd International Conference on Neural Information Processing Systems},
pages = {6776–6786},
numpages = {11},
location = {Montr{\'e}al, Canada},
series = {NIPS'18}
}

@article{zhu2024realm,
  title={Realm: Rag-driven enhancement of multimodal electronic health records analysis via large language models},
  author={Zhu, Yinghao and Ren, Changyu and Xie, Shiyun and Liu, Shukai and Ji, Hangyuan and Wang, Zixiang and Sun, Tao and He, Long and Li, Zhoujun and Zhu, Xi and others},
  journal={arXiv preprint arXiv:2402.07016},
  year={2024}
}

@article{
ye2026tsreasoner,
title={{TS}-Reasoner: Domain-Oriented Time Series Inference Agents for Reasoning and Automated Analysis},
author={Wen Ye and Wei Yang and Defu Cao and Yizhou Zhang and Lumingyuan Tang and Jie Cai and Yan Liu},
journal={Transactions on Machine Learning Research},
issn={2835-8856},
year={2026},
url={https://openreview.net/forum?id=yhy7Vigjcf},
note={}
}

@inproceedings{park2020learning,
  title     = {Learning Memory-Guided Normality for Anomaly Detection},
  author    = {Park, Hyunjong and Noh, Jongyoun and Ham, Bumsub},
  booktitle = {Proceedings of the IEEE/CVF Conference on Computer Vision and Pattern Recognition (CVPR)},
  pages     = {14372--14381},
  year      = {2020}
}

@inproceedings{wu2024stanhop,
  title     = {{STanHop}: Sparse Tandem Hopfield Model for Memory-Enhanced Time Series Prediction},
  author    = {Wu, Dennis and Hu, Jerry Yao-Chieh and Li, Weijian and Chen, Bo-Yu and Liu, Han},
  booktitle = {International Conference on Learning Representations (ICLR)},
  year      = {2024}
}

@inproceedings{mmnet2025,
  author={Xiaoye Miao and Han Shi and Yi Yuan and Daozhan Pan and Yangyang Wu and Xiaohua Pan},
  title={MMNet: Missing-Aware and Memory-Enhanced Network for Multivariate Time Series Imputation},
  year={2025},
  cdate={1735689600000},
  pages={3208-3216},
  url={https://doi.org/10.24963/ijcai.2025/357},
  booktitle={IJCAI},
}

@article{yu2025prime,
  title   = {Imputation with Inter-Series Information from Prototypes for Healthcare Time Series},
  author  = {Yu, Z.-H. and Ma, L.-T. and Wang, Y.-S. and others},
  journal = {Journal of Computer Science and Technology},
  volume  = {40},
  number  = {6},
  pages   = {1499--1511},
  year    = {2025},
  doi     = {10.1007/s11390-025-4333-3}
}

@inproceedings{kang2026craft,
author = {Zhang, Yingwei and Bu, Ke and Zhuang, Zhuoran and Xie, Tao and Yu, Yao and Li, Dong and Guo, Yang and Lv, Detao},
title = {CRAFT: time series forecasting with cross-future behavior awareness},
year = {2025},
isbn = {978-1-956792-06-5},
url = {https://doi.org/10.24963/ijcai.2025/785},
doi = {10.24963/ijcai.2025/785},
booktitle = {Proceedings of the Thirty-Fourth International Joint Conference on Artificial Intelligence},
articleno = {785},
numpages = {9},
location = {Montreal, Canada},
series = {IJCAI '25}
}

@inproceedings{
cao2026enhancing,
title={Enhancing Multivariate Time Series Forecasting with Global Temporal Retrieval},
author={Fanpu Cao and Lu Dai and Jindong Han and Hui Xiong},
  booktitle = {International Conference on Learning Representations (ICLR)},
year={2026},
}

@article{zhang2025timeraf,
  title   = {{TimeRAF}: Retrieval-Augmented Foundation Model for Zero-Shot Time Series Forecasting},
  author  = {Zhang, Huanyu and Xu, Chang and Zhang, Yi-Fan and Zhang, Zhang and Wang, Liang and Bian, Jiang},
  journal = {IEEE Transactions on Knowledge and Data Engineering},
  volume  = {37},
  number  = {9},
  pages   = {5654--5665},
  year    = {2025},
  doi     = {10.1109/TKDE.2025.3579137}
}

@inproceedings{lyu2026tsmemory,
  title     = {{TS-Memory}: Plug-and-Play Memory for Time Series Foundation Models},
  author    = {Lyu, Sisuo and Zhong, Siru and Chen, Tiegang and Ruan, Weilin and Liu, Qingxiang and Lv, Taiqiang and Wen, Qingsong and Wong, Raymond Chi-Wing and Liang, Yuxuan},
  booktitle = {Proceedings of the 32nd ACM SIGKDD Conference on Knowledge Discovery and Data Mining},
  pages     = {3562--3572},
  year      = {2026}
}

@inproceedings{yu2026memts,
author = {Yu, Xiaoyun and Fan, Li and Qiu, Xiangfei and Dong, Nanqing and Huang, Yonggui and Qi, Honggang and Pu, Geguang and Ouyang, Wanli and Chen, Xi and Hu, Jilin},
title = {MEMTS: Internalizing Domain Knowledge via Parameterized Memory for Retrieval-Free Domain Adaptation of Time Series Foundation Models},
year = {2026},
doi = {10.1145/3770855.3818032},
booktitle = {Proceedings of the 32nd ACM SIGKDD Conference on Knowledge Discovery and Data Mining},
numpages = {12},
}

@article{lee2025hamn,
  title   = {Memory Augmented Coherent Probabilistic Forecasts for Hierarchically Related Time Series},
  author  = {Lee, Junyong and Byun, Yunseon and Koo, Byoungmo and Baek, Jun-Geol},
  journal = {Neurocomputing},
  volume  = {653},
  pages   = {131075},
  year    = {2025},
  doi     = {10.1016/j.neucom.2025.131075}
}

@inproceedings{zhou2026seraf,
  title     = {Semantics-Enhanced Retrieval-Augmented Time Series Forecasting},
  author    = {Zhou, Shiqiao and Wu, Zipeng and Sch{\"o}ner, Holger and Fouch{\'e}, Edouard and Wilson, I. A. G. and Wang, Shuo},
  booktitle = {ICML 2026 Workshop on Forecasting as a New Frontier of Intelligence},
  year      = {2026},
  note      = {Workshop paper},
  eprint    = {2606.14941},
  archivePrefix = {arXiv}
}

@inproceedings{lee2022learning,
  title     = {Learning to Remember Patterns: Pattern Matching Memory Networks for Traffic Forecasting},
  author    = {Lee, Hyunwook and Jin, Seungmin and Chu, Hyeshin and Lim, Hongkyu and Ko, Sungahn},
booktitle={Proceedings of the International Conference on Learning Representations (ICLR)},   year      = {2022},
  url       = {https://openreview.net/forum?id=wwDg3bbYBIq}
}

@inproceedings{jiang2023spatiotemporal,
  title   = {Spatio-Temporal Meta-Graph Learning for Traffic Forecasting},
  author  = {Jiang, Renhe and Wang, Zhaonan and Yong, Jiawei and Jeph, Puneet and Chen, Quanjun and Kobayashi, Yasumasa and Song, Xuan and Fukushima, Shintaro and Suzumura, Toyotaro},
  booktitle={Proceedings of the AAAI Conference on Artificial Intelligence},
  volume  = {37},
  number  = {7},
  pages   = {8078--8086},
  year    = {2023},
  doi     = {10.1609/aaai.v37i7.25976}
}

@data{caltrans_pems,
doi = {10.21227/yj7k-b369},
url = {https://dx.doi.org/10.21227/yj7k-b369},
author = {Li. Yan},
publisher = {IEEE Dataport},
title = {PeMS},
year = {2025} }

@article{qiu2024tfb,
  title   = {{TFB}: Towards Comprehensive and Fair Benchmarking of Time Series Forecasting Methods},
  author  = {Qiu, Xiangfei and Hu, Jilin and Zhou, Lekui and Wu, Xingjian and Du, Junyang and Zhang, Buang and Guo, Chenjuan and Zhou, Aoying and Jensen, Christian S. and Sheng, Zhenli and Yang, Bin},
  journal = {Proceedings of the VLDB Endowment},
  volume  = {17},
  number  = {9},
  pages   = {2363--2377},
  year    = {2024},
  doi     = {10.14778/3665844.3665863}
}

@inproceedings{godahewa2021monash,
  title     = {Monash Time Series Forecasting Archive},
  author    = {Godahewa, Rakshitha and Bergmeir, Christoph and Webb, Geoffrey I. and Hyndman, Rob J. and Montero-Manso, Pablo},
  booktitle = {Neural Information Processing Systems Track on Datasets and Benchmarks},
  year      = {2021},
}

@article{dau2019ucr,
  title   = {The {UCR} Time Series Archive},
  author  = {Dau, Hoang Anh and Bagnall, Anthony and Kamgar, Kaveh and Yeh, Chin-Chia Michael and Zhu, Yan and Gharghabi, Shaghayegh and Ratanamahatana, Chotirat Ann and Keogh, Eamonn},
  journal = {IEEE/CAA Journal of Automatica Sinica},
  volume  = {6},
  number  = {6},
  pages   = {1293--1305},
  year    = {2019},
  doi     = {10.1109/JAS.2019.1911747}
}

@inproceedings{liu2024tsbad,
  title     = {The Elephant in the Room: Towards A Reliable Time-Series Anomaly Detection Benchmark},
  author    = {Liu, Qinghua and Paparrizos, John},
  booktitle = {Advances in Neural Information Processing Systems (NeurIPS)},  volume    = {37},
  year      = {2024},
  note      = {Datasets and Benchmarks Track},
  doi       = {10.52202/079017-3437},
}

@inproceedings{toye2025imputation,
  title     = {Benchmarking Missing Data Imputation Methods for Time Series Using Real-World Test Cases},
  author    = {Toye, Adedolapo Aishat and Celik, Asuman and Kleinberg, Samantha},
  booktitle = {Proceedings of the Sixth Conference on Health, Inference, and Learning},
  series    = {Proceedings of Machine Learning Research},
  volume    = {287},
  pages     = {480--501},
  publisher = {PMLR},
  year      = {2025},
  url       = {https://proceedings.mlr.press/v287/toye25a.html}
}

@article{shao2025basicts,
  title   = {Exploring Progress in Multivariate Time Series Forecasting: Comprehensive Benchmarking and Heterogeneity Analysis},
  author  = {Shao, Zezhi and Wang, Fei and Xu, Yongjun and Wei, Wei and Yu, Chengqing and Zhang, Zhao and Yao, Di and Sun, Tao and Jin, Guangyin and Cao, Xin and Cong, Gao and Jensen, Christian S. and Cheng, Xueqi},
  journal = {IEEE Transactions on Knowledge and Data Engineering},
  volume  = {37},
  number  = {1},
  pages   = {291--305},
  year    = {2025},
  doi     = {10.1109/TKDE.2024.3484454}
}

@inproceedings{liu2023largest,
  title     = {{LargeST}: A Benchmark Dataset for Large-Scale Traffic Forecasting},
  author    = {Liu, Xu and Xia, Yutong and Liang, Yuxuan and Hu, Junfeng and Wang, Yiwei and Bai, Lei and Huang, Chao and Liu, Zhenguang and Hooi, Bryan and Zimmermann, Roger},
  booktitle = {Advances in Neural Information Processing Systems (NeurIPS)},  volume    = {36},
  year      = {2023},
  note      = {Datasets and Benchmarks Track},
}

@article{makridakis2022m5,
  title   = {{M5} Accuracy Competition: Results, Findings, and Conclusions},
  author  = {Makridakis, Spyros and Spiliotis, Evangelos and Assimakopoulos, Vassilios},
  journal = {International Journal of Forecasting},
  volume  = {38},
  number  = {4},
  pages   = {1346--1364},
  year    = {2022},
  doi     = {10.1016/j.ijforecast.2021.11.013}
}

@inproceedings{zhang2024probts,
  title     = {{ProbTS}: Benchmarking Point and Distributional Forecasting across Diverse Prediction Horizons},
  author    = {Zhang, Jiawen and Wen, Xumeng and Zhang, Zhenwei and Zheng, Shun and Li, Jia and Bian, Jiang},
  booktitle = {Advances in Neural Information Processing Systems (NeurIPS)},  volume    = {37},
  year      = {2024},
  note      = {Datasets and Benchmarks Track},
  doi       = {10.52202/079017-1523},
}

@article{ansari2024chronos,
  title   = {Chronos: Learning the Language of Time Series},
  author  = {Ansari, Abdul Fatir and Stella, Lorenzo and T{\"u}rkmen, Ali Caner and Zhang, Xiyuan and Mercado, Pedro and Shen, Huibin and Shchur, Oleksandr and Rangapuram, Syama Sundar and Arango, Sebastian Pineda and Kapoor, Shubham and Zschiegner, Jasper and Maddix, Danielle C. and Wang, Hao and Mahoney, Michael W. and Torkkola, Kari and Wilson, Andrew Gordon and Bohlke-Schneider, Michael and Wang, Bernie},
  journal = {Transactions on Machine Learning Research},
  year    = {2024},
  url     = {https://openreview.net/group?id=TMLR}
}

@inproceedings{cohen2025boom,
  title     = {This Time is Different: An Observability Perspective on Time Series Foundation Models},
  author    = {Cohen, Ben and Khwaja, Emaad and Doubli, Youssef and Lemaachi, Salahidine and Lettieri, Chris and Masson, Charles and Miccinilli, Hugo and Ram{\'e}, Elise and Ren, Qiqi and Rostamizadeh, Afshin and du Terrail, Jean and Toon, Anna-Monica and Wang, Kan and Xie, Stephan and Xu, Zongzhe and Zhukova, Viktoriya and Asker, David and Talwalkar, Ameet S. and Abou-Amal, Othmane},
  booktitle = {Advances in Neural Information Processing Systems (NeurIPS)},  volume    = {38},
  year      = {2025},
}

@inproceedings{liu2024timemmd,
  title     = {{Time-MMD}: Multi-Domain Multimodal Dataset for Time Series Analysis},
  author    = {Liu, Haoxin and Xu, Shangqing and Zhao, Zhiyuan and Kong, Lingkai and Kamarthi, Harshavardhan and Sasanur, Aditya B. and Sharma, Megha and Cui, Jiaming and Wen, Qingsong and Zhang, Chao and Prakash, B. Aditya},
  booktitle = {Advances in Neural Information Processing Systems (NeurIPS)},  volume    = {37},
  year      = {2024},
  note      = {Datasets and Benchmarks Track},
  doi       = {10.52202/079017-2476},
}

@inproceedings{williams2025context,
  title     = {Context is Key: A Benchmark for Forecasting with Essential Textual Information},
  author    = {Williams, Andrew Robert and Ashok, Arjun and Marcotte, {\'E}tienne and Zantedeschi, Valentina and Subramanian, Jithendaraa and Riachi, Roland and Requeima, James and Lacoste, Alexandre and Rish, Irina and Chapados, Nicolas and Drouin, Alexandre},
booktitle={Proceedings of the International Conference on Machine Learning (ICML)},
  volume    = {267},
  pages     = {66887--66944},
  publisher = {PMLR},
  year      = {2025},
  url       = {https://proceedings.mlr.press/v267/williams25a.html}
}

@inproceedings{chang2025timeimm,
  title     = {{Time-IMM}: A Dataset and Benchmark for Irregular Multimodal Multivariate Time Series},
  author    = {Chang, Ching and Hwang, Jeehyun and Shi, Yidan and Wang, Haixin and Wang, Wei and Peng, Wen-Chih and Chen, Tien-Fu},
  booktitle = {Advances in Neural Information Processing Systems (NeurIPS)},  volume    = {38},
  year      = {2025},
  note      = {Datasets and Benchmarks Track},
}

@inproceedings{chen2025trace,
  title     = {{TRACE}: Grounding Time Series in Context for Multimodal Embedding and Retrieval},
  author    = {Chen, Jialin and Zhao, Ziyu and Nurbek, Gaukhar and Feng, Aosong and Maatouk, Ali and Tassiulas, Leandros and Gao, Yifeng and Ying, Rex},
  booktitle = {Advances in Neural Information Processing Systems (NeurIPS)},  volume    = {38},
  year      = {2025},
}

@inproceedings{gwiazda2026timeseriesexam,
  title     = {{TimeSeriesExamAgent}: Creating {TimeSeries} Reasoning Benchmarks at Scale},
  author    = {Gwiazda, Malgorzata and Cai, Yifu and Goswami, Mononito and Choudhry, Arjun and Dubrawski, Artur},
booktitle={Proceedings of the International Conference on Learning Representations (ICLR)},   year      = {2026},
  url       = {https://iclr.cc/virtual/2026/poster/10010760}
}

@inproceedings{wang2025itformer,
  title     = {{ITFormer}: Bridging Time Series and Natural Language for Multi-Modal {QA} with Large-Scale Multitask Dataset},
  author    = {Wang, Yilin and Lei, Peixuan and Song, Jie and Hao, Yuzhe and Chen, Tao and Zhang, Yuxuan and Jia, Lei and Li, Yuanxiang and Wei, Zhongyu},
booktitle={Proceedings of the International Conference on Machine Learning (ICML)},
  series    = {Proceedings of Machine Learning Research},
  volume    = {267},
  pages     = {63324--63344},
  publisher = {PMLR},
  year      = {2025},
}

@inproceedings{kong2025timemqa,
  title     = {Time-{MQA}: Time Series Multi-Task Question Answering with Context Enhancement},
  author    = {Kong, Yaxuan and Yang, Yiyuan and Hwang, Yoontae and Du, Wenjie and Zohren, Stefan and Wang, Zhangyang and Jin, Ming and Wen, Qingsong},
  booktitle = {Proceedings of the 63rd Annual Meeting of the Association for Computational Linguistics (Volume 1: Long Papers)},
  pages     = {29736--29753},
  publisher = {Association for Computational Linguistics},
  address   = {Vienna, Austria},
  year      = {2025},
  doi       = {10.18653/v1/2025.acl-long.1437},
}

@article{xie2025chatts,
  title   = {{ChatTS}: Aligning Time Series with {LLM}s via Synthetic Data for Enhanced Understanding and Reasoning},
  author  = {Xie, Zhe and Li, Zeyan and He, Xiao and Xu, Longlong and Wen, Xidao and Zhang, Tieying and Chen, Jianjun and Shi, Rui and Pei, Dan},
  journal = {Proceedings of the VLDB Endowment},
  volume  = {18},
  number  = {8},
  pages   = {2385--2398},
  year    = {2025},
  doi     = {10.14778/3742728.3742735}
}

@inproceedings{huang2025manyminds,
  title     = {Many Minds, One Goal: Time Series Forecasting via Sub-task Specialization and Inter-agent Cooperation},
  author    = {Huang, Qihe and Zhou, Zhengyang and Li, Yangze and Yang, Kuo and Wang, Binwu and Wang, Yang},
  booktitle = {Advances in Neural Information Processing Systems (NeurIPS)},  volume    = {38},
  year      = {2025},
}

@inproceedings{li2025timeseriesgym,
  title     = {{TimeSeriesGym}: A Scalable Benchmark for (Time Series) Machine Learning Engineering Agents},
  author    = {Li, Xinyu and Cai, Yifu and Goswami, Mononito and Wili{\'n}ski, Micha{\l} and Welter, Gus and Dubrawski, Artur},
  booktitle = {Women in Machine Learning Workshop at NeurIPS},
  year      = {2025},
  note      = {Workshop paper},
  url       = {https://openreview.net/forum?id=8M3qAX6e5M}
}

@inproceedings{tan2025syntsbench,
  title     = {{SynTSBench}: Rethinking Temporal Pattern Learning in Deep Learning Models for Time Series},
  author    = {Tan, Qitai and Chen, Yiyun and Li, Mo and Gu, Ruiwen and Su, Yilin and Zhang, Xiao-Ping},
  booktitle = {Advances in Neural Information Processing Systems (NeurIPS)},  volume    = {38},
  year      = {2025},
  note      = {Datasets and Benchmarks Track},
}

@inproceedings{zumarraga2026tshaystack,
  title     = {{TS-Haystack}: A Multi-Scale Retrieval Benchmark for Time Series Language Models},
  author    = {Zumarraga, Nicolas and Kaar, Thomas and Wang, Ning and Xu, Maxwell A. and Rosenblattl, Max and Kreft, Markus and O'Sullivan, Kevin and Schmiedmayer, Paul and Langer, Patrick and Jakob, Robert},
  booktitle = {ICLR 2026 Workshop on Time Series in the Age of Large Models},
  year      = {2026},
  note      = {Workshop paper},
  url       = {https://openreview.net/forum?id=vGMayMuH83}
}

@article{alexandrov2020gluonts,
  title   = {{GluonTS}: Probabilistic and Neural Time Series Modeling in Python},
  author  = {Alexandrov, Alexander and Benidis, Konstantinos and Bohlke-Schneider, Michael and Flunkert, Valentin and Gasthaus, Jan and Januschowski, Tim and Maddix, Danielle C. and Rangapuram, Syama and Salinas, David and Schulz, Jasper and Stella, Lorenzo and T{\"u}rkmen, Ali Caner and Wang, Yuyang},
  journal = {Journal of Machine Learning Research},
  volume  = {21},
  number  = {116},
  pages   = {1--6},
  year    = {2020},
  url     = {https://www.jmlr.org/papers/v21/19-820.html}
}

@article{middlehurst2024aeon,
  title   = {aeon: a Python Toolkit for Learning from Time Series},
  author  = {Middlehurst, Matthew and Ismail-Fawaz, Ali and Guillaume, Antoine and Holder, Christopher and Guijo-Rubio, David and Bulatova, Guzal and Tsaprounis, Leonidas and Mentel, Lukasz and Walter, Martin and Sch{\"a}fer, Patrick and Bagnall, Anthony},
  journal = {Journal of Machine Learning Research},
  volume  = {25},
  number  = {289},
  pages   = {1--10},
  year    = {2024},
  url     = {https://www.jmlr.org/papers/v25/23-1444.html}
}

@article{bhatnagar2023merlion,
  title   = {Merlion: End-to-End Machine Learning for Time Series},
  author  = {Bhatnagar, Aadyot and Kassianik, Paul and Liu, Chenghao and Lan, Tian and Yang, Wenzhuo and Cassius, Rowan and Sahoo, Doyen and Arpit, Devansh and Subramanian, Sri and Woo, Gerald and Saha, Amrita and Jagota, Arun Kumar and Gopalakrishnan, Gokulakrishnan and Singh, Manpreet and Krithika, K C and Maddineni, Sukumar and Cho, Daeki and Zong, Bo and Zhou, Yingbo and Xiong, Caiming and Savarese, Silvio and Hoi, Steven and Wang, Huan},
  journal = {Journal of Machine Learning Research},
  volume  = {24},
  number  = {226},
  pages   = {1--6},
  year    = {2023},
  url     = {https://jmlr.org/papers/v24/22-0809.html}
}

@article{labar2006cognitive,
  title={Cognitive neuroscience of emotional memory},
  author={LaBar, Kevin S and Cabeza, Roberto},
  journal={Nature Reviews Neuroscience},
  volume={7},
  number={1},
  pages={54--64},
  year={2006},
  publisher={Nature Publishing Group UK London}
}

@article{kensinger2020retrieval,
  title={Retrieval of emotional events from memory},
  author={Kensinger, Elizabeth A and Ford, Jaclyn H},
  journal={Annual review of psychology},
  volume={71},
  number={1},
  pages={251--272},
  year={2020},
  publisher={Annual Reviews}
}

@inproceedings{cai2025disentangling,
  title={Disentangling long-short term state under unknown interventions for online time series forecasting},
  author={Cai, Ruichu and Huang, Haiqin and Jiang, Zhifan and Li, Zijian and Zhou, Changze and Liu, Yuequn and Liu, Yuming and Hao, Zhifeng},
  booktitle={Proceedings of the AAAI Conference on Artificial Intelligence},
  volume={39},
  number={15},
  pages={15641--15649},
  year={2025}
}

@inproceedings{chen2025learning,
  title={Learning to Extrapolate and Adjust: Two-Stage Meta-Learning for Concept Drift in Online Time Series Forecasting.},
  author={Chen, Weiqi and Zhu, Zhaoyang and Zhang, Yifan and Shen, Lefei and Yang, Linxiao and Wen, Qingsong and Sun, Liang},
  booktitle={IJCAI},
  pages={4869--4877},
  year={2025}
}

@article{zhan2025continuous,
  title={Continuous evolution pool: Taming recurring concept drift in online time series forecasting},
  author={Zhan, Tianxiang and Jin, Ming and He, Yuanpeng and Liang, Yuxuan and Deng, Yong and Pan, Shirui},
  journal={arXiv preprint arXiv:2506.14790},
  year={2025}
}

@inproceedings{zhong2025timeVLM,
  title = 	 {Time-{VLM}: Exploring Multimodal Vision-Language Models for Augmented Time Series Forecasting},
  author =       {Zhong, Siru and Ruan, Weilin and Jin, Ming and Li, Huan and Wen, Qingsong and Liang, Yuxuan},
booktitle={Proceedings of the International Conference on Machine Learning (ICML)},
  pages = 	 {78478--78497},
  year = 	 {2025},
  volume = 	 {267},
  series = 	 {Proceedings of Machine Learning Research},
  month = 	 {13--19 Jul},
  publisher =    {PMLR},
}

@article{vapnik2015learning,
  title={Learning using privileged information: similarity control and knowledge transfer},
  author={Vapnik, Vladimir and Izmailov, Rauf},
  journal={The Journal of Machine Learning Research},
  volume={16},
  number={1},
  pages={2023--2049},
  year={2015},
}

@inproceedings{karlsson2022privileged,
  title = 	 { Using time-series privileged information for provably efficient learning of prediction models },
  author =       {K.A. Karlsson, Rickard and Willbo, Martin and Hussain, Zeshan M. and Krishnan, Rahul G. and Sontag, David and Johansson, Fredrik},
  booktitle = 	 {Proceedings of The 25th International Conference on Artificial Intelligence and Statistics},
  pages = 	 {5459--5484},
  year = 	 {2022},
  volume = 	 {151},
  series = 	 {Proceedings of Machine Learning Research},
  month = 	 {28--30 Mar},
  publisher =    {PMLR},
}

@inproceedings{liu2025timekd,
  title={Efficient multivariate time series forecasting via calibrated language models with privileged knowledge distillation},
  author={Liu, Chenxi and Miao, Hao and Xu, Qianxiong and Zhou, Shaowen and Long, Cheng and Zhao, Yan and Li, Ziyue and Zhao, Rui},
  booktitle={2025 IEEE 41st International Conference on Data Engineering (ICDE)},
  pages={3165--3178},
  year={2025},
  organization={IEEE}
}

@article{assaad2022survey,
  title={Survey and evaluation of causal discovery methods for time series},
  author={Assaad, Charles K and Devijver, Emilie and Gaussier, Eric},
  journal={Journal of Artificial Intelligence Research},
  volume={73},
  pages={767--819},
  year={2022}
}

@article{hewamalage2021recurrent,
  title={Recurrent neural networks for time series forecasting: Current status and future directions},
  author={Hewamalage, Hansika and Bergmeir, Christoph and Bandara, Kasun},
  journal={International Journal of Forecasting},
  volume={37},
  number={1},
  pages={388--427},
  year={2021},
  publisher={Elsevier}
}

@article{gong2025causal,
  title={Causal discovery from temporal data: An overview and new perspectives},
  author={Gong, Chang and Zhang, Chuzhe and Yao, Di and Bi, Jingping and Li, Wenbin and Xu, Yongjun},
  journal={ACM Computing Surveys},
  volume={57},
  number={4},
  pages={1--38},
  year={2024},
  publisher={ACM New York, NY}
}

@article{cai2025causal,
  title={Causal-oriented representation learning for time-series forecasting based on the spatiotemporal information transformation},
  author={Cai, Sihua and Peng, Hao and Liu, Rui and Chen, Pei},
  journal={Communications Physics},
  volume={8},
  number={1},
  pages={242},
  year={2025},
  publisher={Nature Publishing Group UK London}
}

@inproceedings{nguyenspectral,
  title={Spectral Text Fusion: A Frequency-Aware Approach to Multimodal Time-Series Forecasting},
  author={Nguyen, Huu Hiep and Nguyen, Minh Hoang and Nguyen, Dung and Le, Hung},
  booktitle={The 29th International Conference on Artificial Intelligence and Statistics},
  year={2026}
}

@inproceedings{nguyenreviving,
  title={Reviving Error Correction in Modern Deep Time-Series Forecasting},
  author={Nguyen, Minh Hoang and Nguyen, Huu Hiep and Nguyen, Dung and Do, Kien and Le, Hung and others},
  booktitle={Proceedings of the International Conference on Machine Learning (ICML)},
 year={2026},
}

@INPROCEEDINGS{11392029,
  author={Le, Hung and Abbas, Sherif and Nguyen, Minh Hoang and Do, Van Dai and Nguyen, Huu Hiep and Nguyen, Dung},
  booktitle={2025 IEEE International Conference on Data Mining (ICDM)}, 
  title={Accelerating Long-Term Molecular Dynamics with Physics-Informed Time-Series Forecasting}, 
  year={2025},
  volume={},
  number={},
  pages={1340-1349},
  doi={10.1109/ICDM65498.2025.00143}}

@inproceedings{tu2024powerpm,
  title={Powerpm: Foundation model for power systems},
  author={Tu, Shihao and Zhang, Yupeng and Zhang, Jing and Fu, Zhendong and Zhang, Yin and Yang, Yang},
  booktitle = {Advances in Neural Information Processing Systems (NeurIPS)},  volume={37},
  pages={115233--115260},
  year={2024}
}

@article{jiang2024probabilistic,
  title={Probabilistic electricity price forecasting based on penalized temporal fusion transformer},
  author={Jiang, He and Pan, Sheng and Dong, Yao and Wang, Jianzhou},
  journal={Journal of Forecasting},
  volume={43},
  number={5},
  pages={1465--1491},
  year={2024},
  publisher={Wiley Online Library}
}

@inproceedings{xu2023transehr,
  title={Transehr: Self-supervised transformer for clinical time series data},
  author={Xu, Yanbo and Xu, Shangqing and Ramprassad, Manav and Tumanov, Alexey and Zhang, Chao},
  booktitle={Machine learning for health (ML4h)},
  pages={623--635},
  year={2023},
  organization={PMLR}
}

@inproceedings{staniek2024early,
  title = 	 {Early Prediction of Causes (not Effects) in Healthcare by Long-Term Clinical Time Series Forecasting},
  author =       {Staniek, Michael and Fracarolli, Marius and Hagmann, Michael and Riezler, Stefan},
  booktitle={Machine learning for health (ML4h)},
  year = 	 {2024},
  volume = 	 {252},
  organization={PMLR}
}

@inproceedings{li2025MAD,
  title = 	 {M$^2$AD: Multi-Sensor Multi-System Anomaly Detection through Global Scoring and Calibrated Thresholding},
  author =       {Alnegheimish, Sarah and He, Zelin and Reimherr, Matthew and Chandrayan, Akash and Pradhan, Abhinav and D'Angelo, Luca},
  booktitle = 	 {Proceedings of The 28th International Conference on Artificial Intelligence and Statistics},
  pages = 	 {4384--4392},
  year = 	 {2025},
  editor = 	 {Li, Yingzhen and Mandt, Stephan and Agrawal, Shipra and Khan, Emtiyaz},
  volume = 	 {258},
  series = 	 {Proceedings of Machine Learning Research},
  publisher =    {PMLR},
}

@article{zhang2025survey,
  title={A survey on the memory mechanism of large language model-based agents},
  author={Zhang, Zeyu and Dai, Quanyu and Bo, Xiaohe and Ma, Chen and Li, Rui and Chen, Xu and Zhu, Jieming and Dong, Zhenhua and Wen, Ji-Rong},
  journal={ACM Transactions on Information Systems},
  volume={43},
  number={6},
  pages={1--47},
  year={2025},
  publisher={ACM New York, NY}
}

@inproceedings{zhong2024memorybank,
  title={Memorybank: Enhancing large language models with long-term memory},
  author={Zhong, Wanjun and Guo, Lianghong and Gao, Qiqi and Ye, He and Wang, Yanlin},
  booktitle={Proceedings of the AAAI Conference on Artificial Intelligence},
  volume={38},
  number={17},
  pages={19724--19731},
  year={2024}
}

@inproceedings{song2020spatial,
  author       = {Chao Song and
                  Youfang Lin and
                  Shengnan Guo and
                  Huaiyu Wan},
  title        = {Spatial-Temporal Synchronous Graph Convolutional Networks: {A} New
                  Framework for Spatial-Temporal Network Data Forecasting},
  booktitle    = {{AAAI}},
  pages        = {914--921},
  publisher    = {{AAAI} Press},
  year         = {2020}
}

\clearpage
\setcounter{section}{0}

\renewcommand{\thesection}{S\arabic{section}}

\section{Supplementary Material: Glossary}
\label{sec:glossary}

This glossary collects the time-series-memory terminology introduced or used consistently throughout the survey, together with a small number of foundational concepts the survey builds on. Full treatments, formal definitions, and citations are given in the referenced sections; entries here are deliberately short.

\small
\begin{description}[leftmargin=1.35em,style=sameline,itemsep=3pt,font=\normalfont\bfseries]

\item[Agentic memory (agentic curated memory).] An external store $\mathcal{M}$ over which a controller $\pi$ exercises a \emph{write}, \emph{read}, and \emph{forget} policy, so the store is actively revised in light of new interactions rather than fixed once built. Explicit in representation, policy-bounded in capacity, controller-issued in read/write, and online in persistence (\Cref{sec:agentic}).

\item[Auditability.] The property that a prediction can be traced back to the specific stored episode, prototype, or record that produced it, e.g., via the read weights of \Cref{eq:explicit-read}. Cited as a benefit of explicit, retrieval, and agentic memory over internal memory (\Cref{sec:whymemory}).

\item[Capacity (axis).] One of the four axes for comparing memory mechanisms: \emph{bounded} capacity is fixed in dimension independent of sequence length (e.g., a hidden state or a fixed slot bank); \emph{unbounded} capacity grows with the data available, as in a retrieval index (\Cref{sec:4axes}).

\item[Causal memory.] A proposed extension in which temporal memory retains causal structure, i.e., dependencies, delayed effects, and intervention outcomes, rather than only associations, so that stored experience can support counterfactual reasoning (\Cref{sect: chall8}).

\item[Controller.] The component (rule-based, learned, or an LLM-based agent) that issues the write, read, and forget operations over an agentic memory store, denoted $\pi$ in \Cref{eq:agentic-lifecycle} (\Cref{sec:agentic-mechanism}).

\item[Curated retrieval.] An intermediate regime, between passive retrieval and fully online agentic memory, in which a store's contents are chosen or distilled by an agent but then remain read-only at deployment; most surveyed ``agentic'' time-series systems fall here rather than in the fully online regime (\Cref{sec:agentic-design}).

\item[Effective temporal memory capacity.] A proposed evaluation notion: the degree to which actionable, temporally distant information remains recoverable and usable to condition future predictions, characterized by how retrieval fidelity degrades with temporal distance (\Cref{sect: chall1}).

\item[Episodic memory.] An agentic memory type that records specific past events, e.g., a context/forecast/error triple from a single interaction; typically written per instance and read at prediction or reflection time, with fast decay under regime change (\Cref{tab:agentic-types}).

\item[Evolution operator ($\mathcal{E}$).] The agentic-memory operator that consolidates, revises, and evicts stored content after admission. It is the identity in retrieval memory, where the store is built once and thereafter only read, and is non-trivial only where a forget policy actually runs (\Cref{sec:retrieval}, \Cref{sec:agentic-mechanism}, \Cref{sec:agentic-design}).

\item[Explicit memory.] A store of individually addressable entries (slots, vectors, or records), each written and read as a unit rather than entangled in weights or a dense state; subdivided into verbatim, slot, and learned memory (\Cref{sec:explicit}).

\item[External memory.] Historical information maintained outside the model's latent state in a separately addressable store that the model can read from and, depending on the mechanism, write to. External memory can retain information beyond the model's fixed internal capacity and may incorporate content not seen during training. It encompasses explicit, retrieval, and agentic memory, whose capacity and ability to incorporate new content vary by subclass (\Cref{sec:prelim}). 


\item[Forget policy.] The evolution operator $\mathcal{E}$ acts on content already stored in memory, determining what should be consolidated, revised, demoted, or discarded, e.g., when knowledge becomes stale under a regime shift. Together, $\mathcal{F}$ and $\mathcal{E}$ form the write operator $\mathcal{W}_\phi$ in \Cref{eq:generic-memory}, with $\mathcal{F}$ operating on newly generated content and $\mathcal{E}$ on existing memory (\Cref{sec:agentic-mechanism}).

\item[Foundation model (time-series foundation model, TSFM).] A model pretrained on a broad mixture of time-series data and applied across datasets, frequencies, and domains, often zero-shot; its verbatim attention cache and its use as a retrieval backbone recur throughout the survey, and it raises the confound that an external memory's apparent gain may already be encoded parametrically during pretraining (\Cref{sec:explicit-raw}, \Cref{sec:resources}).

\item[Instance retrieval memory.] Retrieval memory that retains concrete historical cases, i.e., windows, trajectories, or context--future pairs, so a query can be grounded in specific observed precedents (\Cref{sec:retrieval-instance}).

\item[Internal memory.] Historical information encoded in dynamically updated, fixed-dimensional latent states rather than retained as separate, addressable entries. Learned parameters govern how these states are updated but do not themselves constitute memory. Internal memory is subdivided into recurrent-state and structured-state memory (\Cref{sec:parametric}).

\item[Knowledge retrieval memory.] Retrieval memory that stores information other than another realization of the target series, e.g., text, exogenous events, or structured domain knowledge, providing context about the underlying process rather than a trajectory to imitate (\Cref{sec:retrieval-knowledge}).

\item[Latent retrieval memory.] Retrieval memory whose persistently stored and retrieved object is itself a compact derived representation (an embedding, latent state, or learned token) rather than the original observation (\Cref{sec:retrieval-latent}).

\item[Learned memory.] Explicit memory that stores compact representations learned from data, such as prototypes, centroids, or shared pattern banks, and explains a query as a mixture of a few such entries rather than many concrete episodes (\Cref{sec:explicit-learned}).

\item[Learning Using Privileged Information (LUPI).] A foundational paradigm in which a model exploits information available during training but unavailable at inference (e.g., via teacher--student distillation); proposed as a training signal for supervising what a time-series memory system should retain, compress, or discard (\Cref{sect: chall7}).

\item[Long-horizon dependence (P1).] The problem setting in which useful predictive information lies far outside any feasible input window, e.g., a value tied to the previous year's seasonal peak (\Cref{sec:p1}).

\item[Lookback window.] The immediate span of recent observations, $\mathbf{x}_{t-L+1:t}$, that a model conditions on directly; memory is defined relative to this window as information the model uses that is \emph{not} present in it (\Cref{sec:prelim}).

\item[Memory (of a model).] The memory of a time-series model refers to any mechanism that makes historical information available to the predictor beyond the information contained in its current prediction target.(Definition~\ref{def:memory}, \Cref{sec:prelim}).

\item[Memory management (evaluation measure).] A proposed benchmark quantity scoring the quality of write, update, and forget decisions over repeated episodes under a fixed storage budget (\Cref{sec:resources}).

\item[Nearest-neighbor search.] A hard (arg max) form of content-based addressing that retrieves the single closest stored entry to a query rather than a weighted mixture; the sharpened limit of \Cref{eq:explicit-read} (\Cref{sec:explicit}).

\item[Non-stationarity / regime shift / drift (P3).] The problem setting in which the data-generating process changes over time, so historical data can become uninformative or actively misleading; the central motivation for forget/update policies in agentic memory (\Cref{sec:p3}).

\item[Persistence (axis).] One of the four axes for comparing memory mechanisms, describing how long stored information survives: \emph{per-sequence} (resets with each new input), \emph{per-dataset} (a fixed store or index, frozen after training/construction), or \emph{online} (continues to evolve across an interaction) (\Cref{sec:4axes}).

\item[Procedural memory.] An agentic memory type that stores knowledge of how to perform a task, e.g., which tool or retrieval strategy to use in a given situation; typically written once evidence accrues and decays slowly, since it can outlive individual episodic entries (\Cref{tab:agentic-types}).

\item[Read/write mechanism (axis).] One of the four axes for comparing memory mechanisms, describing how information enters and leaves a store: gated recurrence, structured linear recurrence, content attention, nearest-neighbor search, or controller-issued operations (\Cref{sec:4axes}).

\item[Recurrent-state memory.] Internal memory in which a hidden state $\mathbf{h}_t$ is updated at each time step from the previous state and the current observation (e.g., RNN, LSTM, GRU), so it depends on the entire history $\mathbf{x}_{1:t}$ while remaining fixed in dimension (\Cref{sec:internal-recurrent}).

\item[Recurring patterns (P2).] The problem setting in which current conditions resemble a past event whose outcome carries predictive value, e.g., matching a demand spike to a similar one the previous year; internal state compression blends past episodes and is ill-suited to it, motivating explicit, addressable memory (\Cref{sec:p2}).

\item[Representation (axis).] One of the four axes for comparing memory mechanisms: \emph{implicit} representation is entangled in weights or a dense state; \emph{explicit} representation consists of identifiable slots, vectors, or records (\Cref{sec:4axes}).

\item[Retrieval-augmented generation (RAG).] The foundational (originally NLP) paradigm of conditioning a model's output on documents retrieved from an external corpus; the conceptual ancestor of retrieval-augmented memory for time series (\Cref{sec:fourclasses}).

\item[Retrieval-augmented memory.] Memory that retains information in an external store and selectively recalls a query-dependent subset (top-$K$ by similarity) to condition the current prediction; capacity is unbounded and grows independently of the predictor. Subdivided into instance, latent, and knowledge retrieval memory (\Cref{sec:retrieval}).

\item[Retrieval quality (evaluation measure).] A proposed benchmark quantity measuring whether the relevant record appears among the retrieved items, assessed against random, no-retrieval, and (where possible) oracle controls (\Cref{sec:resources}).

\item[Semantic memory.] An agentic memory type that stores generalized knowledge distilled from past observations, e.g., recurring regimes, calendar rules, or exogenous facts, rather than specific episodes; written once evidence accrues, with medium decay (\Cref{tab:agentic-types}).

\item[Slot collapse.] A failure mode of explicit memory in which multiple slots converge on similar content, leaving an effective capacity far below the nominal number of slots $N$ (\Cref{sec:explicit-write}).

\item[Slot memory.] Explicit memory built from dedicated, learned-addressing locations (descended from the Neural Turing Machine and Differentiable Neural Computer) whose contents are model states rather than raw observations (\Cref{sec:explicit-slot}).

\item[Sparse and hierarchical data (P4).] The problem setting in which an individual series lacks sufficient local history (e.g., a new product or newly deployed sensor) and must draw on related series or higher levels of aggregation (\Cref{sec:p4}).

\item[Structured-state memory (state-space model, SSM).] Internal memory in which a fixed-dimensional hidden state evolves through parameterized, often linear, state transitions (e.g., S4, Mamba) rather than nonlinear gating, admitting a closed-form compression map (\Cref{eq:ssm-unrolled}) and, in selective variants, input-dependent transitions (\Cref{sec:internal-structured}).

\item[Temporal validity (evaluation measure).] A proposed benchmark quantity testing whether a model appropriately rejects or down-weights memories originating from an obsolete regime (\Cref{sec:resources}).

\item[Verbatim memory.] Explicit memory that gives every observed element its own entry, encoded individually and never merged or summarized into a coarser state, so forgetting occurs only through truncation; exemplified by the attention context of a Transformer (\Cref{sec:explicit-raw}).

\item[Working memory.] Non-persistent, within-episode state that an agent accumulates during a single analysis (intermediate evidence, tool outputs, partial conclusions) and discards once the episode ends, distinguishing it from the three persistent agentic memory types (\Cref{sec:agentic-methods}).

\item[Write policy.] The agentic-memory policy $\mathcal{F}$ that determines which artifacts of an interaction are persisted to the store and in what abstracted form (\Cref{sec:agentic-mechanism}).

\end{description}


\providecommand{\authorfill}[1]{\textcolor{red}{\textbf{[#1]}}}

\section{Supplementary Material: Search and Selection Protocol}
\label{sec:selection}

This section documents how the surveyed corpus was assembled. We report the time window, the search terms, the inclusion and exclusion criteria, and the resulting composition of the corpus, so that the coverage claims in \Cref{tab:relatedsurveys} and the white space identified in \Cref{tab:taskmatrix} can be independently assessed.

\subsection{Scope and Time Window}
\label{sec:selection-window}

The primary search window is \textbf{2015--2026}, matching the temporal axis of \Cref{fig:time_series_memory_evolution}. The lower bound is set at 2015 because the memory mechanisms this survey is organized around, namely explicitly addressable stores, retrieval augmentation, and agentically curated stores, only begin to appear in the time-series literature after the neural external-memory architectures of 2014--2016. The search was last refreshed on \textbf{04/09/2026}.

Four works published before this window are retained as \emph{foundational exceptions}: the recurrent formulation~\cite{elman1990finding}, gated recurrence~\cite{hochreiter1997lstm,cho2014learning}, and the Neural Turing Machine~\cite{graves2014neural}. These are included not as surveyed time-series methods but because the mechanisms reviewed in \Cref{sec:parametric} and \Cref{sec:explicit} are direct descendants of them, and the memory lineage is not intelligible without them. 

\subsection{Sources and Search Terms}
\label{sec:selection-terms}

We searched arXiv and Google Scholar and also manually reviewed the proceedings of NeurIPS, ICML, ICLR, KDD, IJCAI, AAAI, and ICDM, as well as relevant journals represented in our reference set, including IEEE TPAMI, TMLR, International Journal of Forecasting, Journal of Forecasting, Neurocomputing, and Engineering Applications of AI. Because similar memory mechanisms are often described using different terms—for example, a memory bank'' in one paper may be called a prototype set'' or ``datastore'' in another—we did not rely on keyword searches alone. We complemented the initial search with backward snowballing, by examining the references of included papers, and forward snowballing, by following later works that cite the key papers in each memory category.

\Cref{tab:search-terms} lists the query terms. Each mechanism term was conjoined with at least one time-series term.

\begin{table*}[t]
\centering\small
\renewcommand{\arraystretch}{1.3}
\caption{Search terms. Each mechanism term (right) was conjoined with at least one domain term (top block) and, where a class-specific vocabulary exists, with its class terms. Terms are lower-cased and matched on title, abstract, and keywords.}
\label{tab:search-terms}
\setlength{\tabcolsep}{5pt}
\begin{tabularx}{\textwidth}{l >{\raggedright\arraybackslash}X}
\toprule
\textbf{Group} & \textbf{Terms} \\
\midrule
\rowcolor{rowfill}
Domain (required)
& ``time series'', ``temporal'', ``forecasting'', ``spatiotemporal'', ``sequential data'', ``multivariate series'', ``anomaly detection'', ``imputation'', ``time-series classification'' \\

Cross-cutting memory
& ``memory'', ``memory-augmented'', ``external memory'', ``memory network'', ``memory module'', ``memory bank'', ``long-term dependency'', ``long-range dependence'' \\

\rowcolor{rowfill}
\mh{cInternal}{Internal}
& ``recurrent'', ``LSTM'', ``GRU'', ``hidden state'', ``state space model'', ``structured state'', ``selective state'', ``Mamba'', ``linear recurrence'' \\

\mh{cExplicit}{Explicit}
& ``slot memory'', ``memory slot'', ``addressable memory'', ``neural Turing machine'', ``differentiable neural computer'', ``prototype'', ``pattern bank'', ``shapelet'', ``codebook'', ``Hopfield'', ``key-value memory'' \\

\rowcolor{rowfill}
\mh{cRetrieval}{Retrieval}
& ``retrieval-augmented'', ``retrieval augmentation'', ``RAG'', ``nearest neighbour retrieval'', ``datastore'', ``exemplar retrieval'', ``analogue retrieval'', ``retrieve and refine'', ``in-context examples'' \\

\mh{cAgentic}{Agentic}
& ``agent'', ``agentic'', ``LLM agent'', ``tool use'', ``episodic memory'', ``semantic memory'', ``procedural memory'', ``reflection'', ``experience replay buffer'', ``self-improving'' \\

\rowcolor{rowfill}
Evaluation
& ``benchmark'', ``long-context'', ``needle in a haystack'', ``retrieval benchmark'', ``memory evaluation'' \\
\bottomrule
\end{tabularx}
\end{table*}

\subsection{Inclusion and Exclusion Criteria}
\label{sec:selection-criteria}

We screened the retrieved papers in two stages, first by title and abstract and then by full text. The inclusion and exclusion criteria are summarized in \Cref{tab:selection-criteria}. At the full-text stage, we did not require a paper to explicitly use the term \emph{memory}. Instead, we asked whether its method could be described using the four axes in \Cref{sec:4axes}: what information is retained, how it is written, how it is retrieved, and how long it remains available. Papers for which these aspects could not be clearly identified were considered outside the scope of the survey.

\begin{table*}[t]
\centering\small
\renewcommand{\arraystretch}{1.3}
\caption{Inclusion and exclusion criteria. IC = inclusion, EC = exclusion. A work was retained only if it satisfied every inclusion criterion and triggered no exclusion criterion.}
\label{tab:selection-criteria}
\setlength{\tabcolsep}{5pt}
\begin{tabularx}{\textwidth}{l >{\raggedright\arraybackslash}X}
\toprule
\textbf{ID} & \textbf{Criterion} \\
\midrule
\rowcolor{rowfill}
IC1 & \textbf{Temporal data.} The method operates on time series, event sequences, or spatiotemporal streams as its primary modality. \\ IC2 & \textbf{Identifiable memory mechanism.} The work can be placed on all four axes of \Cref{sec:4axes}: what is retained, how it is written, how it is read, and how long it persists. \\ \rowcolor{rowfill} IC3 & \textbf{Memory is load-bearing.} Retention beyond the immediate input window is a stated mechanism of the contribution, not an incidental property of the backbone. \\ IC4 & \textbf{Task relevance.} The work targets forecasting, classification, anomaly detection, imputation, or the emerging reasoning and decision-making settings of \Cref{sec:crosscutting}. \\ \rowcolor{rowfill} IC5 & \textbf{Retrievable and in English.} Full text available; peer-reviewed, or a preprint meeting EC5. \\ \midrule EC1 & \textbf{Non-temporal modality only.} Memory architectures evaluated solely on text, static vision, or tabular data, unless they are an anchor mechanism a surveyed time-series method descends from (\Cref{sec:selection-window}). \\ \rowcolor{rowfill} EC2 & \textbf{Classical methods.} Statistical and non-deep approaches (ARIMA, exponential smoothing, classical nearest-neighbour forecasting) that predate the memory framing. \\ \rowcolor{rowfill} EC3 & \textbf{Unvetted preprints.} Preprints were admitted only where they are the sole source for an emerging mechanism, and are marked as such; this applies chiefly to the agentic and 2026 retrieval literature, where archival versions do not yet exist. \\ \rowcolor{rowfill} EC4 & \textbf{Superseded versions.} Where a workshop paper, preprint, and archival version describe the same system, only the most mature version is counted. \\
\bottomrule
\end{tabularx}
\end{table*}

\subsection{Threats to Validity}
\label{sec:selection-threats}

Three limitations should be noted. First, \emph{terminological drift}: there is no shared vocabulary for temporal memory, so keyword-based searches may miss relevant work that describes similar mechanisms using domain-specific terms. We used backward and forward snowballing to reduce this risk, but some slot-memory and prototype-based methods may still have been missed. Second, \emph{recency bias}: the most recent literature, particularly work from 2025--2026, includes preprints whose archival status is not yet settled. The emerging literature on agentic memory should therefore be viewed as a current snapshot rather than a complete or definitive record. Third, \emph{classification judgement}: some works fall near the boundaries between memory classes and require judgement when applying our taxonomy. For example, Titans~\cite{behrouz2025titans} learns a write rule at test time but is classified as explicit because its memory store is parametric, while iTransformer~\cite{liu2024itransformer} is classified as verbatim despite consolidating information across time. Different choices about these boundaries may therefore lead to slightly different class assignments. To make these decisions transparent, the accompanying paper collection records the classification assigned to each included work, allowing the choices to be inspected and challenged.


\end{document}